\documentclass[10pt,twocolumn,letterpaper]{article}

\usepackage{3dv}
\usepackage{times}
\usepackage{float}  
\usepackage{subfigure}
\usepackage{epsfig}
\usepackage{graphicx}
\usepackage{wrapfig}
\makeatletter
\def\adl@drawiv#1#2#3{%
        \hskip.5\tabcolsep
        \xleaders#3{#2.5\@tempdimb #1{1}#2.5\@tempdimb}%
                #2\z@ plus1fil minus1fil\relax
        \hskip.5\tabcolsep}
\newcommand{\cdashlinelr}[1]{%
  \noalign{\vskip\aboverulesep
           \global\let\@dashdrawstore\adl@draw
           \global\let\adl@draw\adl@drawiv}
  \cdashline{#1}
  \noalign{\global\let\adl@draw\@dashdrawstore
           \vskip\belowrulesep}}
\makeatother

\usepackage{tikz}
\usepackage{comment}
\usepackage{amsmath}
\usepackage{amssymb}
\usepackage{color}
\usepackage{caption}
\usepackage{booktabs}
\usepackage{multirow}
\usepackage{multicol}

\usepackage[pagebackref=true,breaklinks=true,colorlinks,bookmarks=false]{hyperref}
\usepackage[accsupp]{axessibility} 
\usepackage{xspace}

\makeatletter
\DeclareRobustCommand\onedot{\futurelet\@let@token\@onedot}
\def\@onedot{\ifx\@let@token.\else.\null\fi\xspace}
\def\eg{\emph{e.g}\onedot} 
\def\ie{\emph{i.e}\onedot}

\def\etal{\emph{et al}\onedot}
\makeatother

\usepackage[capitalize]{cleveref}
\crefname{section}{Sec.}{Secs.}
\Crefname{section}{Section}{Sections}
\Crefname{table}{Table}{Tables}
\crefname{table}{Tab.}{Tabs.}

\definecolor{tan}{rgb}{0.82, 0.71, 0.55}
\definecolor{navyblue}{rgb}{0.0, 0.0, 0.5}

\threedvfinalcopy 

\def\threedvPaperID{283} 

\ifthreedvfinal\fi
\begin{document}

\newcommand{\myparagraph}[1]{\noindent\textbf{#1}\,\,}
\title{HoW-3D: Holistic 3D Wireframe Perception from a Single Image}

\author{Wenchao Ma$^1$ \qquad Bin Tan$^1$ \qquad Nan Xue$^{1}$\thanks{Corresponding author} \qquad Tianfu Wu$^2$ \qquad Xianwei Zheng$^3$ \qquad Gui-Song Xia$^{1}$\\
$^1$ School of Computer Science, Wuhan University, Wuhan, China\\
$^2$ Department of ECE, North Carolina State University, Raleigh, USA\\
$^3$ LIESMARS, Wuhan University, Wuhan, China\\\\
\emph{Code\&Dataset}: \url{https://github.com/Wenchao-M/HoW-3D}
}

\maketitle
\begin{abstract}
This paper studies the problem of holistic 3D wireframe perception (HoW-3D), a new task of perceiving both the visible 3D wireframes and the invisible ones from single-view 2D images. 
As the non-front surfaces of an object cannot be directly observed in a single view, estimating the non-line-of-sight (NLOS) geometries in HoW-3D is a fundamentally challenging problem and remains open in computer vision. 
We study the problem of HoW-3D by proposing an ABC-HoW benchmark, which is created on top of CAD models sourced from the ABC-dataset with 12k single-view images and the corresponding holistic 3D wireframe models. 
With our large-scale ABC-HoW benchmark available, we present a novel Deep Spatial Gestalt (DSG) model to learn the visible junctions and line segments as the basis and then infer the NLOS 3D structures from the visible cues by following the Gestalt principles of human vision systems. 
In our experiments, we demonstrate that our DSG model performs very well in inferring the holistic 3D wireframes from single-view images. 
Compared with the strong baseline methods, our DSG model outperforms the previous wireframe detectors in detecting the invisible line geometry in single-view images and is even very competitive with prior arts that take high-fidelity PointCloud as inputs on reconstructing 3D wireframes.
\end{abstract}

\section{Introduction}
\begin{figure}
    \centering
     \subfigure[\label{fig:teaser-visible}]{
      \includegraphics[trim={0 14 0 10},clip, width=0.22\linewidth]{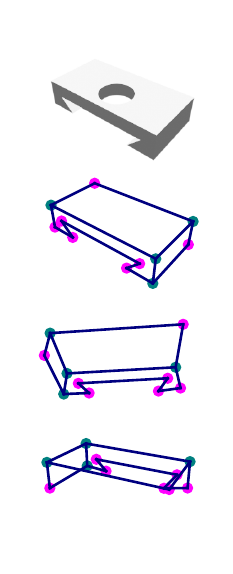}
     }
     \hfill
     \subfigure[\label{fig:teaser-mesh}]{
      \includegraphics[trim={0 14 0 10},clip, width=0.22\linewidth]{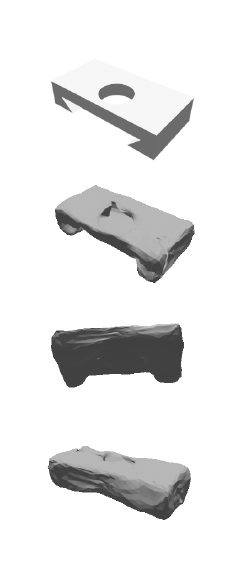}
    }
     \hfill
     \subfigure[\label{fig:teaser-pointcloud}]{
      \includegraphics[trim={0 14 0 10},clip, width=0.22\linewidth]{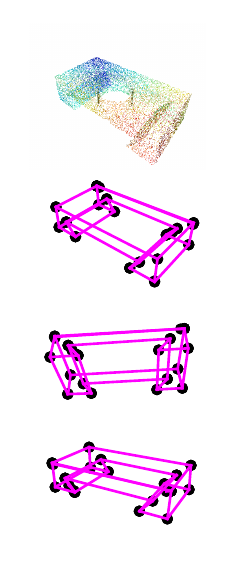}
     }
     \hfill
     \subfigure[\label{fig:teaser-how}]{
      \includegraphics[trim={0 14 0 10},clip, width=0.22\linewidth]{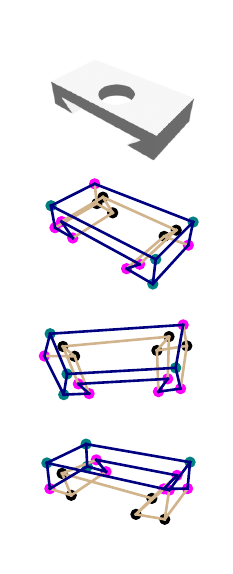}
    }
    \caption{\small An illustrative comparison between different specifications of 3D geometric perception. (a) Conventional Single-View 3D Wireframe Perception, (b) Single-View 3D Mesh Reconstruction, (c) Holistic 3D Wireframe Perception from Point Clouds, and (d) Holistic 3D Wireframe Perception from \textbf{Single-View Images}. The inputs of these specifications are displayed in the top row, and the resulted 3D models are shown in different views.
    }
    \label{fig:teaser}
    \vspace{-4mm}
\end{figure}

Since the pioneering work of David Marr~\cite{Marr82}, 3D reconstruction from 2D images has been recognized as one main goal and remained one main grand challenge, of computer vision. Among many different forms, the 3D wireframe that aims to compute the vectorized line drawing graphs~\cite{Zhou-Manhattan-Wireframe} provides a parsimonious object and scene representation for object and scene understanding, object manipulation, and interaction, and computer-aided design.

As illustrated in Fig.~\ref{fig:teaser}, extracting the 3D models (including meshes and wireframe models) from input images is a fundamentally challenging task, especially from a single-view 2D image (\ie, monocular) due to the ill-posed nature of 2D-3D projection: The non-front surfaces of a 3D object cannot be directly observed with only one viewpoint. 
Thus, perceiving 3D wireframe from an input 2D image entails estimating the non-line-of-sight geometry. In the literature, the problem has been addressed in a simplified setting which either only aims at recovering the visible lines (Fig.~\ref{fig:teaser-visible}) in~\cite{Zhou-Manhattan-Wireframe}, learning the mesh models (\cref{fig:teaser-mesh}) with strong class priors~\cite{pixel2mesh}, or assumes a high-fidelity, often dense and accurate 3D point cloud can be computed at a first-place (Fig.~\ref{fig:teaser-pointcloud}) as done in~\cite{liu2021pc2wf}. 
We ask: \textit{Can we estimate the full 3D wireframe directly from a single-view 2D image (Fig.~\ref{fig:teaser-how}) without resorting to the intermediate point could representation? And, how far can we go to see the invisible directly from a single 2D image?}

One hallmark of human vision is the capability of perceiving 3D from 2D inputs in an effective and efficient way. The capability is attributed to the strong and sophisticated bottom-up/top-down integrative information processing system in our brain. Before the 3D wireframe representation re-emerged recently as a desirable representation of object and scene, it has been a long history of single-view 3D reconstruction based on mesh or point cloud/volume representation. To counter the ill-posed nature of the problem, prior visual knowledge and/or commonsense has to be integrated. In the pioneering work of single-view 3D reconstruction~\cite{han-full-primal-sketch, Make3D}, a set of manually designed Gestalt rules is introduced in~\cite{han-full-primal-sketch} accounting for the structural regularities between detected visual primitives such as line segments, junctions, and rectangles, while a set of commonsense knowledge regarding the height distribution of a person or a building and the size distribution of a tree or a car in a scene is utilized in~\cite{Make3D}. Although promising and impressive progress has been made ever since, they are only available for limited scenarios. 

With the resurgence of deep neural networks (DNNs) in the past decade, single-view 3D reconstruction has witnessed remarkable progress~\cite{pixel2mesh,chang2015shapenet,xiao2013sun3d}, thanks to the end-to-end deep learning capability of DNNs distilling semantic priors for depth estimation and shape reconstruction of objects in a scene from large-scale datasets. Despite powerful,  they are mostly pure bottom-up data-driven and have the known issue of not generalizing well beyond training. When the task is to infer the full 3D wireframe from a single-view 2D image (Fig.~\ref{fig:teaser-how}), prior arts on single-view 3D reconstruction lack the capability of inferring the invisible boundary geometries. 
For example in \cref{fig:teaser-mesh}, the mesh reconstruction approach proposed in~\cite{pixel2mesh} uses a prior topology to adaptively deform the mesh template, and thus can not handle the unknown topology in 3D wireframe perception. 
Alternatively, simplified settings as shown in Fig.~\ref{fig:teaser-visible} and Fig.~\ref{fig:teaser-pointcloud} are used. More specifically, the Gestalt principles as adopted and advocated   by~\cite{han-full-primal-sketch,Make3D} should be integrated into deep learning. To that end, the pure bottom-up data-driven design needs to be revisited to resemble the bottom-up/top-down integrative information processing paradigm well known for 3D perception and understanding in human vision.

This paper makes the first attempt to study \emph{the holistic 3D wireframe perception from single-view images} (dubbed as HoW-3D), as shown in ~\cref{fig:teaser-how}. The key idea is to learn the line geometry between the projective 2D counterparts and the corresponding Euclidean 3D parameters using a proposed Deep Spatial Gestalt (DSG) model. As illustrated in Fig.~\ref{fig:model-architecture}, the proposed DSG model consists of three components: (i) A bottom-up visible 2D-2.5D wireframe perception module built on a recently proposed Holistic Attraction Field Map representation for 2D wireframe parsing~\cite{xue2020holistically}; (ii) A top-down query-based design for inferring invisible (hidden)  junctions from the bottom-up data evidence built on the Transformer encoder and decoder framework proposed in the DETR framework~\cite{carion2020end}; (iii) A bottom-up/top-down integrative reasoning module for 3D wireframe perception and refinement built on Graph Neural Networks.
To train and benchmark the proposed HoW-3D task, we build the ABC-HoW dataset based on the large-scale ABC dataset~\cite{koch2019abc} for geometric deep learning with CAD models. We rendered 12258 images and the holistic 3D wireframe models from 991 objects and divided them into the training and testing splits by objects to avoid the potential overfitting. Statistically, our ABC-HoW dataset contains 11179 images of 900 objects and 1079 images of 91 objects for training and testing, respectively.

\myparagraph{Our Contributions.} This paper makes three main contributions in the field of single-view 3D wireframe perception as follows:
\begin{itemize}
\vspace{-2mm}
\item[-] It presents a dataset ABC-HoW for a new and challenging task of HoW-3D, which enables studying and benchmarking single-view 3D reconstruction methods from recovering only the visible wireframes to the holistic one.
\item[-] It proposes a DNN-based bottom-up/top-down integrative approach, the Deep Spatial Gestalt (DSG) model, to explicitly learn the relationship between the visible primitives of the 3D wireframe and the invisible ones. 
\item[-] Compared with the state-of-the-art on 2D visible wireframe parsing~\cite{xue2020holistically} and 3D wireframe reconstruction from point clouds~\cite{liu2021pc2wf}, our proposed DSG model obtains superior performance to HAWP~\cite{xue2020holistically} in detecting invisible parts of the single-view image and shows competitive results with the PC2WF~\cite{liu2021pc2wf} on holistic 3D wireframe reconstruction.
\end{itemize}

\section{Related Work}
\myparagraph{Deep Single-view Shape Reconstruction.}
Recently, the deep learning based approaches significantly advanced this goal by learning the 3D models in a data-driven way with a large-scale ShapeNet~\cite{chang2015shapenet,modelNet} in different forms of the voxels~\cite{choy20163d,girdhar2016learning}, point clouds~\cite{fan2017point}, triangle meshes~\cite{huang2015single,pixel2mesh}, and implicit functions~\cite{chen2019learning,mescheder2019occupancy}. Although impressive, those approaches mainly focus on learning the category-specific shape priors (\eg, chairs and planes) for 3D perception. For the task of HoW-3D, those approaches can be an initialization to obtain the 3D shapes and then leverage a follow-up scheme (such as PolyFit~\cite{Polyfit} or PC2WF~\cite{liu2021pc2wf}) to get the holistic 3D wireframes. In contrast to those approaches, we focus on the learning of 3D wireframe models in a geometric perspective without incurring any semantic labels.

\myparagraph{Wireframe Perception and Reconstruction.}
Detecting the line segments or wireframes in the 2D image is a longstanding and basic problem, which has always been concerned by researchers~\cite{burns1986extracting,Matas2000robust,Gioi2010LSD,almazan2017mcmlsd,huang2018learning,xue2019learning,RegionalAttraction,zhang2019ppgnet,zhou2019end,xue2020holistically,xu2021letr,XueXBZS18,XiaDG14}. With the development of deep learning and the emergence of large-scale datasets~\cite{huang2018learning}, 2D deep wireframe models ~\cite{huang2018learning,xue2019learning,RegionalAttraction,xue2020holistically,zhang2019ppgnet,zhou2019end,xu2021letr} achieved credible performance. For the 3D wireframe perception, previous unsupervised methods ~\cite{lin2015line,lu2019lfast,nan2017ployfit} detect the 3D line segments or polygonal surfaces from the point clouds of the buildings. Benefiting from the large-scale CAD dataset~\cite{koch2019abc}, recent works ~\cite{PIE-Net,liu2021pc2wf,ComplexGen} achieve promising 3D wireframe reconstruction results based on the high fidelity 3D point clouds in a supervised manner.
In contrast to those works, we make efforts on the perception of 3D holistic wireframes from single-view images rather than the costly point clouds by exploiting the gestalt principles with deep neural networks.

\myparagraph{Single-view 3D Geometry Structure Reconstruction.}
Early works studied line drawings as a kind of primal sketch~\cite{Marr82} for 3D shape reconstruction, ~\cite{Marill91emulating,lipson1996optimization,shoji20013,liu2007plane,wang20093d} tried to reconstruct the holistic 3D wireframe from the holistic 2D wireframe line drawing generated by hand drawing or orthogonal projection. These methods usually rely on the structural assumptions of 3D objects to obtain the 3D wireframe models via optimization processes. And recent work~\cite{FaceFormer} approaches this problem in a data-driven way by projecting the CAD models in ABC dataset~\cite{koch2019abc} into line drawings. For the image-based approaches, Zhou~\etal~\cite{Zhou-Manhattan-Wireframe} proposed a CNN-based method to reconstruct the visible 3D wireframes of buildings in the image under the Manhattan assumption. There are also excellent works~\cite{WU-3D-INN-IJCV,reddy2019occlusion,li2019deep, Suwajanakorn2018discovery, Cootes1995activate,zia2013detialed,zia2015towards,han-full-primal-sketch} that attempted to detect both the visible and invisible parts or keypoints in the images to reconstruct the holistic 3D geometry structure of the specific class of objects (vehicles for example) with strong semantic priors. 

\myparagraph{3D Shape Datasets.}
There have been many large-scale 3D shape datasets~\cite{chang2015shapenet,modelNet, Things10k, PrincetonSB} that boosted the research of 3D object reconstruction and understanding by using meshes or point clouds. Most recently, the ABC~\cite{koch2019abc} dataset provided the analytic boundary representation of surfaces and curves for the 3D models, which motivates recent works~\cite{PIE-Net,liu2021pc2wf, FaceFormer, ComplexGen} to use the selected objects from the ABC dataset for the 3D parametric boundary perception from the sampled point clouds or line drawings. Besides the sampled ABC dataset, PC2WF~\cite{liu2021pc2wf} also prepared a small-scale \emph{Furniture} dataset that consists of 250 objects from Google 3D Warehouse with only 5 categories, which encodes strong class priors for 3D wireframe perception. For the task of single-view HoW-3D without incurring the semantic labels, to the best of our knowledge, there is no off-the-shelf dataset for either training a DNN or evaluating the reconstruction results. Accordingly, we present our ABC-HoW benchmark based on the original ABC dataset for HoW-3D.

In summary, most current works either requires ideal inputs (line drawing and high-quality point clouds) or are limited by a specific class of objects with strong semantic priors. Although Zhou~\etal~\cite{Zhou-Manhattan-Wireframe} studied the 3D wireframe perception from single-view images, they only focus on the depth estimation of visible 3D junctions and line segments. In this work, we aim to study the Gestalt principle to break the front-view surface to perceive non-line-of-sight geometric elements for the holistic 3D wireframe reconstruction without any semantics prior to single-view images. 

\section{The ABC-HoW Dataset}
\begin{figure}
\begin{minipage}{0.9\linewidth}
 \resizebox{\linewidth}{!}{
  \begin{tabular}{c|cccc|cccc}
    \toprule
         & \multicolumn{4}{c|}{Train (11179 samples) } & \multicolumn{4}{c}{Test (1079 samples)}\\
         & min & max & mean & std& min & max & mean & std\\\midrule
    $J_{\text{vis}}$ & 5 & 30 & 11.67 & 4.22 & 5 & 25 & 11.2 & 3.39\\
    $J_{\text{hidden}}$ & 1 & 20 & 3.67 & 2.60 & 1 & 15 & 3.27 & 2.00 \\\midrule
    $L_{\text{vis}}$ & 6 & 39 & 14.23 & 5.16 & 6 & 33 & 13.71 & 4.17\\
    $L_{\text{hidden}}$ & 2 & 37 & 8.66 & 5.28 & 3 & 22 &  7.84 & 3.88 \\
    \bottomrule
    \end{tabular}
    }
\vspace{-3mm}
  \captionof{subfigure}{Statistics for the number of visible and invisible junctions and lines in ABC-HoW dataset}
  \label{fig:abc-how-stat}
 \end{minipage}
 \begin{minipage}{0.9\linewidth}
    \centering
    \begin{minipage}{0.6\linewidth}
        \centering
            \includegraphics[width=0.95\linewidth]{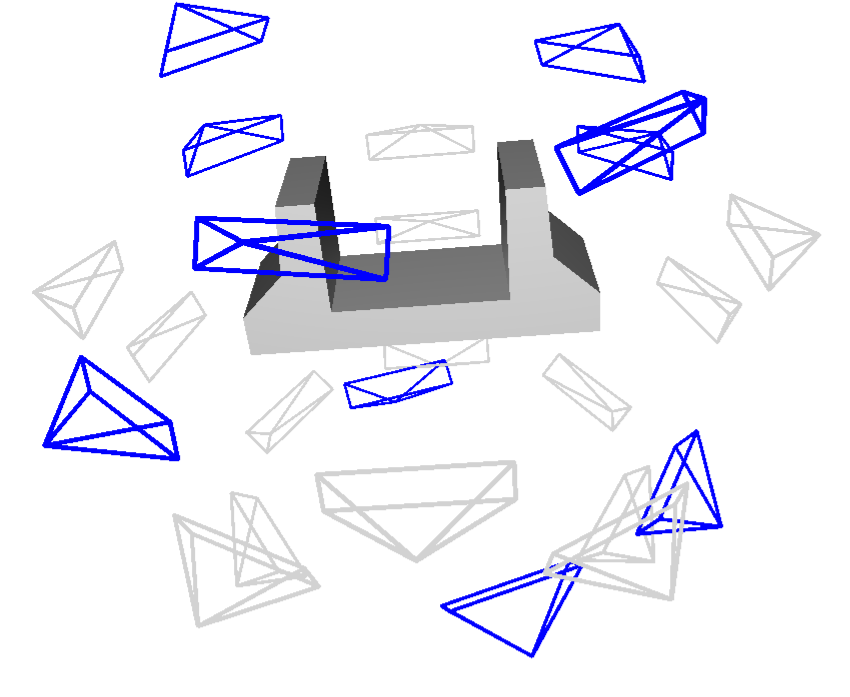}
        \end{minipage}
    \begin{minipage}{0.35\linewidth}
        \centering
            \includegraphics[width=0.7\linewidth]{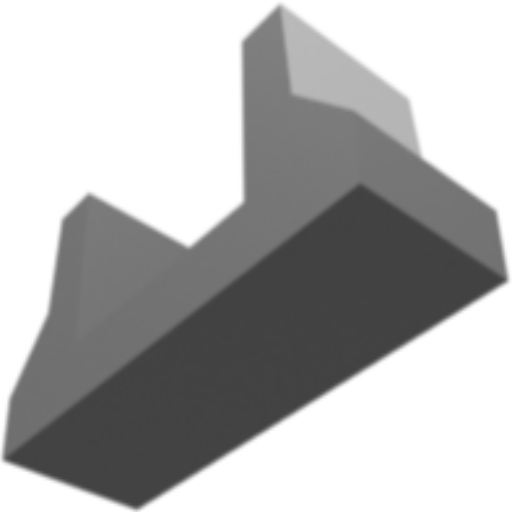}
            \includegraphics[width=0.7\linewidth]{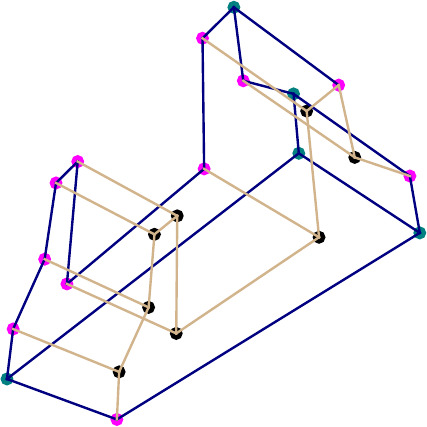}
    \end{minipage}
    \vspace{-3mm}
    \captionof{subfigure}{The rendering process for ABC-HoW.}
    \label{fig:abc-how-render}
    \vspace{-3mm}

 \end{minipage}
  \caption{The statistics (a) and illustration (b) of our ABC-HoW dataset. For each selected object from the ABC dataset~\cite{koch2019abc}, we render the images and 3D wireframe models for the randomly-generated view candidates, and then manually select the meaningful viewpoints (marked in \textcolor{blue}{blue}) for the HoW-3D task. The right side displays a pair of the rendered image and the corresponding 3D wireframe model. The visible and hidden line segments in the rendering viewpoint are marked in \textcolor{navyblue}{navyblue} and \textcolor{tan}{tan}, respectively.
    }
   \vspace{-5mm}
   \label{fig:how-3d-data}
    
\end{figure}

In this section, we describe the detail of the proposed ABC-HoW dataset for holistic 3D wireframe perception. We create our ABC-HoW dataset based on the large-scale ABC dataset~\cite{koch2019abc}. As shown in Fig.~\ref{fig:how-3d-data}, our dataset has 12k samples rendered from 991 objects, we split them into 11179 images from 900 objects and 1079 images from 91 objects for training and testing, respectively. Each data sample contains a rendered image and a 3D wireframe model in the rendering viewpoint. The rendering process is described in Fig.~\ref{fig:how-3d-data}(b). For the paired 3D wireframe model in each data sample, we label the visibility of all the lines and junctions with different colors. 

\subsection{CAD Model Preparation}
We select a subset of CAD models that are dominated by straight line segments as the main physical object boundaries from the first 10 chunks of the ABC dataset~\cite{koch2019abc}, which yields a total of 3000 CAD models for the HoW-3D task. 
For each CAD model, we rendered $24$ images from the randomly generated sparse views and filter out some objects by the following criteria:
(1) Removing the duplicated or extremely similar objects such as cubes with different sizes to avoid unexpected data contamination;
(2) Filtering out the objects with some too ``thin" parts to achieve the HoW-3D task;
(3) Considering the self-occlusion fact in the image formation process, there are some perceptually meaningless viewpoints that cannot achieve the HoW-3D even by a human should also be removed. We further demonstrate this in the supplementary material.
Finally, the $991$ CAD models and their corresponding perceptually meaningful viewpoints remained to get the paired data samples of input images and the 3D wireframe models. 

\subsection{Single-View Holistic 3D Wireframe Models}
We denote the 3D wireframe graph in the camera coordinate of a given viewpoint by $\mathcal{W}^{3d} = (\mathcal{J}^{3d},\mathcal{L})$, with the junction set $\mathcal{J}^{3d}=\{(X^m, Y^m, Z^m)\}_{m=1}^{|J^{3d}|} \subset \mathbb{R}^3$ and index set $\mathcal{L} = \{(m,n)\}$ that indicates the endpoint indices of line segments. The corresponding 2D wireframe graph $\mathcal{W}^{2d} = (\mathcal{J}^{2d}, L)$ can be obtained by projecting 3D junctions $\mathcal{J}^{3d}$ onto the 2D image plane by $J^{2d} = \Pi (K\cdot J^{3d}) = (x,y) \in\mathbb{R}^2 $, where $\Pi$ is the perspective projection operation and $K$ is the camera intrinsic. 
The index set $\mathcal{L}$ remains unchanged because the projection operation does not affect the adjacency relation between the points. 
After that, it is obvious that only a subgraph remains visible in the rendered image $I$. Therefore, we mark the visibility of the $m$-th junction by $v^m \in \{0,1\}$. If it is visible, we set the $v^m = 1$ and vice versa. Based on the visibility of junctions, the visibility of line segments can be defined by their endpoints. Specifically, if there is at least one invisible junction as the endpoint of a line segment, it is regarded as the invisible one. Thus we split the junction set $\mathcal{J}^{2d}$ into visible junction set $\mathcal{J}_v = \{j^m \in \mathcal{J}\ \mid v^m = 1\}$ and the hidden set $\mathcal{J}_h = \{J^m \in \mathcal{J}\ \mid v^m = 0\}$, and the visible line set $\mathcal{L}_v = \{(m,n) \mid v^m \land v^n = 1 \}$
and the invisible line set $\mathcal{L}_h = \{(m,n) \mid v^m\vee v^n = 0 \}$.

In summary, for each data sample in the ABC-HoW dataset,c 
it contains a 2D image $I \in \mathbb{R}^{H\times W\times 3}$, the 3D junction set $\mathcal{J}^{3d}$, the line index set $\mathcal{L}$ for the line segments, and the visibility label $\mathcal{V} \in \{0,1\}^{|\mathcal{J}|}$ to represent the holistic 3D wireframe graph $W^{3d} = (\mathcal{J}^{3d}, L, \mathcal{V})$. The 2D wireframe graphs (with invisible primitives) can be obtained by replacing the 3D junctions $\mathcal{J}^{3d}$ with 2D points $\mathcal{J}^{2d}$.

\section{Deep Spatial Gestalt Model}
\begin{figure*}
  \centering
    \includegraphics[width=0.8\linewidth]{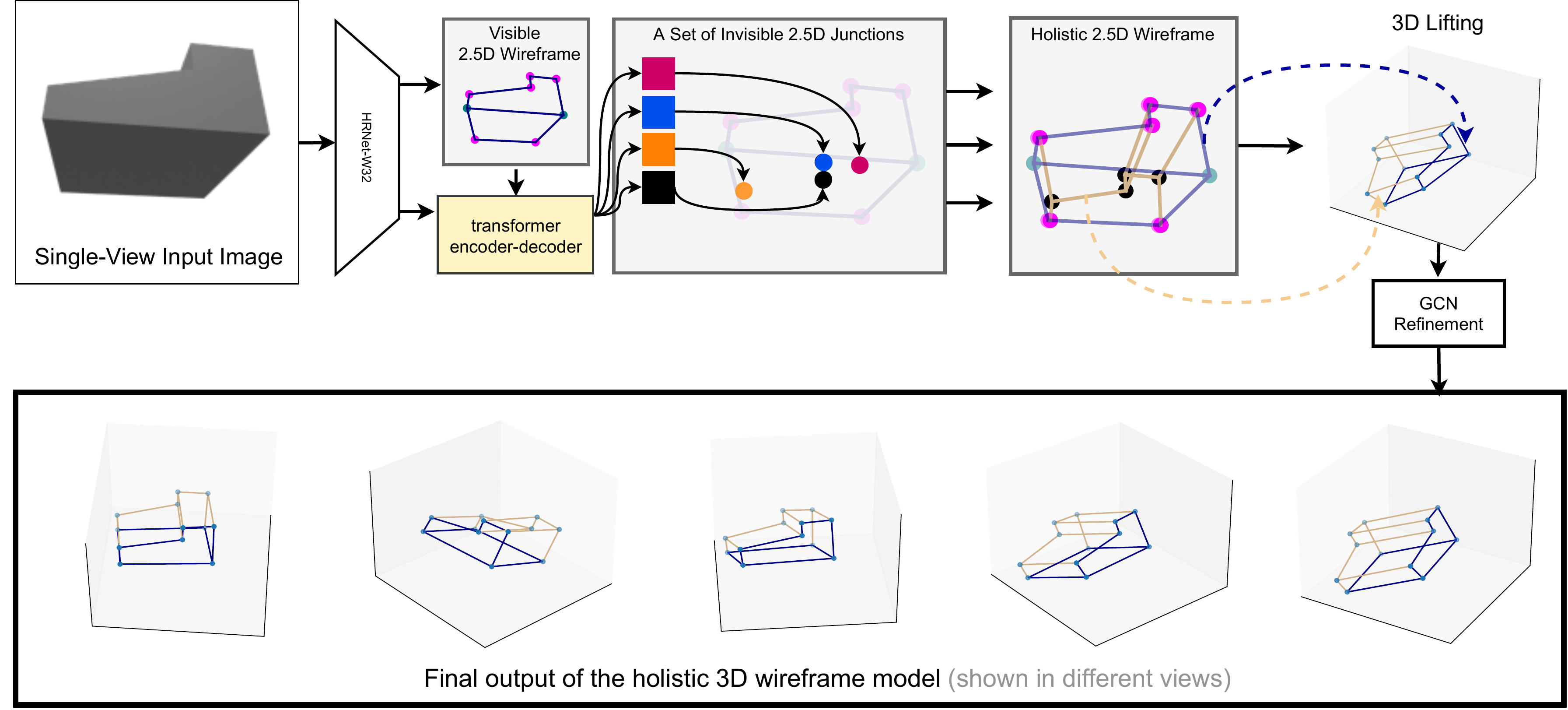}
  \caption{The algorithmic flow of our proposed Deep Spatial Gestal model.}
  \label{fig:model-architecture}
  \vspace{-5mm}
\end{figure*}

In this section, we present the detail of the proposed Deep Spatial Gestalt model to mimic the human brain at filling in the hidden junctions and lines in an image and creating a whole 3D wireframe model than the sum of the visible 2D wireframe. 

\subsection{Network Architecture}
As shown in Fig.~\ref{fig:model-architecture}, our DSG model consists of three main steps: (1) perceiving the visible 2.5D wireframe from the input image, (2) filling the hidden 2.5D junctions and lines, and (3) lifting the 2.5D wireframe into 3D Euclidean space. 

\myparagraph{Backbone Network.} We use a modified HRNet-w32~\cite{wang2020deep} as the backbone for the DSG model. The stride of the input convolution is changed from $2$ to $1$, which doubles the spatial dimension of output feature maps. The outputs of the 4-th convolutional stage are utilized as the deep feature maps for the subsequent usage. In detail, for an input image $I\in\mathbb{R}^{3\times H\times W}$, we use the high resolution feature $\mathcal{F}_{hr} \in \mathbb{R}^{32\times \frac{H}{2} \times \frac{W}{2}}$ to learn the visible junctions and line segments,
and use the low resolution feature $\mathcal{F}_{lr}\in\mathbb{R}^{256\times \frac{H}{16} \times \frac{W}{16}}$ for the hidden parts because of the larger receptive field for global information. 

\myparagraph{Perceiving Visible 2.5D Wireframe as Initialization.} We use the state-of-the-art H-AFM ~\cite{xue2020holistically} for the perception of the visible 2D wireframe of the image $I$ and the depth values of the junctions. 
In detail, we learn the holistic attraction field map (HAFM) to generate the line proposals and use 2 convolution layers to predict the junction heatmaps and the depth maps on the top of the feature map $\mathcal{F}_{hr}$ to generate the proposals of 2.5D junctions and line segments. 
To remove the false positive line proposals, we use the LOIPooling~\cite{zhou2019end} layer to learn the confidence for each generated line segment. The line segments that are with confidences smaller than $0.5$ are discarded for the subsequent processes. As our end task requires inferring the hidden line segments, we classify the observable junctions into two categories, visible and fleeting, according to the graph association in the 3D wireframe model. The observable junctions that have one or more connections to the hidden line segment are regarded as fleeting junctions. We call the non-fleeting observable junctions the visible ones.

We denote the set of visible 2D junctions and the fleeting ones by $\mathcal{J}_v^{2d} \subset \mathbb{R}^2$ and $\mathcal{J}_f^{2d} \subset \mathbb{R}^2$, respectively. 
The Z values for the visible and fleeting junctions are denoted by $\mathcal{Z}_{v} = \{Z_1^v, \ldots, Z_{N_{v}}^v\}$ and $\mathcal{Z}_{f} = \{Z_1^f, \ldots, Z_{N_{f}}^f\}$.
The learned line segments of the 2D wireframe are represented by the index set $\mathcal{L}_o  = (m,n), m,n \in \{0,...,N\}$, where $N = |\mathcal{J}_v^{2d}| + |\mathcal{J}_f^{2d}|$. Finally, we yield the observable 2D wireframe graph by $\mathcal{W}_o = (\mathcal{J}_o, \mathcal{L}_o)$, where the observable junction set $\mathcal{J}_o$ is the union of $\mathcal{J}_v$ and $\mathcal{J}_f$. 

\myparagraph{Filling Invisible Parts with Transformers.} As the hidden junctions are occluded by the front-view surfaces of the 3D object, it is challenging to use the convolution layers with limited receptive fields to fill the blank. 
To this end, we resort to using a transformer encode-decoder architecture to decode the hidden junctions via self-attention mechanism~\cite{carion2020end,tan2021planetr}. We term the headnet of learning hidden 2.5D junctions as \emph{HiddenTR}.

Our \emph{HiddenTR} consists of two encoder-decoder transformers to capture the contextual information presented in the image features and the geometric information presented in the observable 2D wireframe. To keep the simplicity, these two transformers follow the standard architecture design with $6$ encoders and decoders as in \cite{carion2020end}.
We term the contextual transformer and the geometric transformer as \emph{HiddenTR-C} and \emph{HiddenTR-G}.

In the encoding stage of \emph{HiddenTR}, the encoder branch of \emph{HiddenTR-C} takes the feature map $\mathcal{F}_{lr}$ as a sequence with a length of $S_{ctx} = H/16\times W/16$, and uses a MLP layer to transform the feature in the 256-dim hidden space. Then, the feature with positional embedding passes all encoding layers step-by-step to output a sequence of 256-dim output features. In a parallel line, the encoder branch of \emph{HiddenTR-G} takes the LOI features with the learned positional embeddings of the $M$ detected line segments as a sequence to embed the geometric features. 

In the decoding stage of \emph{HiddenTR}, two parallel decoders share the same learnable queries $Q_j\in\mathbb{R}^{d\times T}$ to yield $T$ hidden features and then output $T$ hidden 2.5D junctions (2D coordinates on the image plane and the z value) and the confidence score of junctions, denoted by the sets of $\mathcal{J}_h^{2d} = \{(x_1^h, y_1^h), \ldots, (x_T^h, y_T^h) \}\subset \mathbb{R}^2$, $\mathcal{Z}_{h} = \{z_1^h, \ldots, z_{T}^h\}$ and $\mathcal{C} = \{c_1, \ldots, c_{T}\}$, respectively.

Once we have the hidden junctions, the line proposals can be obtained by taking the predicted 2D hidden junctions $\mathcal{J}_{h}^{2d}$ and the fleeting junctions $\mathcal{J}_f^{2d}$ together to enumerate the all-pair proposals. Then we use the LOI pooling~\cite{zhou2019end} to get the line features on the low-resolution feature map $\mathcal{F}_{lr}$. During training, we dynamically sample the top $M = \min(T, 2N_{hj}^{gt})$ to yield hidden line proposals, where $N_{hj}^{gt}$ is the number of GT hidden junctions for the input. In the testing, any hidden junction whose score $<0.5$ is discarded.
The line verification network is a 2-layer MLP to classify the proposals into positive and negative categories with scores.

\myparagraph{Holistic 3D Wireframe Refinement.} With the learned 2.5D junctions $(\mathcal{J}_v^{2d}, \mathcal{Z}_v)$, $(\mathcal{J}_f^{2d}, \mathcal{Z}_f)$, $(\mathcal{J}_h^{2d}, \mathcal{Z}_h)$, it is straightforward to lift a holistic 3D wireframe model by using the learned visible and hidden line segments with the intrinsic matrix $K$.  As the simply 3D lifting operation remains an issue of inaccuracy, we propose a refinement module to adjust the 3D junction locations by using the global structure regularity.
Denoted by $\mathcal{J}^{3d}$ for all the 3D junctions (including the hidden and visible ones) and $\mathcal{L} = (m,n), m,n \in \{0, \ldots \|\mathcal{J}^{3d}\|\} $ for all the line indices, we translate them into a binary adjencent matrix $\mathcal{A} = \{0,1\}^{\|\mathcal{J}^{3d}\| \times \|\mathcal{J}^{3d}\|}$, where $\mathcal{A}(m,n) = 1, if (m, n) \in \mathcal{L}$. We leverage a G-ResNet~\cite{pixel2mesh} $G(\cdot)$ to learn and predict offset vectors $\Delta\mathbf{J}$ from each predicted 3D junction $J \in \mathcal{J}^{3d}$ to its corresponding ground truth by
$
  \Delta\mathbf{J}^{3d} = G(\mathcal{J}^{3d},\mathcal{A}).
$
Then we update the predicted 3D junctions by $\mathcal{J}^{3d} = \mathcal{J}^{3d} + \Delta\mathbf{J}$
as the final 3D junctions to result in the expected holistic 3D wireframe model. An illustrative comparison between the lifted initial 3D wireframe model and the refined 3D wireframe model is shown in the right of Fig.~\ref{fig:model-architecture}.

\subsection{Loss Functions and Network Training}
We train our DSG model in a supervised manner on the proposed ABC-HoW dataset. The loss functions of the four modules in our DSG model are described as follows.

For the learning of 2.5D visible wireframe, we used the loss functions $\mathbb{L}_{\text{hafm}}$, $\mathbb{L}_{\text{junc}}$ and $\mathbb{L}_{\text{ver}}$ to learn the attraction fields of line segments, the junction heatmaps, the labels of line proposals, as done in~\cite{xue2020holistically}. In addition to these loss functions, we learn the depth for the observable junctions by using the $\ell_1$ loss function, denoted by $\mathbb{L}_{depth}$. The total loss function of learning 2.5D visible wireframe is
$
    \mathbb{L}_{V} = \mathbb{L}_{\text{hafm}} + \mathbb{L}_{\text{junc}} + \mathbb{L}_{\text{ver}} + \mathbb{L}_{\text{depth}}.
$

For the learning of 2.5D hidden junctions, the \emph{HiddenTR} headnet predicts a fixed number of $T=30$  junctions with the probabilities (or scores), denoted by  $\{(x_i,y_i, z_i,c_i\}_{i=1}^T$. The number $T$ is greatly larger than the ground truth junction set $\{\hat{x}_i,\hat{y}_i,\hat{z}_i, \hat{c}_i\}_{i=1}^M$ in our dataset. Here, $\hat{c}_i \in \{0,1\}$ denotes whether the junction $\mathbf{X}_i$ is a hidden junction in the image. By searching the minimal-cost bipartite matching $\sigma$ between the sets of prediction and ground truth, we compute the total loss function $\mathbb{L}_{\text{HiddenTR}}$ for \emph{HiddenTR} headnet as 
\begin{equation}
\begin{split}
  \mathbb{L}_{\text{HiddenTR}} = \sum_{i=1}^T -\log_{\hat{c}_i}(c_{\sigma_i}) + \lambda_1 \|(\hat{x}_i, \hat{y}_i) - \\ (x_{\sigma_i}, y_{\sigma_i})\|_1 + \lambda_2 \|\hat{z}_i - z_{\sigma_i}\|_1.
\end{split}
\end{equation}
where $\lambda_1$ and $\lambda_2$ are two hyper-parameter to balance the different loss terms. We set $\lambda_1=5.0$ and $\lambda_2=0.05$ in our experiment.

For the G-ResNet for the 3D wireframe refinement, we also use the $L_1$ loss between the predicted offsets and the ground truth.

\myparagraph{Training Schedule.} As our DSG model has four sequential modules with the generated proposals of line segments and junctions, we train the network in a 4-stage fashion to ease the training. 
In the first stage, we train the 2.5D visible wireframe parsing module. Then, we gradually add the remaining 3 modules one by one in the new training stage, which jointly optimizes the network parameters. In each stage, we train the (partially-added) network in 30 epochs with an initial learning rate of 0.0001. After the training of 15 epochs, the learning rate is decayed by $0.5$. In total, our DSG model is trained with $120$ epochs with the periodic learning rate schedule. During training, we set the batch size to 16 and use 1 GPU card (Nvidia Titan RTX) for the acceleration. The Adam optimizer is used for the training.

\section{Experiments}
This section demonstrates that the HoW-3D task can be solved with single-view input images. To quantitatively evaluate the perceived holistic 3D wireframe model by our DSG model, we set two strong baseline methods to make comparisons. Our data and code are publicly available at \url{https://github.com/Wenchao-M/HoW-3D}.
\subsection{Evaluation metrics and Baseline Configuration.}
\myparagraph{Evaluation Metrics.} We use the vectorized junction AP and structural average precision(sAP) to evaluate the accuracy of our predicted junctions and lines both in 2D and 3D space. For the threshold of evaluating 3D line segments, we add a more strict threshold of 0.01 besides the default thresholds 0.03, 0.05, and 0.07 that are used in PC2WF~\cite{liu2021pc2wf} for evaluation under the same scale setting of 3D models. The thresholds of 2D junctions and line segments are kept the same as HAWP~\cite{xue2020holistically} and L-CNN~\cite{zhou2019end}.
\begin{figure}
    \centering
    \subfigure[{\scriptsize Input}]{
     \begin{minipage}[b]{0.18\linewidth}
      \includegraphics[trim={10 20 10 30},clip, width=1\linewidth]{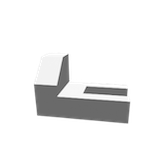}
      \includegraphics[trim={10 20 10 30},clip, width=1\linewidth]{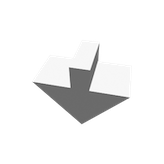}
      \includegraphics[trim={10 20 10 30},clip, width=1\linewidth]{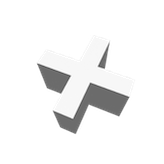}
     \end{minipage}}
     \subfigure[{\scriptsize View1}]{
     \begin{minipage}[b]{0.18\linewidth}
      \includegraphics[trim={10 20 10 30},clip, width=1\linewidth]{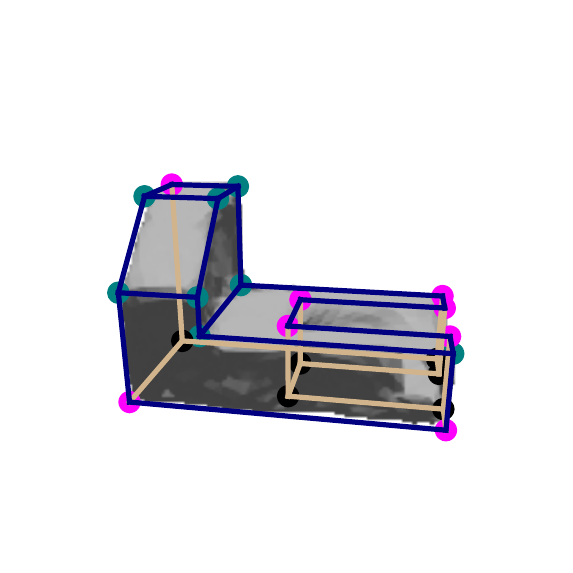}
      \includegraphics[trim={10 20 10 30},clip, width=1\linewidth]{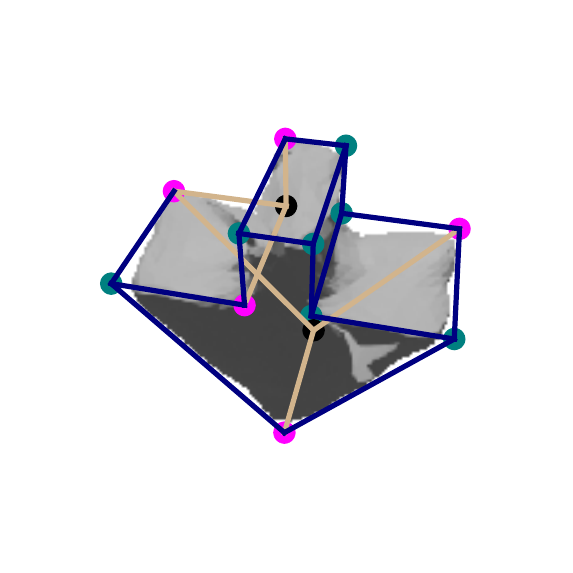}
      \includegraphics[trim={10 20 10 30},clip, width=1\linewidth]{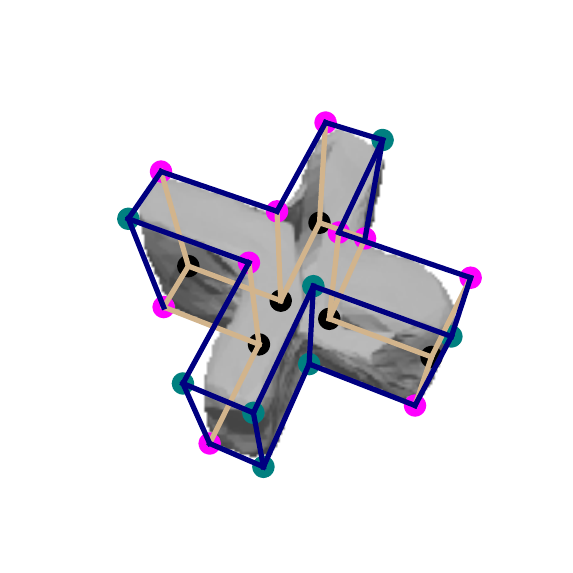}
     \end{minipage}}
     \subfigure[{\scriptsize View2}]{
     \begin{minipage}[b]{0.18\linewidth}
      \includegraphics[trim={10 20 10 30},clip, width=1\linewidth]{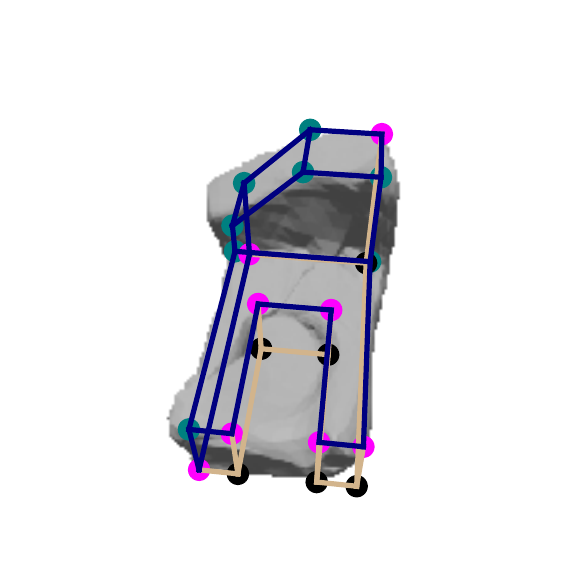}
      \includegraphics[trim={10 20 10 30},clip, width=1\linewidth]{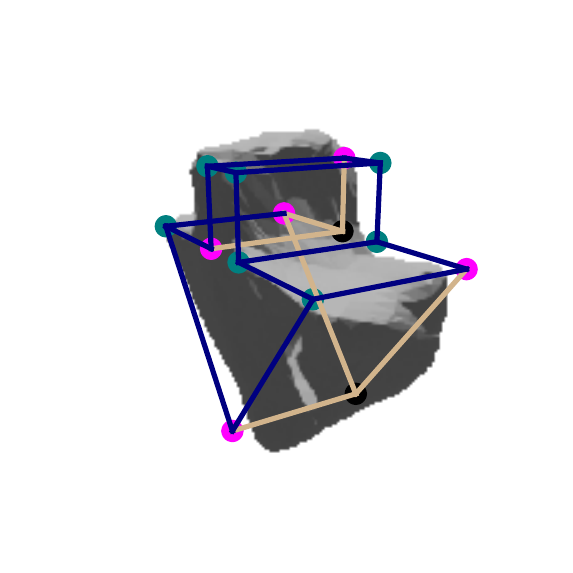}
      \includegraphics[trim={10 20 10 30},clip, width=1\linewidth]{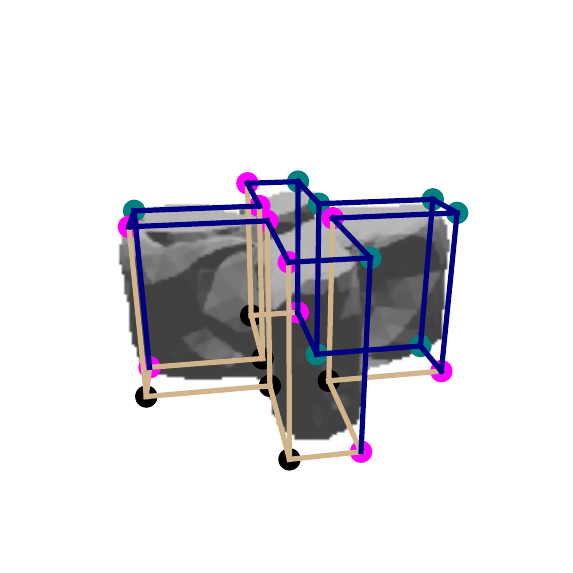}
     \end{minipage}}
     \subfigure[{\scriptsize View3}]{
     \begin{minipage}[b]{0.18\linewidth}
      \includegraphics[trim={10 20 10 30},clip, width=1\linewidth]{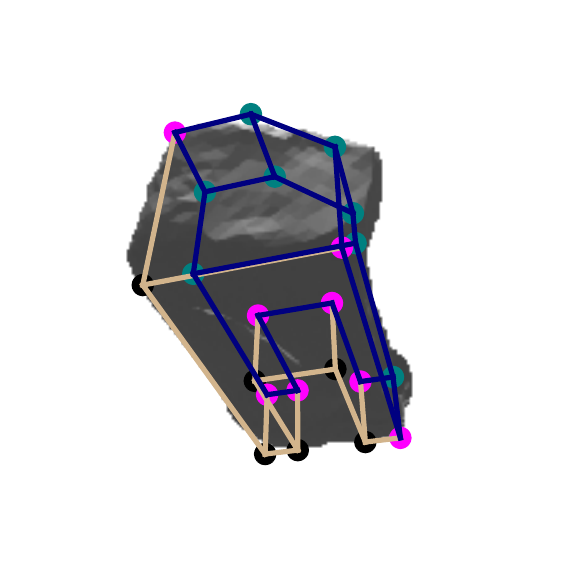}
      \includegraphics[trim={10 20 10 30},clip, width=1\linewidth]{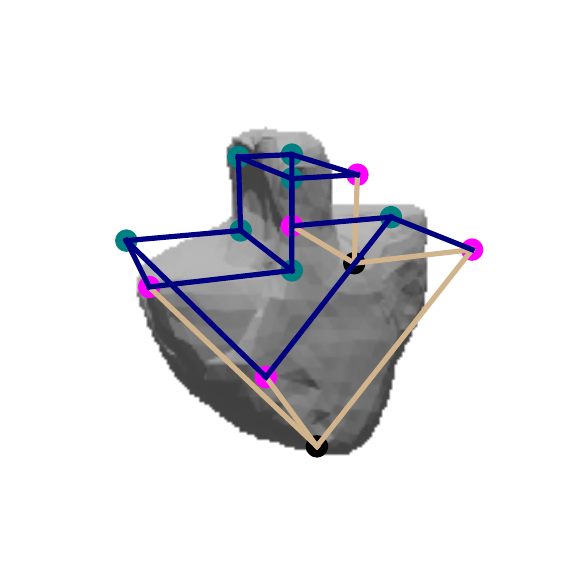}
      \includegraphics[trim={10 20 10 30},clip, width=1\linewidth]{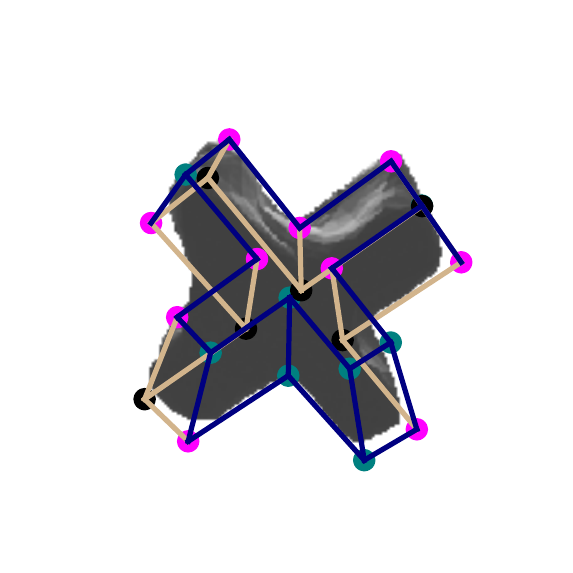}
     \end{minipage}}
     \subfigure[{\scriptsize View4}]{
     \begin{minipage}[b]{0.18\linewidth}
      \includegraphics[trim={10 20 10 30},clip, width=1\linewidth]{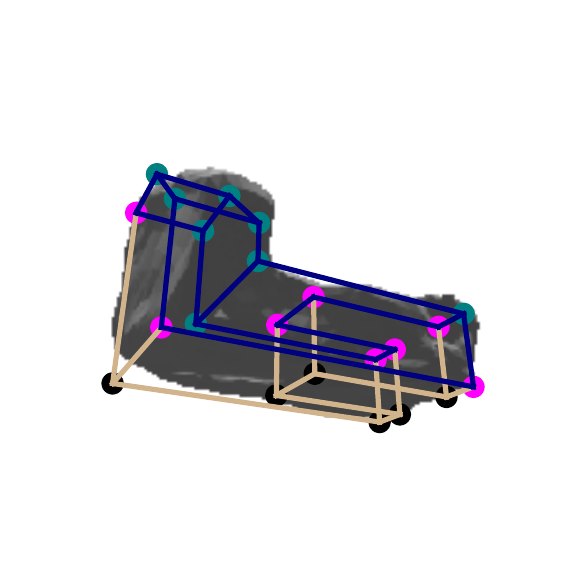}
      \includegraphics[trim={10 20 10 30},clip, width=1\linewidth]{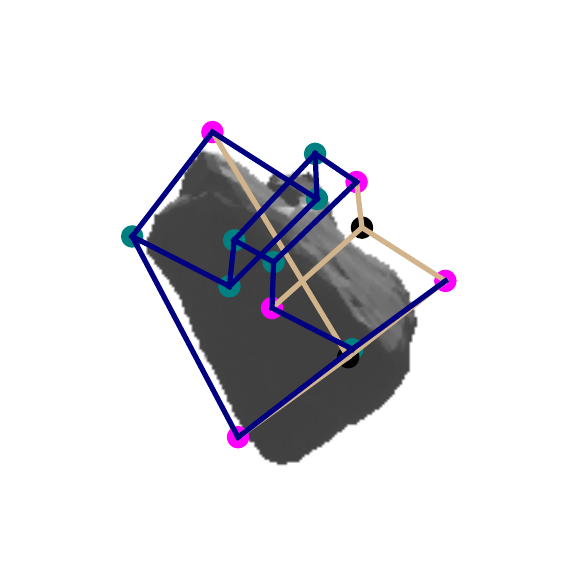}
      \includegraphics[trim={10 20 10 30},clip, width=1\linewidth]{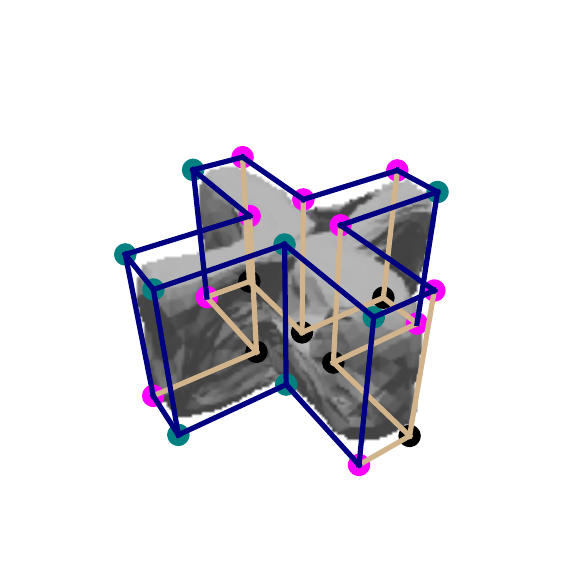}
     \end{minipage}}
    \caption{The results of Pixel2mesh and the corresponding \textbf{GT wireframes}.}
    \label{fig:p2m}
    \vspace{-3mm}
\end{figure}

\myparagraph{Baseline Configuration.} As the existing single-view based 3D wireframe perception approach only focuses on the visible wireframe reconstruction and requires structures under the Manhattan assumption. We consider the following two possible baseline configurations: (1) Directly combine the existing single-view 3D shape reconstruction and the 3D wireframe reconstruction from point clouds. (2) Extending the existing 2D wireframe detector to the holistic one.  

\begin{table}[h]
    \vspace{-2mm}
    \centering
    \caption{The reconstruction accuracy and the junction recall of Pixel2Mesh, where $\tau=10^{-4}$ is same as in~\cite{pixel2mesh}.}
    \vspace{-2mm}
    \label{tab:p2m}
    \resizebox{0.95\linewidth}{!}{
    \begin{tabular}{c|c|c|c|c|c}
    \toprule
     Dataset & $F(\tau)$  & $F(2\tau)$ & CD & Junction recall (0.01) & Junction recall (0.03)\\
     \midrule
     Shapenet         & 59.72   &  74.19  & 0.591 & - & - \\%
     ABC-HoW          & 15.87   &  32.57  & 2.261 & 16.28 & 43.19  \\%
     \bottomrule
    \end{tabular}
    }
    \vspace{-3mm}
\end{table}

\begin{figure*}
\renewcommand{\arraystretch}{0.1}
    \centering
    \resizebox{0.8\linewidth}{!}{
    \begin{tabular}{c|cccc|cccc}
        \toprule
        Input Image & Ours & HAWP-EX & PC2WF & GT & Ours & HAWP-EX & PC2WF & GT \\
         \includegraphics[height=0.1\linewidth]{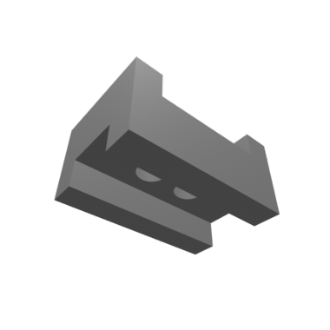}
         & \includegraphics[height=0.1\linewidth]{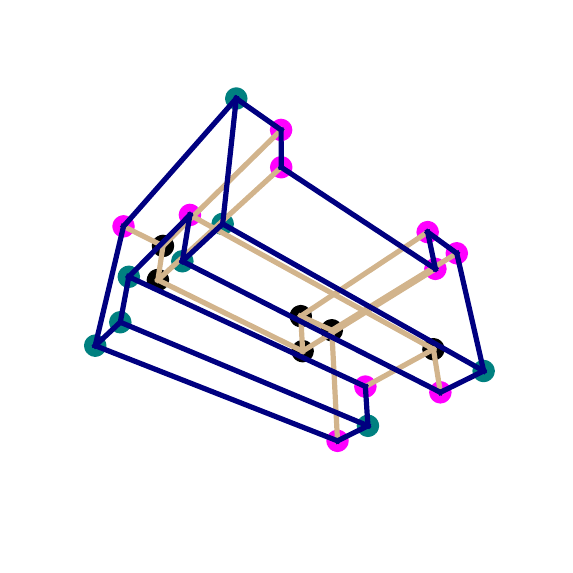}
         & \includegraphics[height=0.1\linewidth]{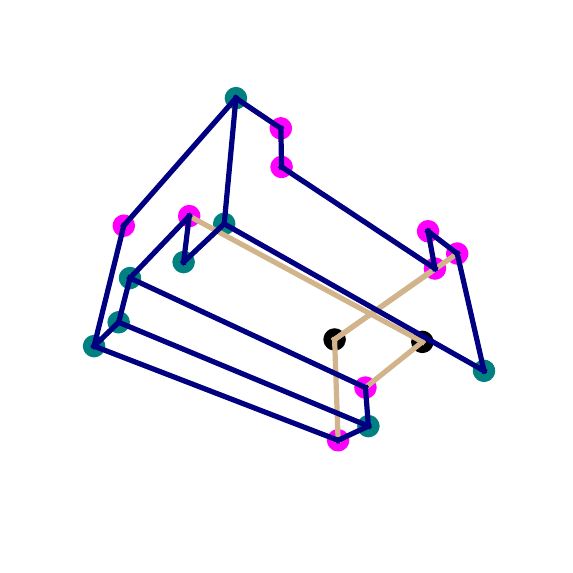}
         & \includegraphics[height=0.1\linewidth]{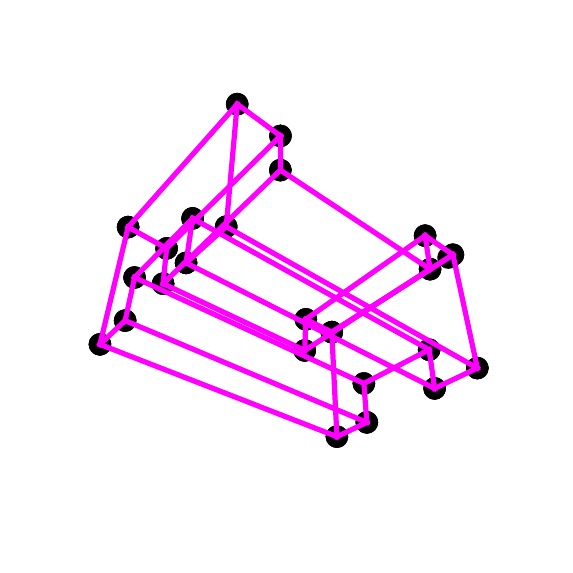}
         & \includegraphics[height=0.1\linewidth]{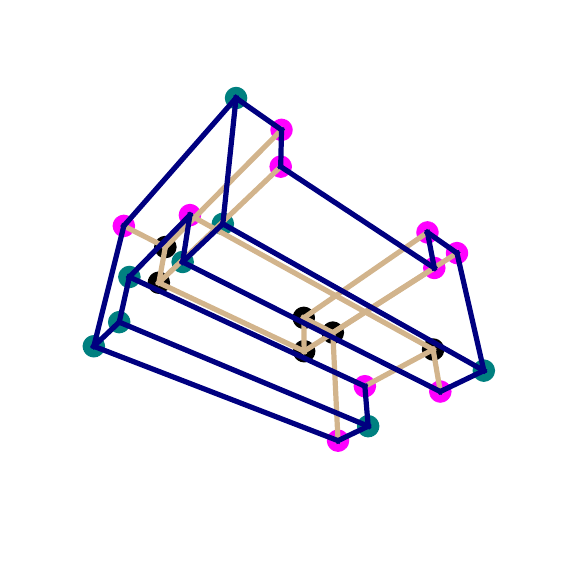}
         & \includegraphics[height=0.1\linewidth]{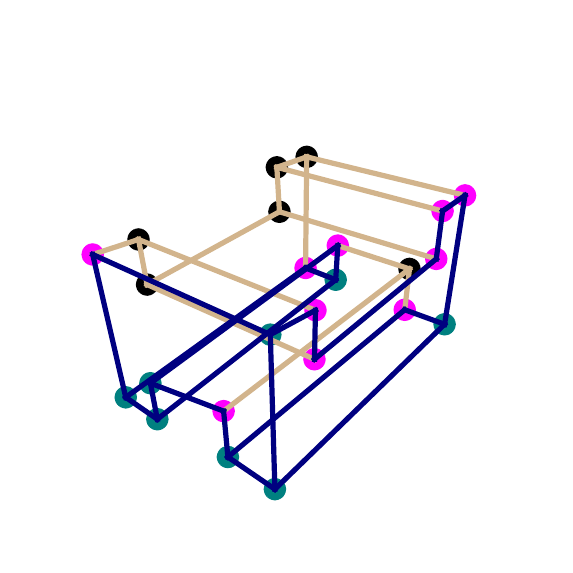}
         & \includegraphics[height=0.1\linewidth]{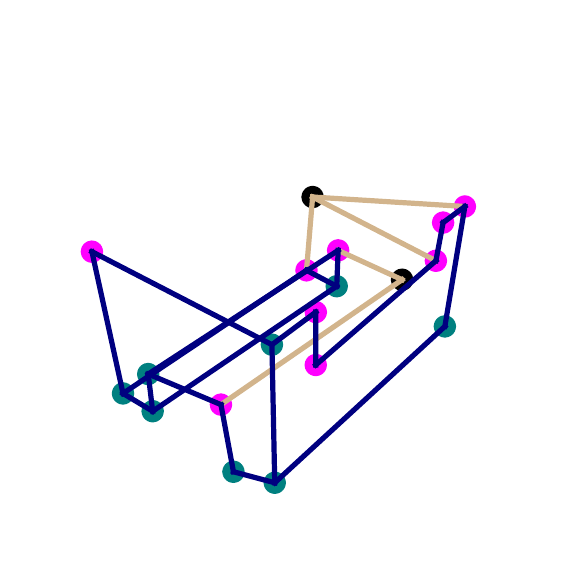}
         & \includegraphics[height=0.1\linewidth]{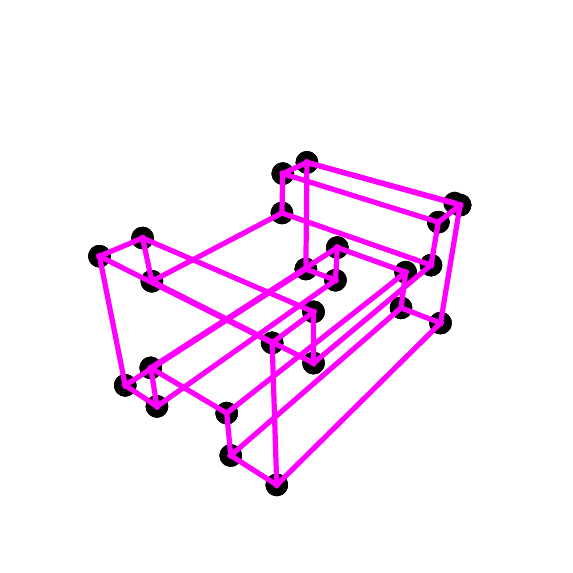}
         & \includegraphics[height=0.1\linewidth]{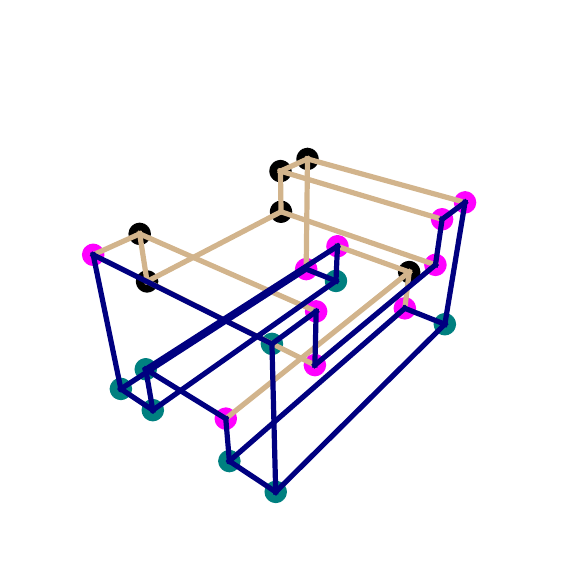}\\
          \includegraphics[height=0.1\linewidth]{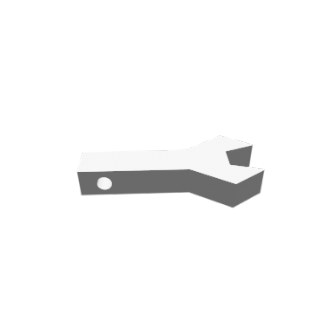}
         & \includegraphics[height=0.1\linewidth]{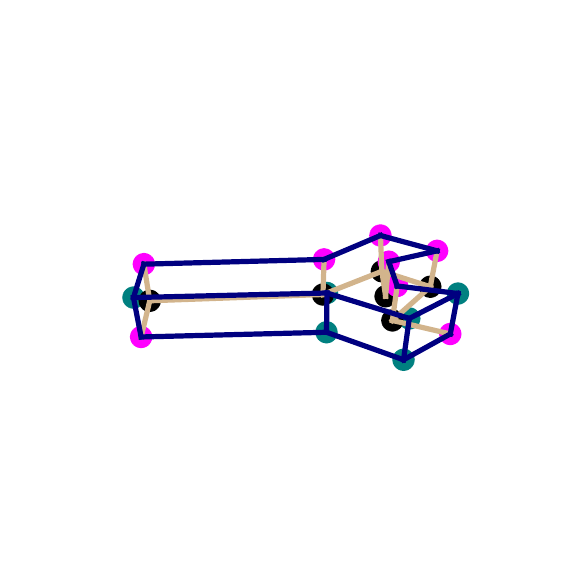}
         & \includegraphics[height=0.1\linewidth]{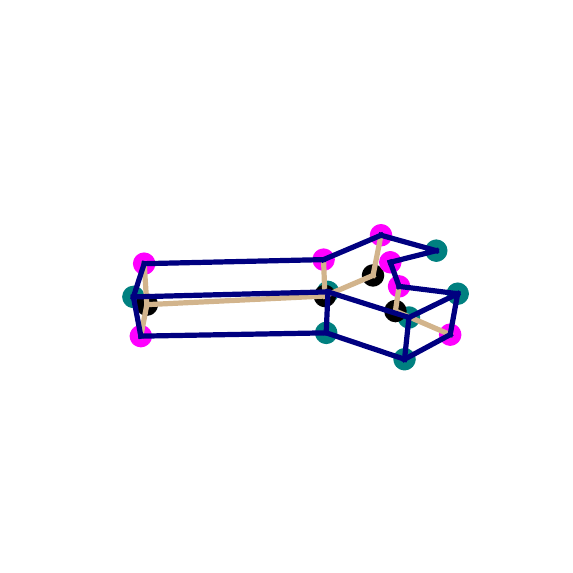}
         & \includegraphics[height=0.1\linewidth]{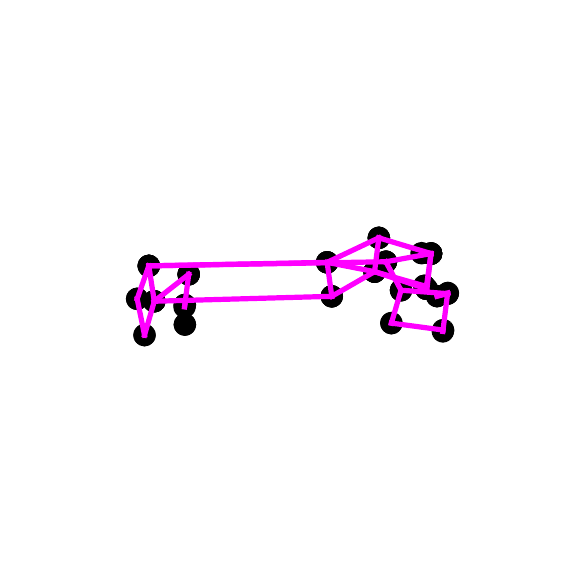}
         & \includegraphics[height=0.1\linewidth]{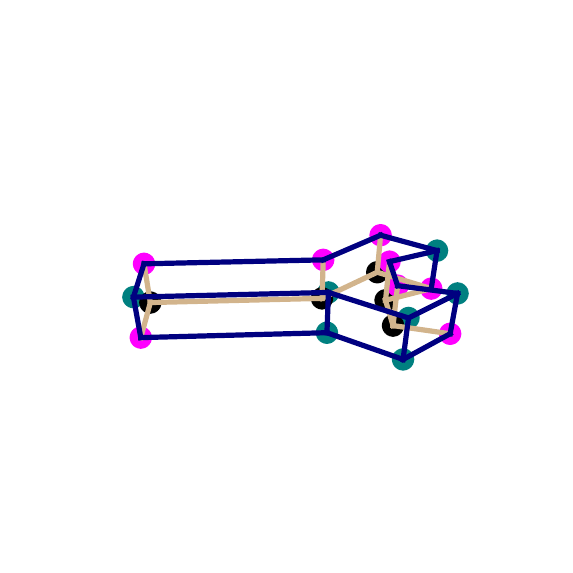}
         & \includegraphics[height=0.1\linewidth]{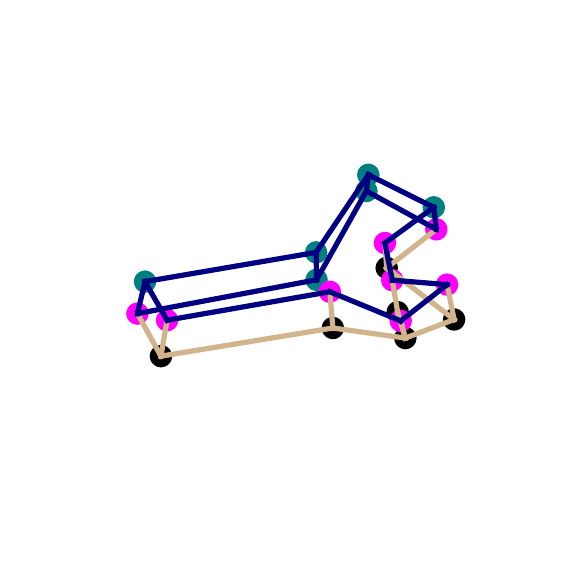}
         & \includegraphics[height=0.1\linewidth]{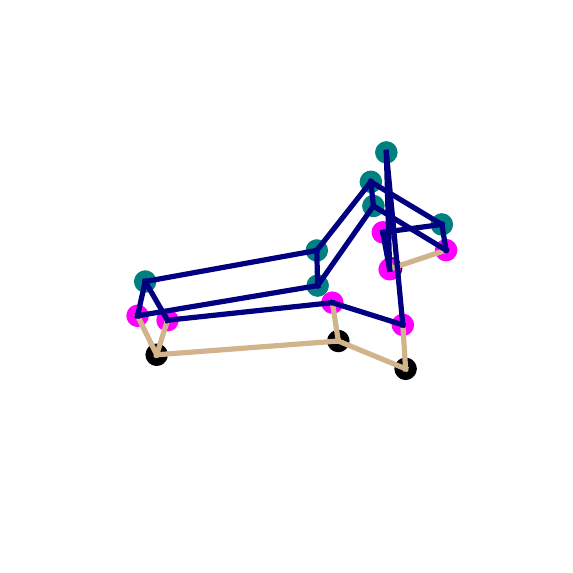}
         & \includegraphics[height=0.1\linewidth]{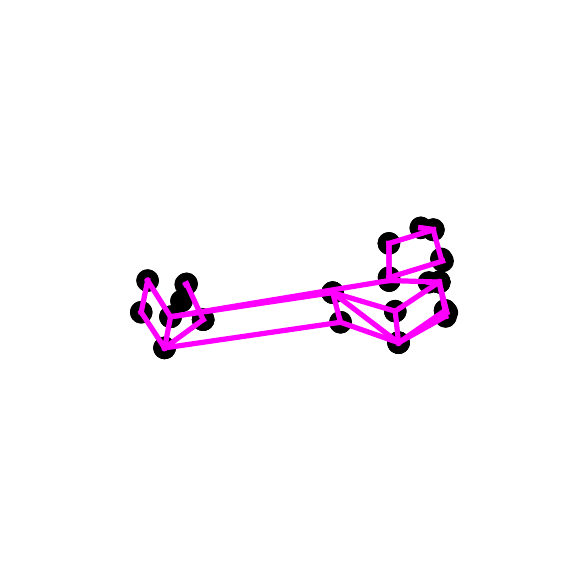}
         & \includegraphics[height=0.1\linewidth]{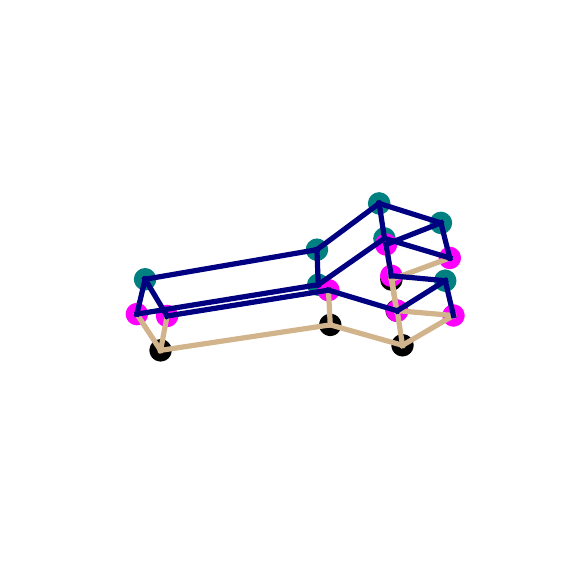}\\
         \includegraphics[height=0.1\linewidth]{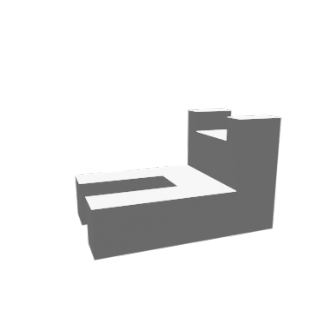}
         & \includegraphics[height=0.1\linewidth]{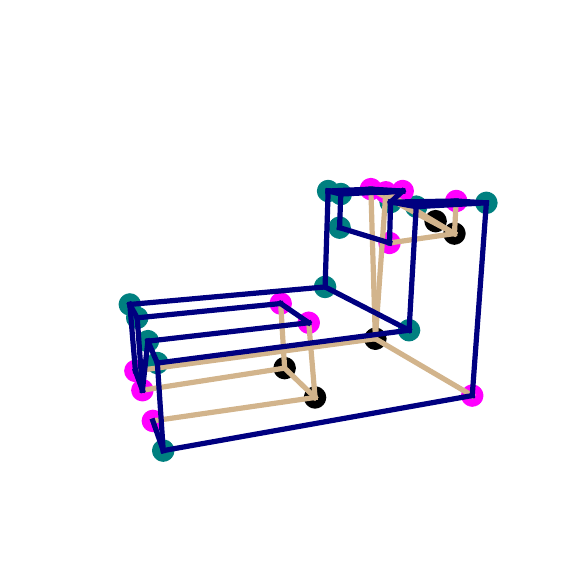}
         & \includegraphics[height=0.1\linewidth]{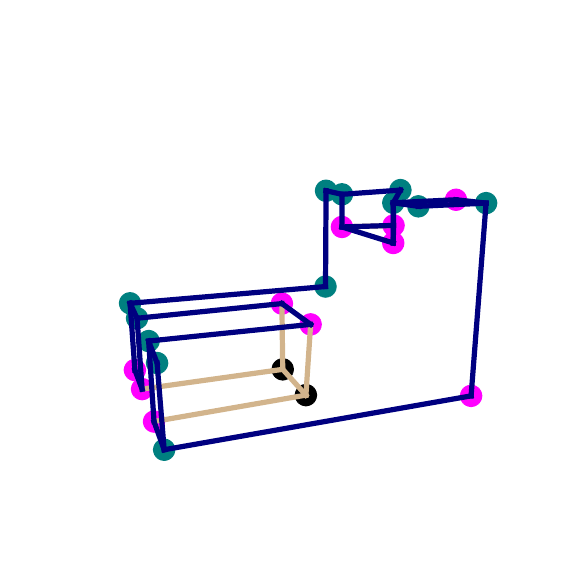}
         & \includegraphics[height=0.1\linewidth]{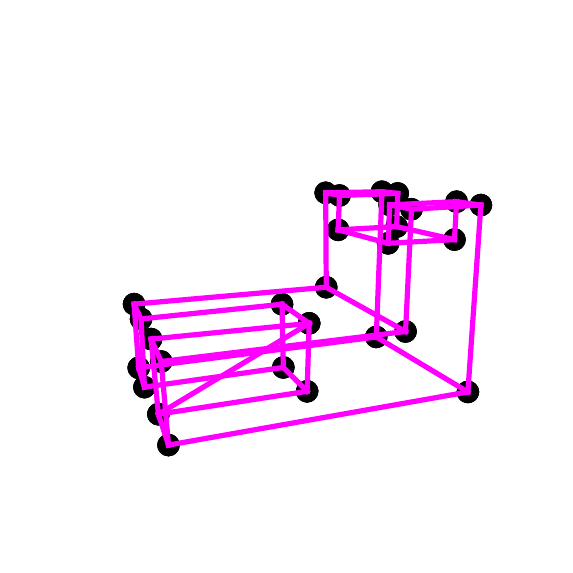}
         & \includegraphics[height=0.1\linewidth]{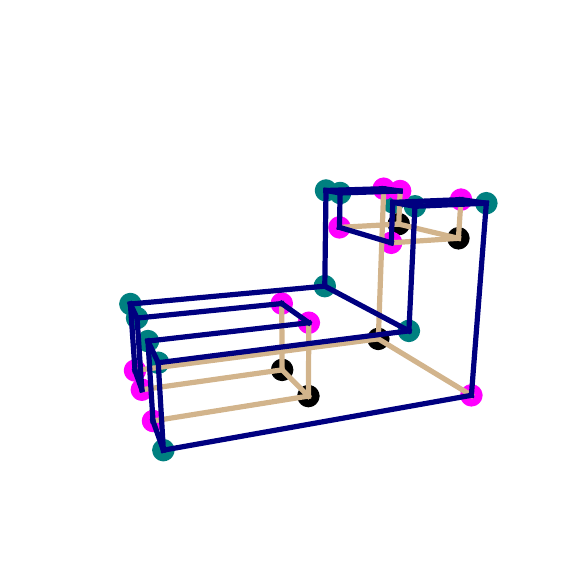}
         & \includegraphics[height=0.1\linewidth]{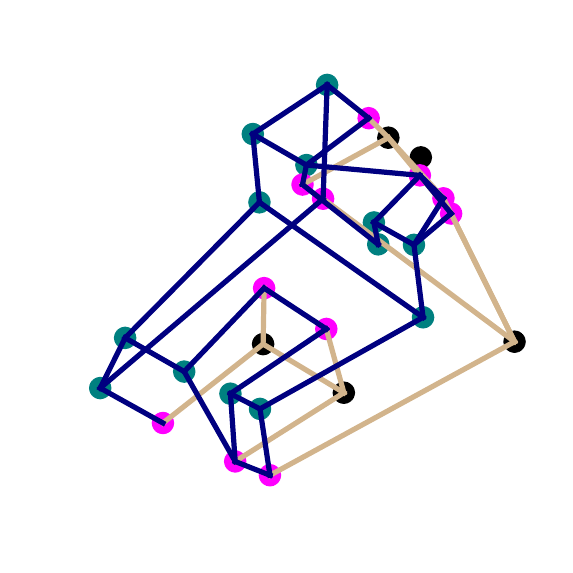}
         & \includegraphics[height=0.1\linewidth]{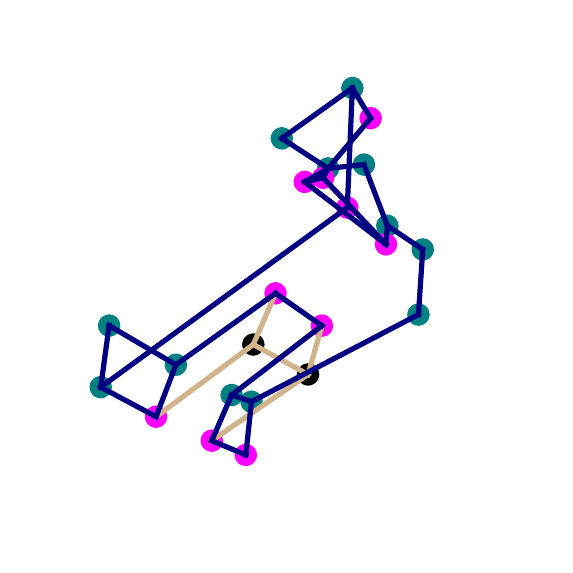}
         & \includegraphics[height=0.1\linewidth]{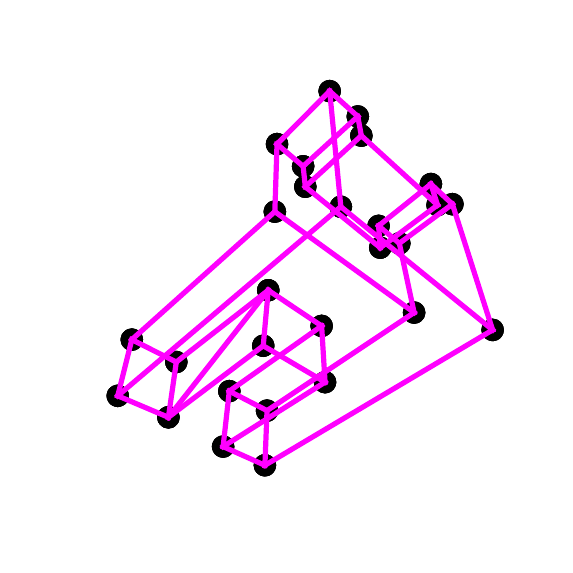}
         & \includegraphics[height=0.1\linewidth]{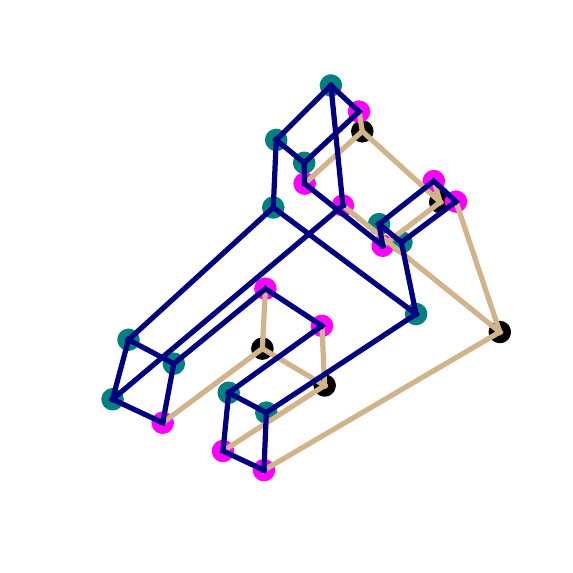}\\
         \includegraphics[height=0.1\linewidth]{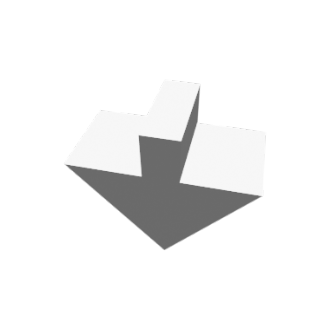}
         & \includegraphics[height=0.1\linewidth]{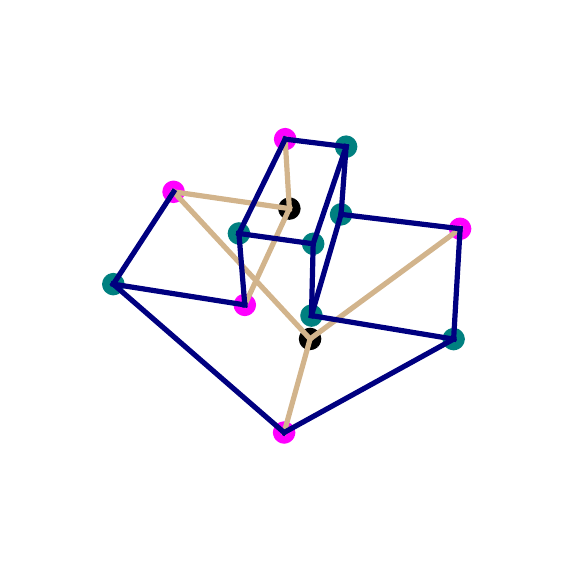}
         & \includegraphics[height=0.1\linewidth]{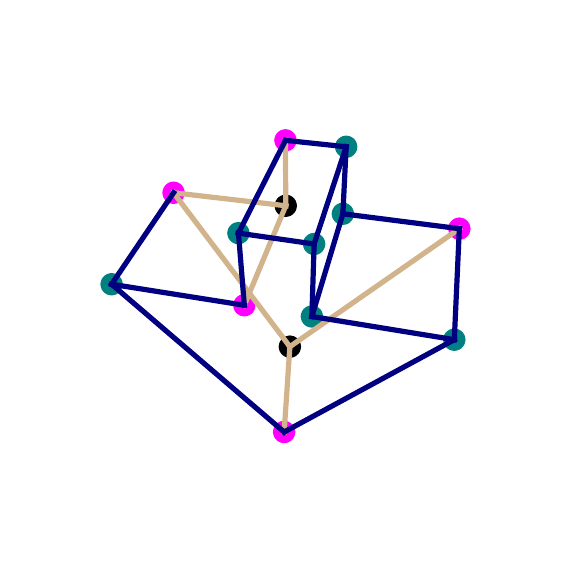}
         & \includegraphics[height=0.1\linewidth]{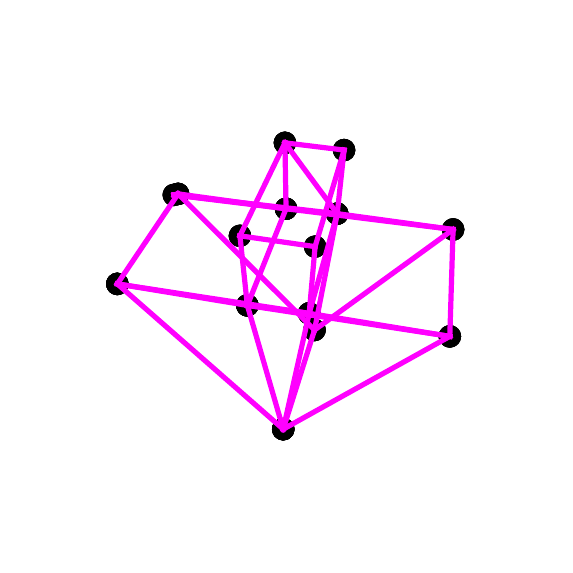}
         & \includegraphics[height=0.1\linewidth]{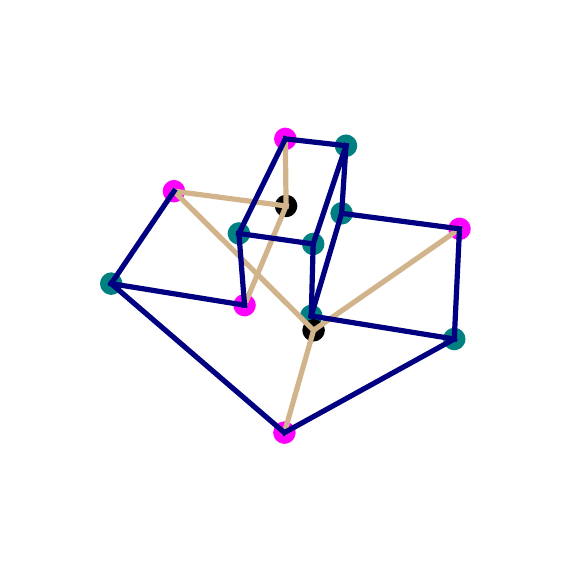}
         & \includegraphics[height=0.1\linewidth]{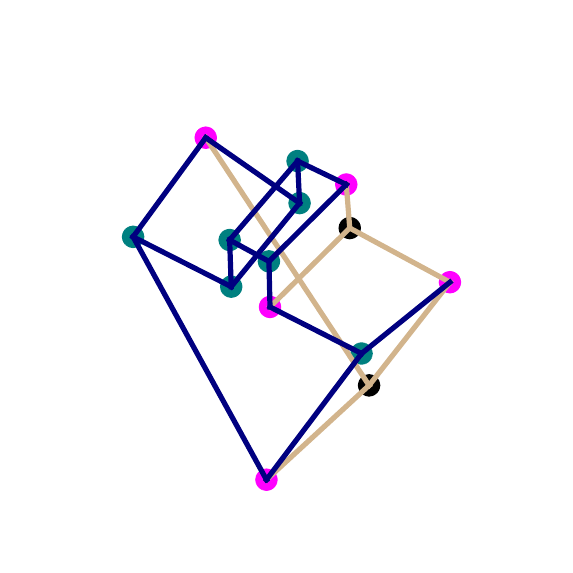}
         & \includegraphics[height=0.1\linewidth]{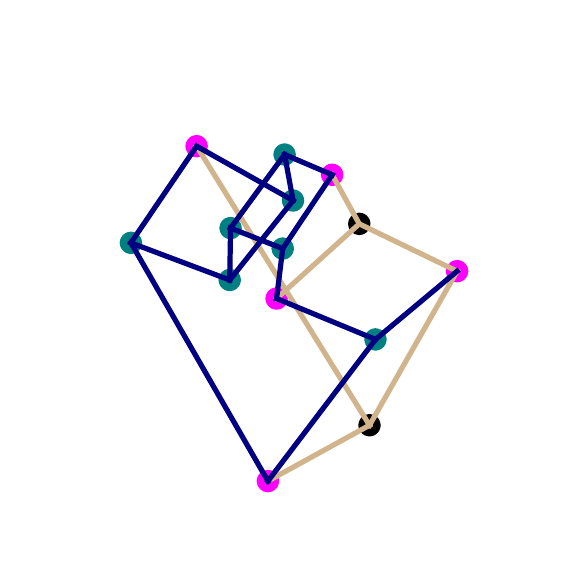}
         & \includegraphics[height=0.1\linewidth]{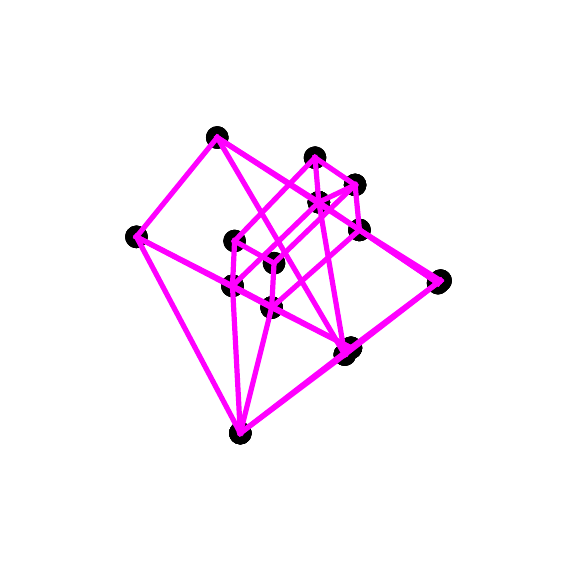}
         & \includegraphics[height=0.1\linewidth]{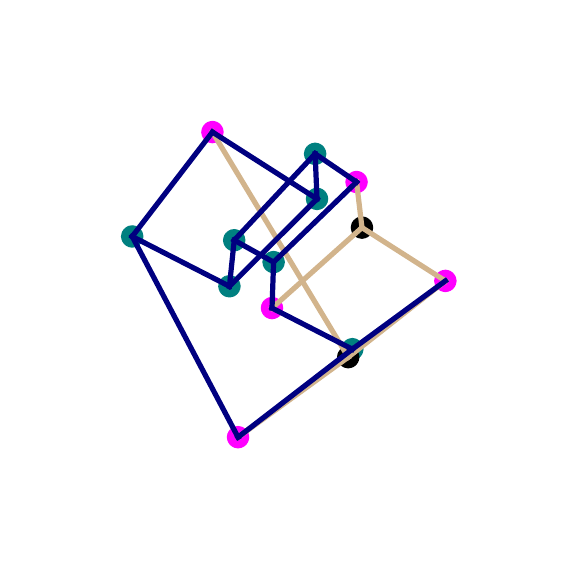}\\
           \includegraphics[height=0.1\linewidth]{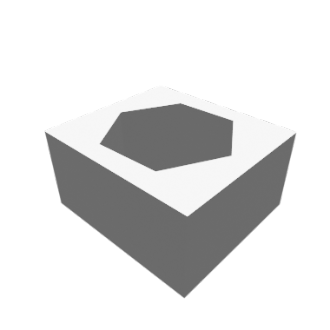}
         & \includegraphics[height=0.1\linewidth]{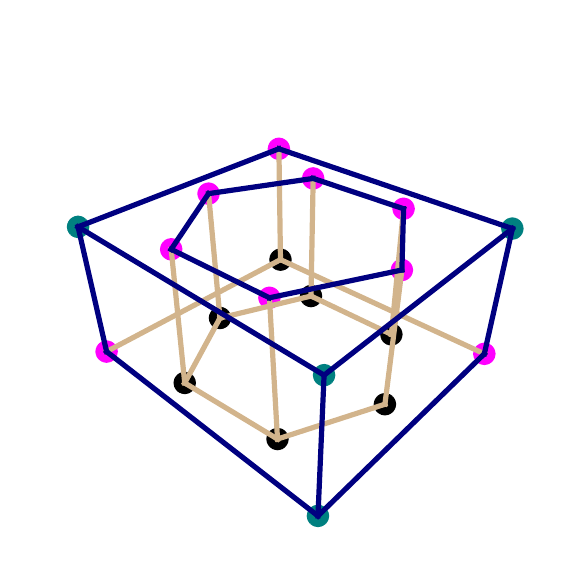}
         & \includegraphics[height=0.1\linewidth]{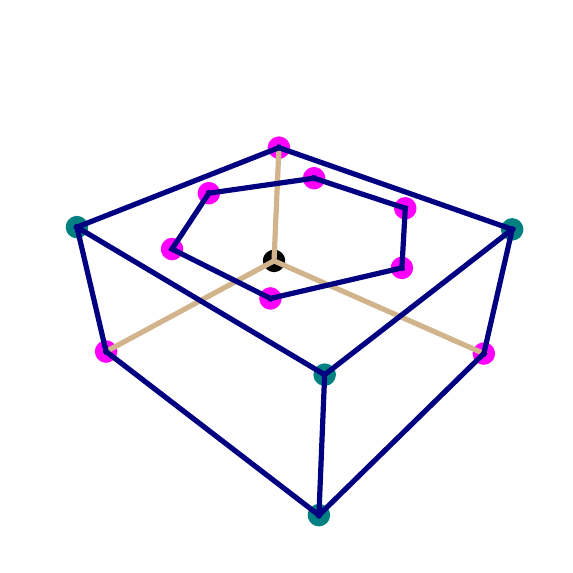}
         & \includegraphics[height=0.1\linewidth]{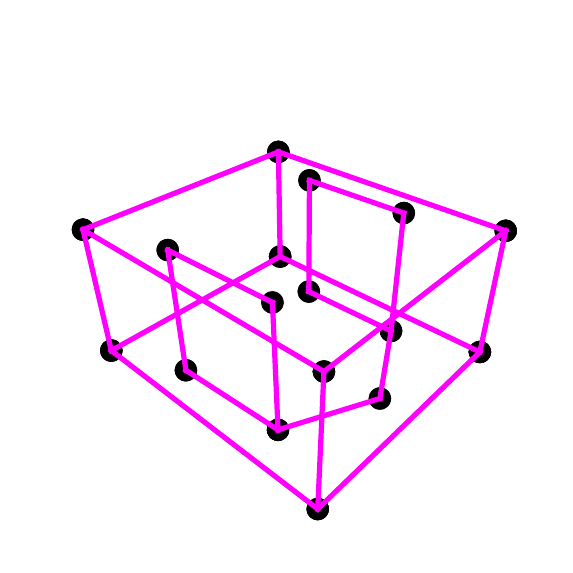}
         & \includegraphics[height=0.1\linewidth]{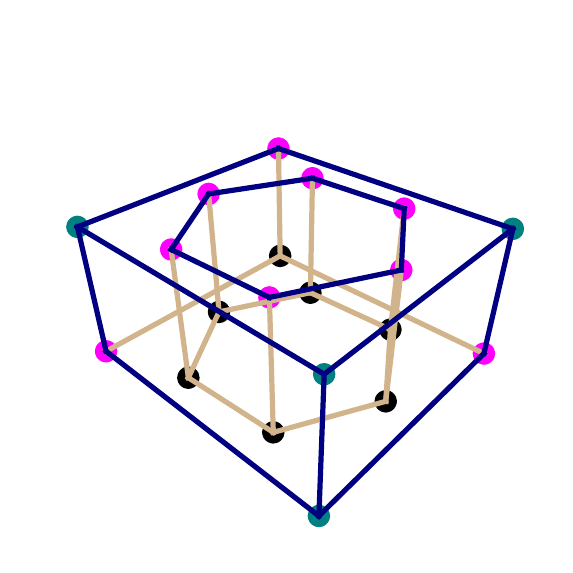}
         & \includegraphics[height=0.1\linewidth]{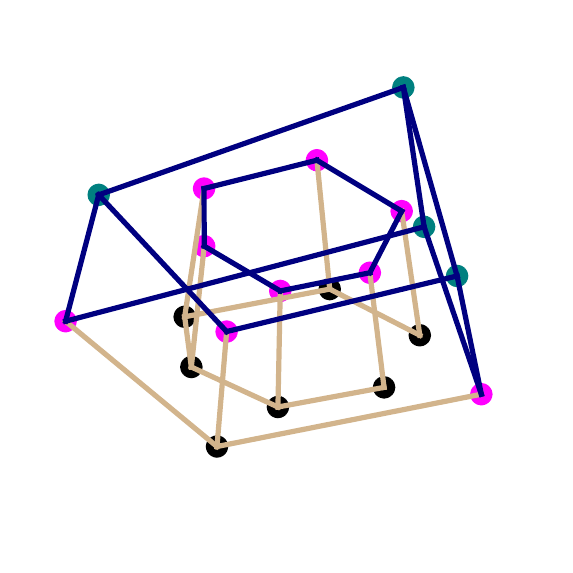}
         & \includegraphics[height=0.1\linewidth]{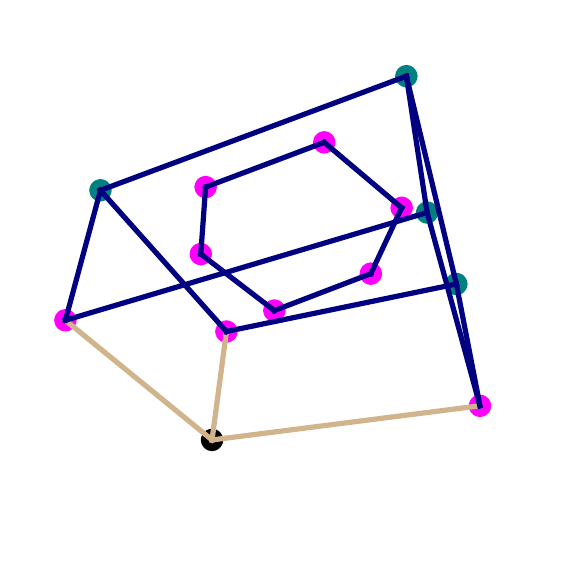}
         & \includegraphics[height=0.1\linewidth]{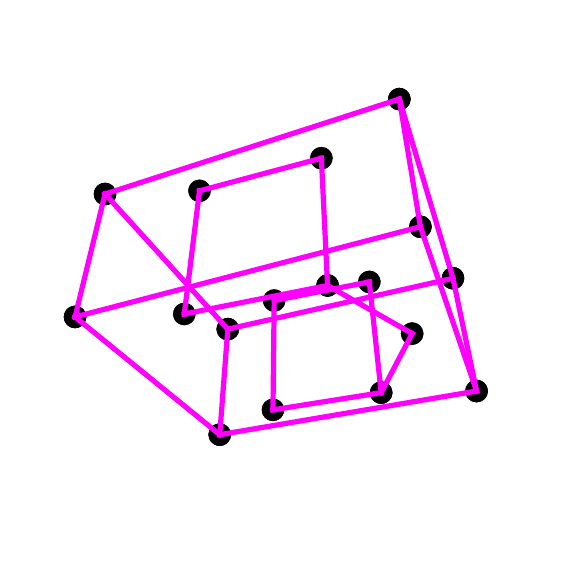}
         & \includegraphics[height=0.1\linewidth]{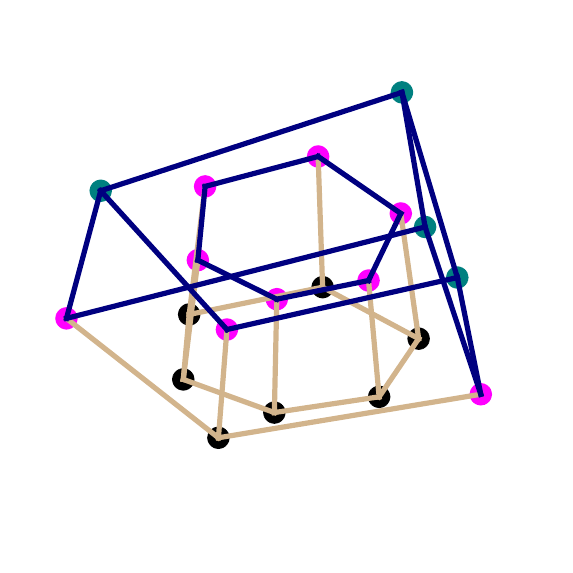}\\
         \includegraphics[height=0.1\linewidth]{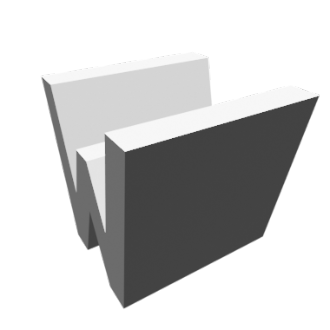}
         & \includegraphics[height=0.1\linewidth]{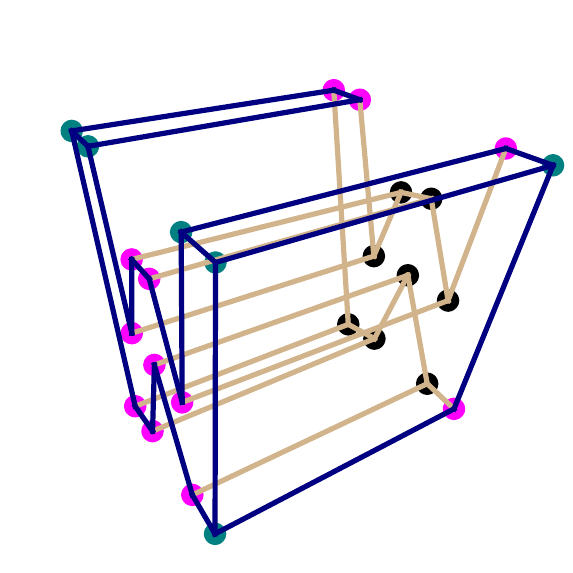}
         & \includegraphics[height=0.1\linewidth]{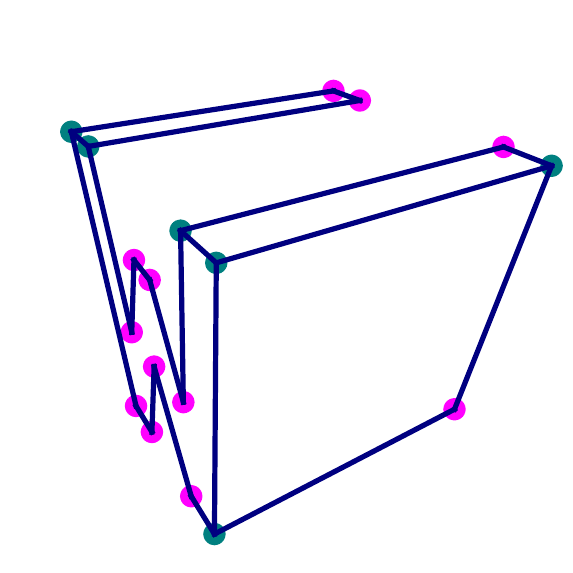}
         & \includegraphics[height=0.1\linewidth]{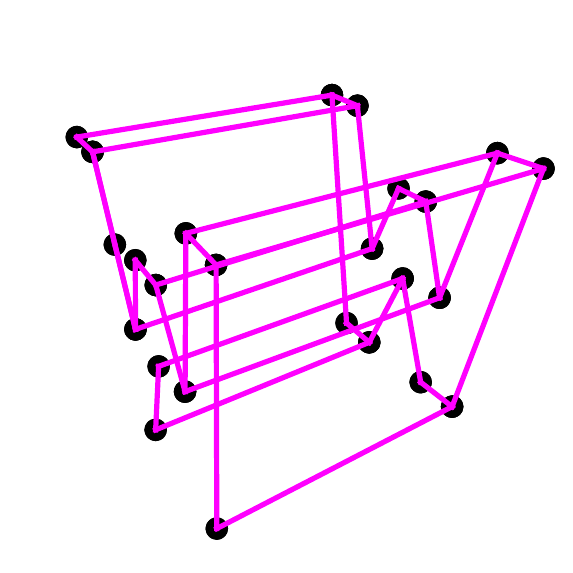}
         & \includegraphics[height=0.1\linewidth]{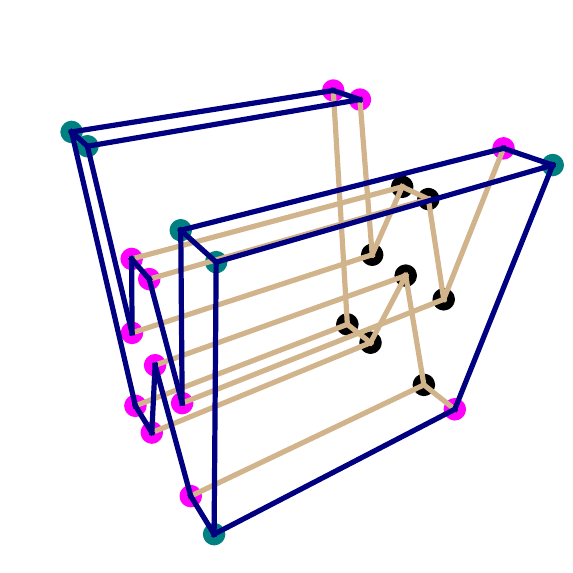}
         & \includegraphics[height=0.1\linewidth]{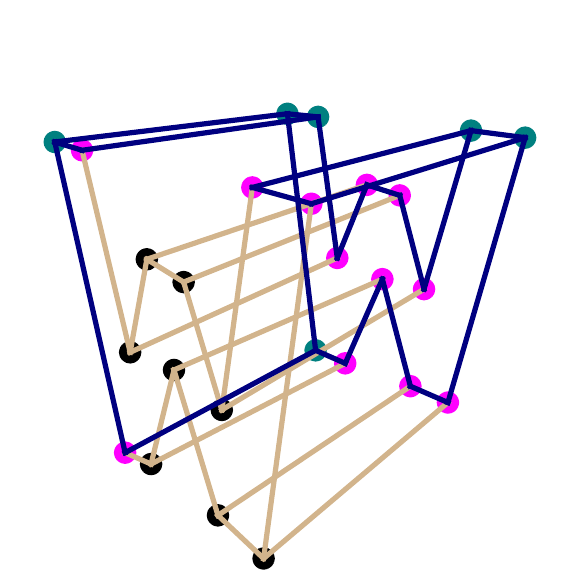}
         & \includegraphics[height=0.1\linewidth]{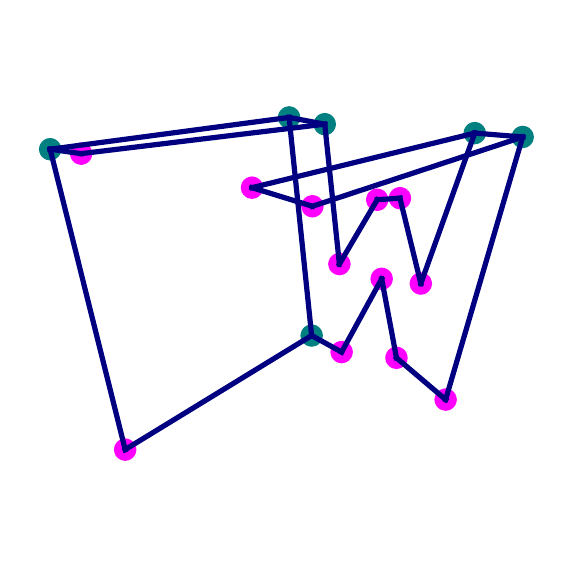}
         & \includegraphics[height=0.1\linewidth]{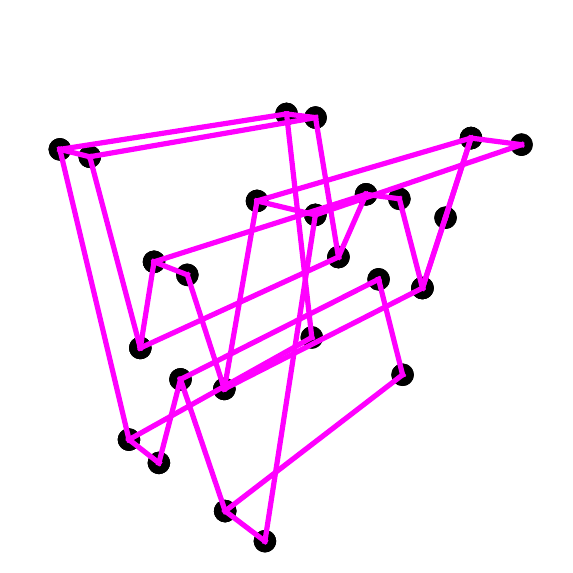}
         & \includegraphics[height=0.1\linewidth]{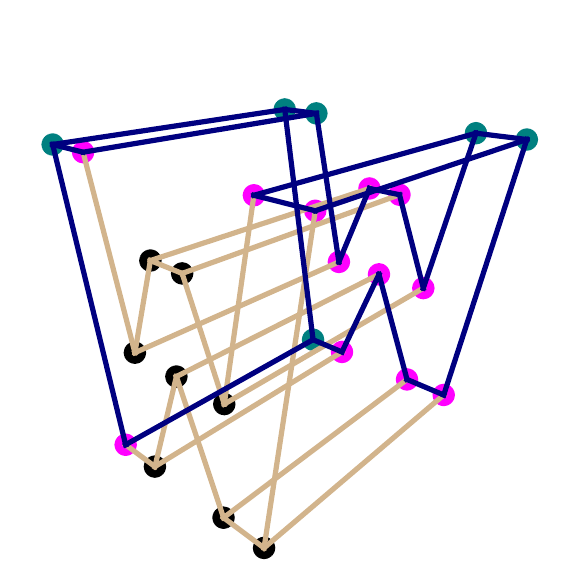}\\
         & \multicolumn{4}{c}{ Input View} & \multicolumn{4}{c}{ Novel View } \\
        \bottomrule
    \end{tabular}
    }
    \vspace{-3mm}
    \caption{Visualization results of our DSG model and baseline methods compared with ground truths in different views.}
    \label{fig:visualization}
    \vspace{-6mm}
\end{figure*}

For the first possible baseline configuration, we train Pixel2mesh~\cite{pixel2mesh} on our ABC-HoW dataset to reconstruct the 3D mesh of the object from single-view images. The training and evaluation settings are the same as in their official implementation. As shown in \cref{tab:p2m} and \cref{fig:p2m}, the mesh reconstruction performance of Pixel2Mesh is dramatically dropped, especially for the invisible parts on our dataset as the objects in our ABC-HoW dataset do not have any semantic priors. Due to the design rationale of PC2WF, a prerequisite of 3D wireframe perception is that the input should contain many support points around each expected junction as 
a base for the junction classification, where the distance between the support point and its corresponding ground truth junction should be less than 0.01. The low-quality mesh reconstruction results would be violated for precisely estimating the holistic 3D wireframes.

To further verify this assertion, we compute the junction recall (without considering the precision) with a relaxed distance threshold (0.03 unit) between the point cloud (sampled with 200k points) of the reconstructed meshes and the corresponding ground truth of 3D wireframe models in \cref{tab:p2m}, which quantitatively demonstrates the challenges of such a possible baseline. Accordingly, we directly take the densely-sampled point clouds (with 200k points per sample) as the ideal inputs of single-view reconstruction. Then, we retrain the best-performing method PC2WF~\cite{liu2021pc2wf} as a ``perfect" baseline, which assumes that perfect meshes can be obtained from the single-view or multi-view images, thereafter leading to very promising results.

For the second possible baseline configuration, we extend HAWP~\cite{xue2020holistically}, the state-of-the-art 2D visible wireframe detector to handle the task of HoW-3D by adding two modules of depth prediction and refinement. Specifically, we first add another hidden branch with the same convolutional architecture as the original visible branch to directly learn both visible and hidden 2D geometries (junctions and lines). Then, a depth estimation branch is applied to obtain the holistic 2.5D wireframe and leverage a GCN to refine the 3D wireframe as used in our DSG model. We term this baseline as ``HAWP-EX".

\subsection{Quantitative and Qualitative Results}

\begin{table*}[!t]
    \centering
    \caption{
        The sAP evaluation on the 3D space for all types (visible and hidden) of line segments under different threshold. 
        For PC2WF, we only report the sAP for ``all" lines as it does not separately handle the lines according to the visibility.
    }
    \vspace{-3mm}
    \label{tab:3D-line-sAP}
    \resizebox{0.85\linewidth}{!}{
    \begin{tabular}{c|cccc|cccc|cccc}
    \toprule
         & \multicolumn{4}{c|}{sAP$_{L_v}$} & 
         \multicolumn{4}{c|}{sAP$_{L_h}$}
         & \multicolumn{4}{c}{sAP$_{L_{\text{all}}}$} \\\midrule
         & 0.01 & 0.03 & 0.05 & 0.07 &0.01 & 0.03 & 0.05 & 0.07 & 0.01 & 0.03 & 0.05 & 0.07\\\cmidrule{2-13}
    PC2WF~\cite{liu2021pc2wf} & - & - &- &- &- &- &- &-  &85.0 & 88.9 & 89.2 & 89.4 \\
    HAWP-EX~\cite{xue2020holistically} & 71.4 & 86.5 &90.0 &91.4 &36.0 &52.3 &56.3 &58.0  & 58.2 &73.8 &77.4 &78.9 \\
    Ours &79.0 & 87.8 & 90.3 & 91.1 &76.3 & 83.5 & 85.1 & 85.9 &77.9 & 86.5 & 88.6 & 89.5 
    \\\bottomrule
    \end{tabular}
    }
\end{table*}

\begin{table*}[!t]
    \vspace{-3mm}
    \centering
    \caption{
        The AP evaluation on the 3D space for all types (visible, fleeting, and hidden) of junctions under the different thresholds. 
        For PC2WF, we only report the AP for ``all" junctions as it does not handle the junctions per the visibility.
    }
    \vspace{-3mm}
    \label{tab:3D-junction-AP}
    \resizebox{0.85\linewidth}{!}{
     \begin{tabular}{c|ccc|ccc|ccc|ccc}
    \toprule
         & \multicolumn{3}{c|}{AP$_{J_v}$} & 
         \multicolumn{3}{c|}{AP$_{J_f}$}
         & \multicolumn{3}{c|}{AP$_{J_h}$} & \multicolumn{3}{c}{AP$_{J_{\text{all}}}$} \\\midrule
         & 0.02 & 0.03 & 0.05 & 0.02 & 0.03 & 0.05 & 0.02 & 0.03 & 0.05 & 0.02 & 0.03 & 0.05\\\cmidrule{2-13}
    PC2WF~\cite{liu2021pc2wf} & - & - &- &- &- &- &- &- &- & 91.9 & 92.0 & 92.1\\
    HAWP-EX~\cite{xue2020holistically} & 20.9 &37.0 &60.9 &18.4 & 34.0 & 57.2 & 0.9 & 3.8 & 16.7 &15.7 & 28.9 & 50.2\\
    Ours & 31.7 & 50.5 & 70.1 & 26.6 & 44.6 & 65.4 & 16.3 & 35.2 & 63.2 & 24.5 & 43.0 & 66.0
    \\\bottomrule
    \end{tabular}
    }
    \vspace{-2mm}
\end{table*}

\begin{table*}[!t]
    \vspace{-2mm}
    \centering
    \begin{minipage}[t]{0.48\linewidth}
        \centering
        \captionof{table}{\small The average precision for different types of junctions.}
        \vspace{-3mm}
        \label{tab:junction-AP-2D}
        \resizebox{!}{0.11\linewidth}{
        \begin{tabular}{c|cc|cc|cc|cc}
    \toprule
         & \multicolumn{2}{c|}{AP$_{J_v}$} & 
         \multicolumn{2}{c|}{AP$_{J_f}$} & \multicolumn{2}{c|}{AP$_{J_h}$}
         & \multicolumn{2}{c}{AP$_{J_{\text{all}}}$} \\\midrule
         & 1.0 & 2.0 & 1.0 & 2.0 &1.0 & 2.0 & 1.0 & 2.0\\\cmidrule{2-9}
    HAWP-EX~\cite{xue2020holistically} &95.8  & 96.0 &94.0 &95.1 &29.8 &54.8 &82.0 &87.3 \\
    Ours & 94.6 & 94.8 &92.0 &92.3 &46.2 &80.9 &81.3 &90.8 \\\bottomrule
    \end{tabular}
    }
    \end{minipage}
    \hfill
    \begin{minipage}[t]{0.48\linewidth}
        \centering
        \captionof{table}{\small The sAP evaluation on the 2D plane for all types of lines}
        \vspace{-3mm}
        \label{tab:Line-AP-2D}
        \resizebox{!}{0.11\linewidth}{
        \begin{tabular}{c|cc|cc|cc}
    \toprule
         & \multicolumn{2}{c|}{sAP$_{L_v}$} & \multicolumn{2}{c|}{sAP$_{L_h}$}
         & \multicolumn{2}{c}{sAP$_{L_{\text{all}}}$} \\\midrule
         & 10.0 & 15.0 & 10.0 & 15.0 &10.0 & 15.0\\\cmidrule{2-7}
    HAWP-EX ~\cite{xue2020holistically} &95.2 &95.2 &60.7 &63.2 &82.7 &83.6  \\
    Ours & 93.3 & 93.3 &83.3 &84.9 &89.8 &90.4 \\\bottomrule
    \end{tabular}
        }
    \end{minipage}
\end{table*}

\begin{table*}[!t]
\vspace{-4mm}
\centering
 \begin{minipage}[t]{0.48\textwidth}
 \centering
 \captionof{table}{\small Ablations on the \emph{HiddenTR}.}
 \vspace{-3mm}
 \label{tab:transformer-ablation}
 \resizebox{!}{0.15\linewidth}{
 \begin{tabular}{c|cc|cccc}
        \toprule
         & \multicolumn{2}{c|}{HiddenTR}  & \multirow{2}*{AP$^1$} & \multirow{2}*{AP$^2$}\\
     & context & geometric & \\\hline
    (a) & No & Yes & 37.2 & 67.0 \\
    (b) & Yes & No & 32.1 & 69.1 \\
    (c) & Yes & Yes & \textbf{46.2} & \textbf{80.9} \\\bottomrule
    \end{tabular}    }
 \end{minipage}
 \hfill
 \begin{minipage}[t]{0.48\textwidth}
 \centering
 \captionof{table}{\small Ablations on the refinement.}
 \vspace{-3mm}
    \label{tab:ablation-GCN-3D}
 \resizebox{!}{0.15\linewidth}{
  \begin{tabular}{c|c|c|c|c}
    \toprule
     setting & AP$^{J_{v}}_{0.05}$ & AP$^{J_{f}}_{0.05}$ & AP$^{J_{h}}_{0.05}$& AP$^{J_{all}}_{0.05}$ \\ 
     \midrule
     w/o refinement & $8.7$ & $7.5$ & $55.2$ &$23.1$\\
     MLP & $46.3$ & $45.5$ & $57.6$ & $49.3$ \\
     GCN & $58.6$ & $56.1$ & $60.8$ & $58.8$ \\
     G-Resnet & $\textbf{70.1}$& $\textbf{65.4}$ &$\textbf{63.2}$ & $\textbf{66.0}$   \\
     \bottomrule
    \end{tabular}
    }
 \end{minipage}
 \vspace{-6mm}
\end{table*}

As reported in \cref{tab:junction-AP-2D} and \cref{tab:Line-AP-2D}, the wireframe detector designed for 2D visible wireframes is failed to detect the invisible parts of the objects from the single-view images, and further leading to a failure in the 3D space (\cref{tab:3D-junction-AP} and \cref{tab:3D-line-sAP}). By contrast, our hierarchical Transformer-based module is more effective to detect the invisible parts. Compared with the ``perfect" baseline that takes the ground truth point clouds as input, our method achieves very competitive sAP scores in the strict thresholds of $0.01$, $0.03$, and $0.05$ for the detected line segments, while being better than PC2WF in a relaxed threshold of $0.07$. For the AP of 3D junctions, as PC2WF takes the point cloud captured in the most ideal configuration and they only need to classify the junction points from the dense point clouds, they obtained the better performance than our DSG model that needs to fill the invisible junctions from single-view images. 

The qualitative evaluation results are shown in \cref{fig:visualization}. The baseline method HAWP-EX remains a significant gap in reconstructing the invisible parts compared with our DSG model, which demonstrate the key difference between the visible and invisible geometry for HoW-3D. We also get better results on the visible parts in the 3D space, as the holistic structure is crucial for the GCN to learn the 3D shape regularities. Although PC2WF~\cite{liu2021pc2wf} can precisely detect 3D junctions, the line verification module that depends on the features extracted from the unstructured point cloud is prone to misclassify the true positive lines. For our method, as the LOI pooling layer takes the features from images, it achieves better robustness for the line classification.

\subsection{Ablation Studies}

\myparagraph{Transformers for Hidden Junctions.} In this experiment, we study the design of learning hidden junctions. We use the predicted 2D location of the hidden junctions to evaluate the average precision under the threshold of $1$ and $2$ pixels. As shown in \cref{tab:transformer-ablation}, we justified that both the global contextual features and the geometric features of visible lines are important for the end task.  

\myparagraph{GCN for Wireframe Refinement.} As reported in \cref{tab:ablation-GCN-3D}, the GCN module is necessary and effective to get a better holistic wireframe perception result. Without using GCN, the AP scores for visible and hidden junctions are greatly lower than the final results. It is worth noting that the AP of the hidden junction is much higher than the visible ones before refinement, because our convolution network is not specialized to predict the depth of the visible junctions, but filling the invisible junction requires the transformer utilizes global context and geometry information to achieve holistic 3D perception. The large performance gap between the MLP and GCN shows that the holistic geometric structure is crucial for refinement.

\section{Conclusion \& Limitations}
This paper studies a new task of HoW-3D for single-view images by proposing an ABC-HoW dataset and a novel Deep Spatial Gestalt (DSG) model, making the first step of perceiving holistic 3D wireframes from single-view images. Our proposed DSG model takes a single-view image as input to hierarchically learn the visible 2.5D wireframe, inferring the hidden 2.5D junctions from the visible geometry, detecting the hidden line segments, and finally resulting in a holistic 3D wireframe model. 
In the experiments, we demonstrated that the proposed DSG model mimics the gestalt principle for human vision systems to fill the invisible geometry from the learned structures.

Our model would be failed when (1) the object contains many tiny or dense parts, or (2) the observation viewpoint contains much less visible cues. We further discuss this in the supplementary material.  Besides, our proposed DSG model only achieves the object-level holistic perception. For the scene-level perception, it is still challenging to infer the hidden geometry from single-view images and we leave it in our future work. We believe that our work sheds light on developing more
powerful geometric deep learning methods for scene-level holistic 3D perception.

{\small
\bibliographystyle{ieee_fullname}
\bibliography{egbib}
}

\newpage
\appendix

\onecolumn
\section*{Supplementary Material}
In the supplementary material, we first discuss the perceptually-meaningful viewpoint for the task of HoW-3D. Then, we present the data samples for our ABC-HoW dataset and provide an additional qualitative comparison to PC2WF\cite{liu2021pc2wf}. In the last, we show the failure case of our DSG model.

\section{Perceptually-meaningful Viewpoints}
\begin{figure}[!h]
    \centering
    \includegraphics[width=0.15\linewidth]{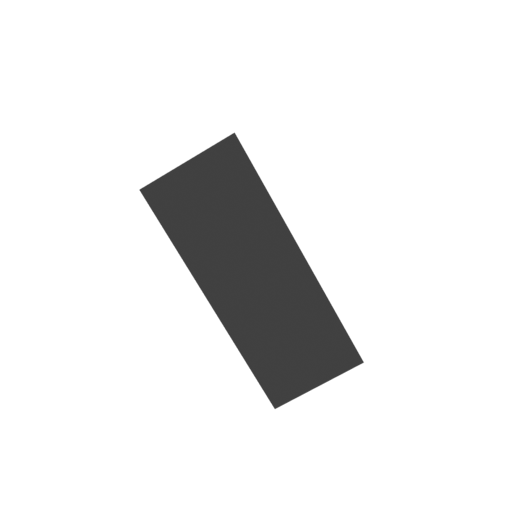}
    \includegraphics[width=0.15\linewidth]{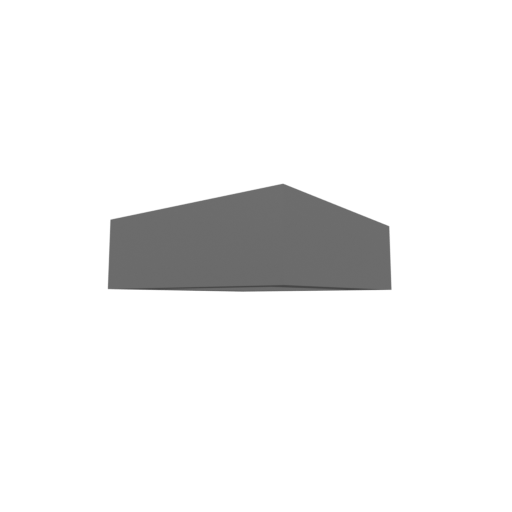}
    \includegraphics[width=0.15\linewidth]{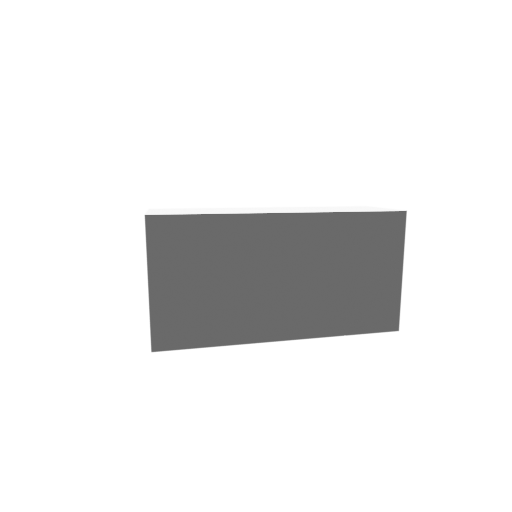}
    \includegraphics[width=0.15\linewidth]{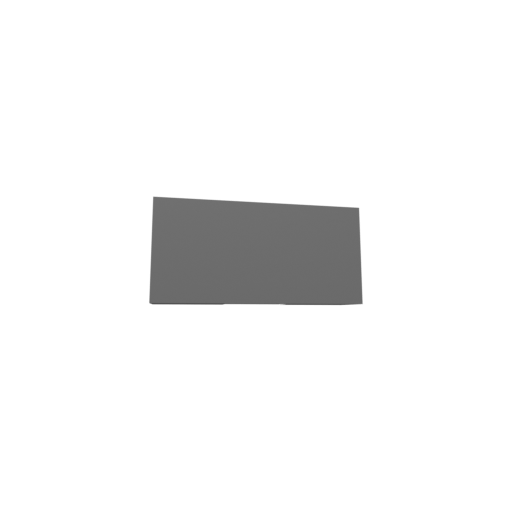}
    \includegraphics[width=0.15\linewidth]{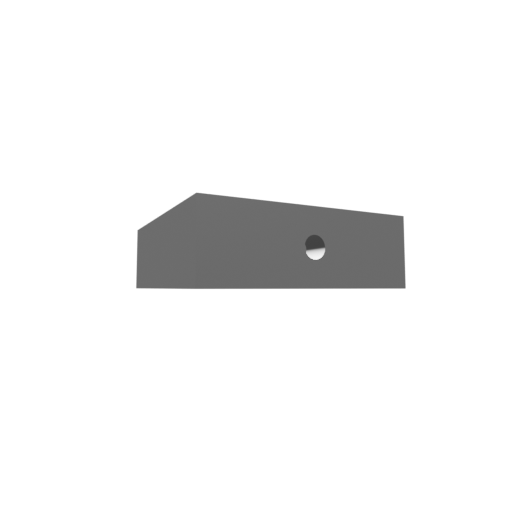}
    \\
    \vspace{-5mm}
    
    \includegraphics[width=0.15\linewidth]{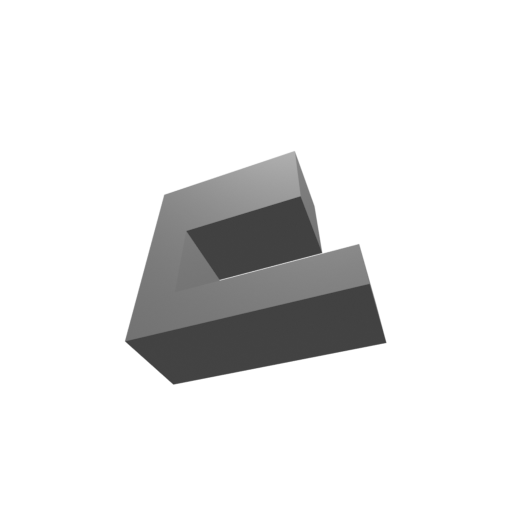}
    \includegraphics[width=0.15\linewidth]{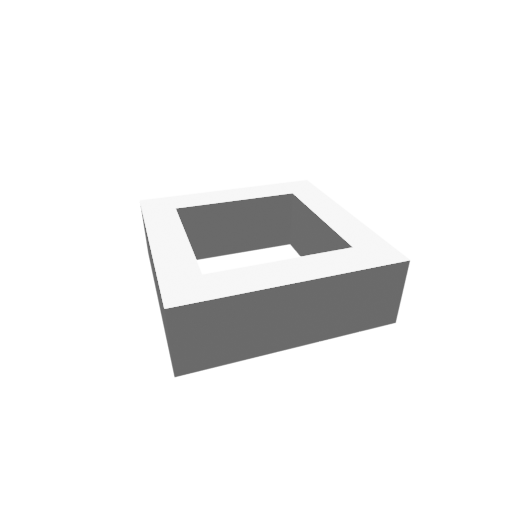}
    \includegraphics[width=0.15\linewidth]{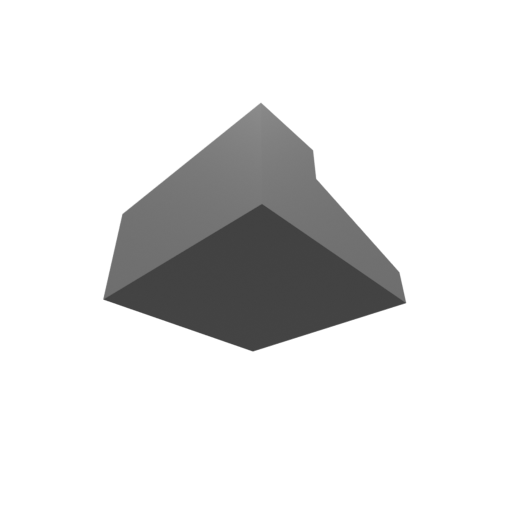}
    \includegraphics[width=0.15\linewidth]{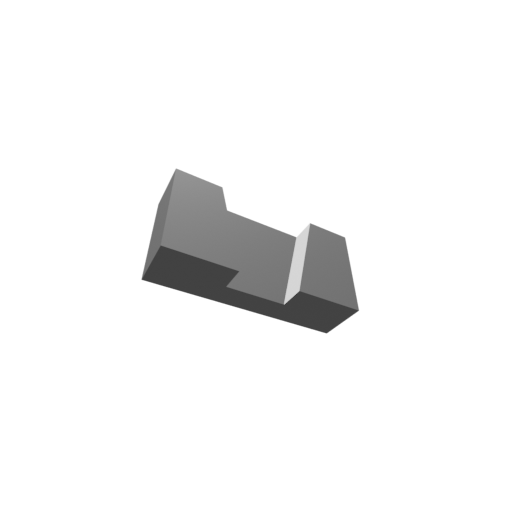}
    \includegraphics[width=0.15\linewidth]{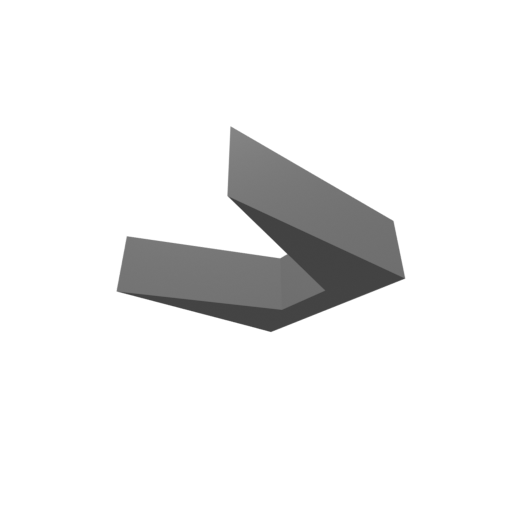}
    \\
    \vspace{-5mm}
    \includegraphics[width=0.15\linewidth]{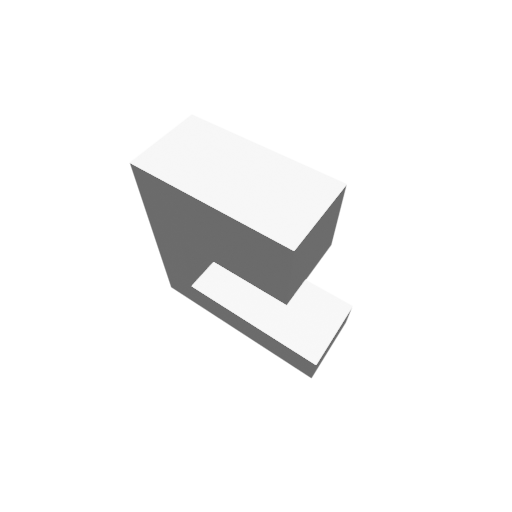}
    \includegraphics[width=0.15\linewidth]{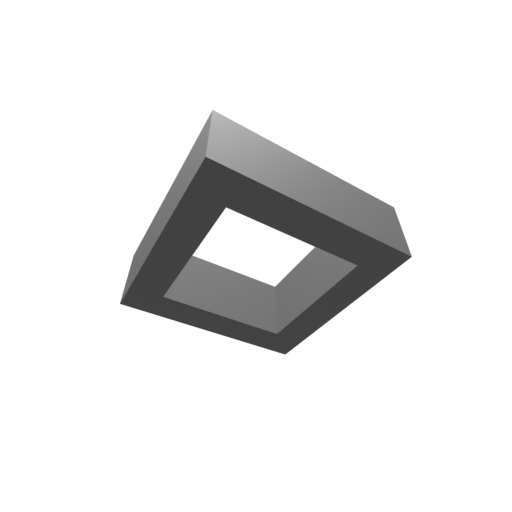}
    \includegraphics[width=0.15\linewidth]{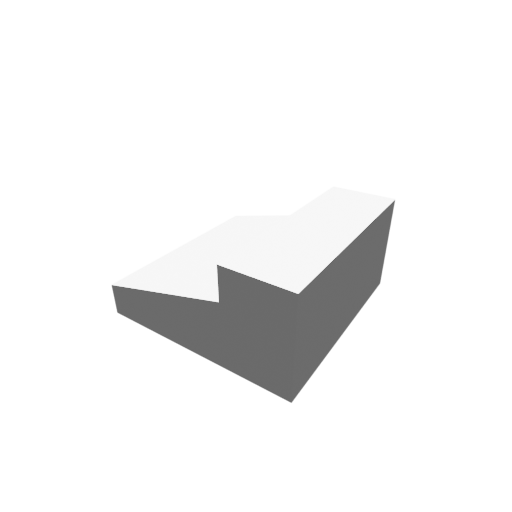}
    \includegraphics[width=0.15\linewidth]{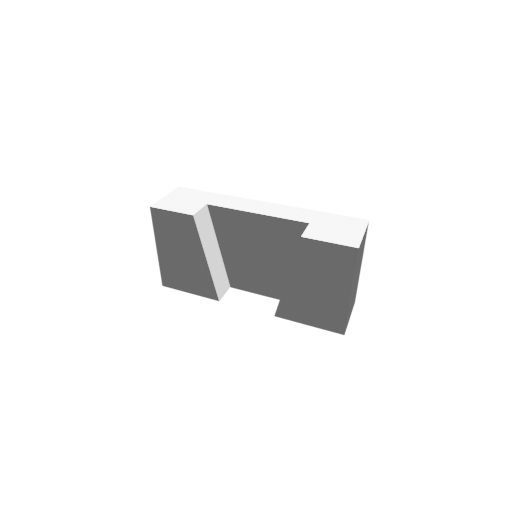}
    \includegraphics[width=0.15\linewidth]{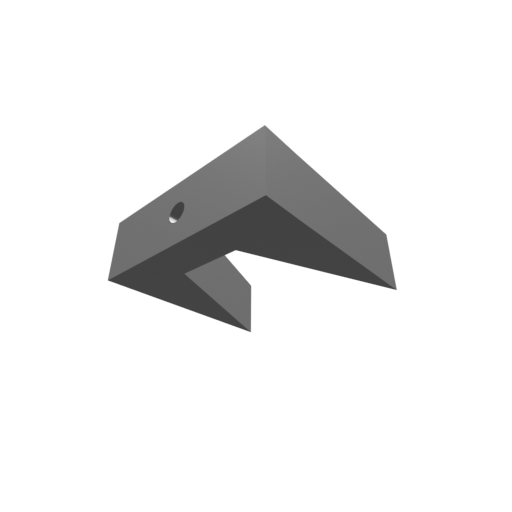}
    \caption{Perceptually-meaningless and the meaningful viewpoints for rendering. In the top row, the rendered images are ambiguous for HoW-3D task. In the middle and bottom rows, we are able to infer the non-line-of-sight geometry.}
    \label{fig:good-viewpoint}
\end{figure}

Since the target of the HoW-3D task is perceiving the holistic structure of 3D objects from single-view images without any semantic prior, it should be noted that \emph{not all viewpoints are feasible to estimate the non-line-of-sight structures}. For example, when the observing viewpoint is occluded by a large 3D plane, it is impossible to guess or infer the structures behind the plane even for humans if we do not know any priors of the scene. Therefore, the rendered image for a 3D object should be perceptually-meaningful for the task of HoW-3D.
To be intuitive, Fig.~\ref{fig:good-viewpoint} visualizes several objects in our ABC-How dataset, rendering them from the perceptually meaningful and the perceptually-meaningless viewpoints. In each column, we show a perceptually-meaningless rendered image in the top row, and two meaningful rendered images in the middle and bottom rows. For the images in the first row, it is very challenging for humans to infer the occluded structures from the single-still images without semantic prior. Therefore, we create the ABC-HoW dataset with a manual viewpoint selection by volunteers. For a rendered image, we ask 3 volunteers to check if it is perceptually meaningless for HoW-3D. If more than half (\ie, two volunteers) vote for the meaningless given rendered image, it will be discarded from the initial ABC-HoW dataset. Furthermore, if there are only 3 (or fewer) rendered images that are perceptually meaningful, we will remove all the images for the corresponding CAD model.
Due to viewpoint checking being time-consuming and expensive, we currently use 24 randomly-generated sparse viewpoints for rendering and checking. 

\section{Data Samples in the ABC-HoW}
\begin{figure}[!h]
    \centering
    \resizebox{0.75\linewidth}{!}{
    \begin{tabular}{cccccc}
        \toprule
         & \includegraphics[height=0.12\linewidth]{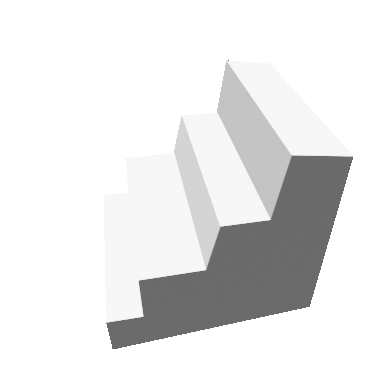} 
         & \includegraphics[height=0.12\linewidth]{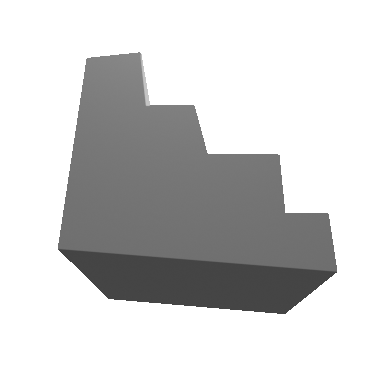}
         & \includegraphics[height=0.12\linewidth]{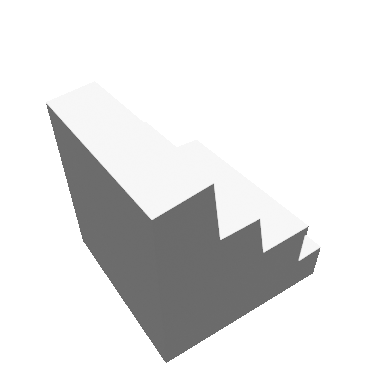}
         & \includegraphics[height=0.12\linewidth]{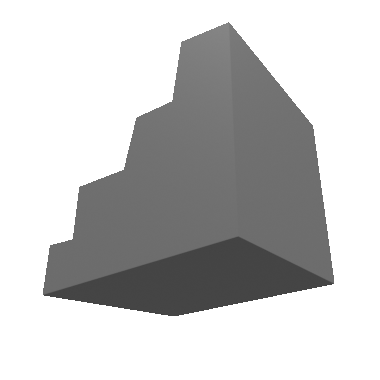}
         & \includegraphics[height=0.12\linewidth]{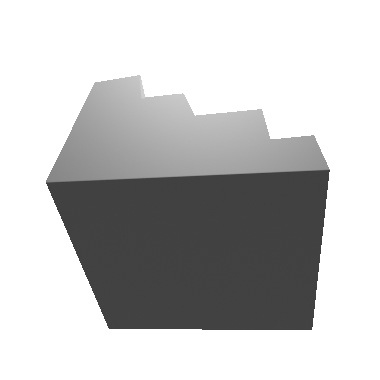}\\
         \multicolumn{6}{c}{
         \begin{tikzpicture}
         \draw [dashed] (3,0) --  ++(8,0);
         \end{tikzpicture}
         }
         \\
         & \includegraphics[height=0.12\linewidth]{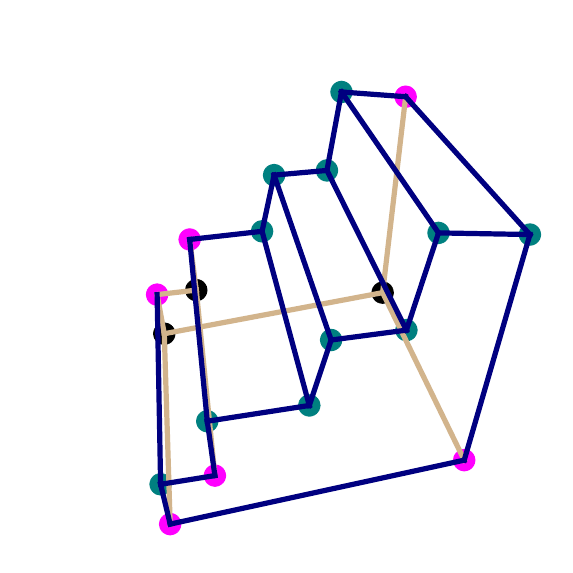}
         & \includegraphics[height=0.12\linewidth]{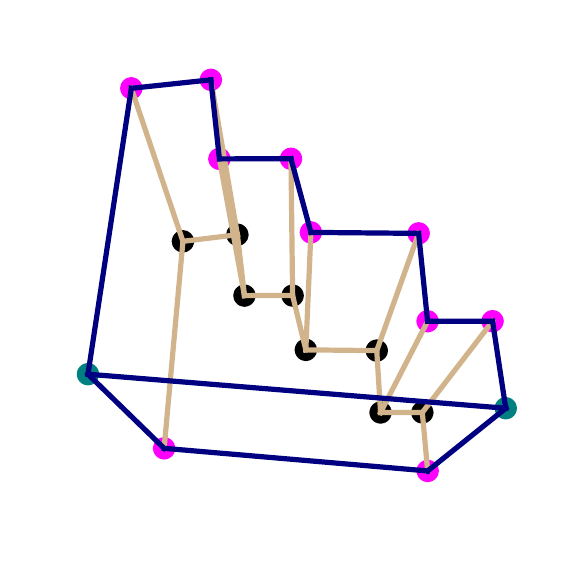}
         & \includegraphics[height=0.12\linewidth]{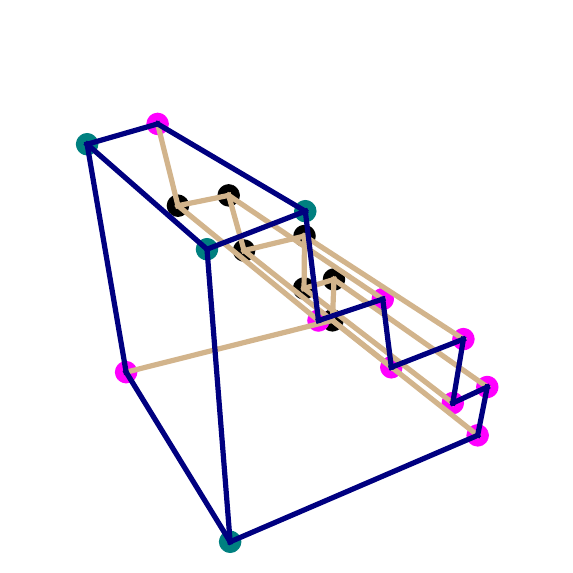}
         & \includegraphics[height=0.12\linewidth]{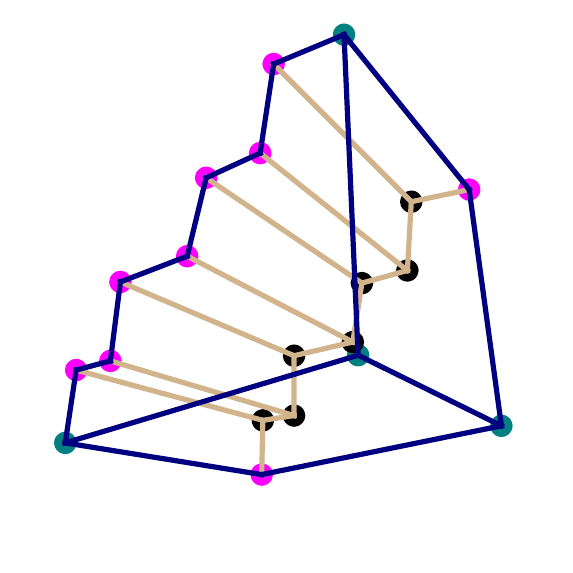}
         & \includegraphics[height=0.12\linewidth]{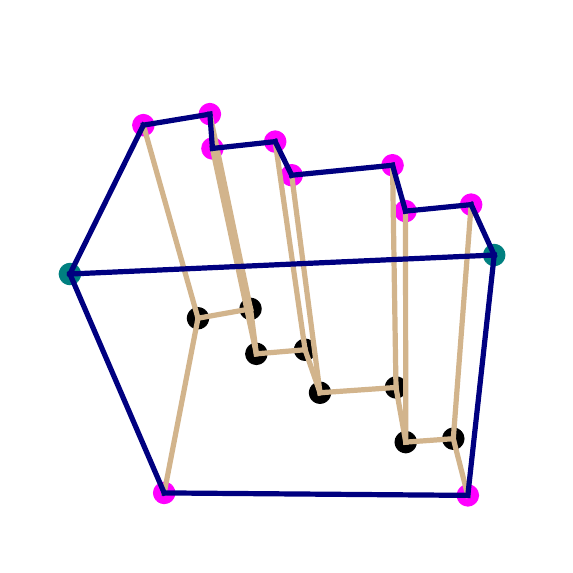}\\
         \midrule
        & \includegraphics[height=0.12\linewidth]{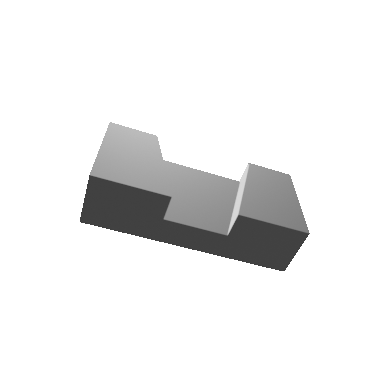} 
         & \includegraphics[height=0.12\linewidth]{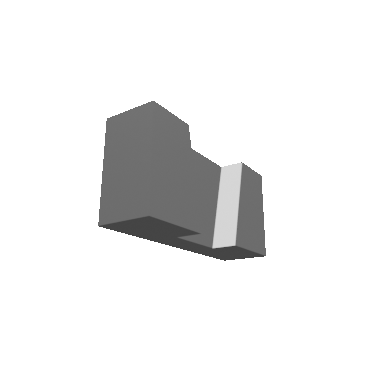}
         & \includegraphics[height=0.12\linewidth]{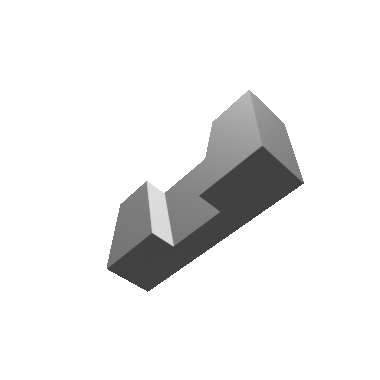}
         & \includegraphics[height=0.12\linewidth]{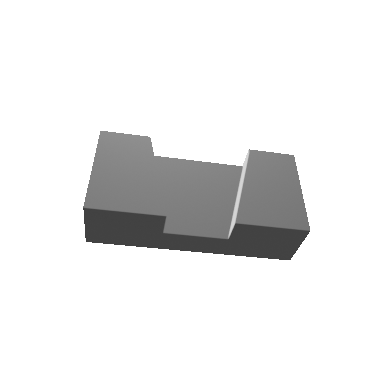}
         & \includegraphics[height=0.12\linewidth]{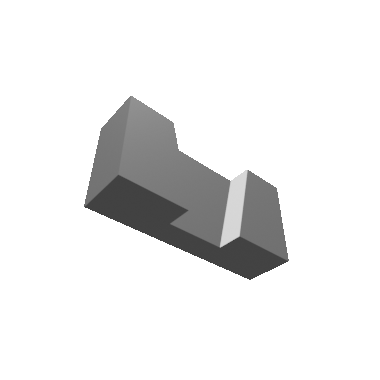}\\
         \multicolumn{6}{c}{
         \begin{tikzpicture}
         \draw [dashed] (3,0) --  ++(8,0);
         \end{tikzpicture}
         }
         \\
         & \includegraphics[height=0.12\linewidth]{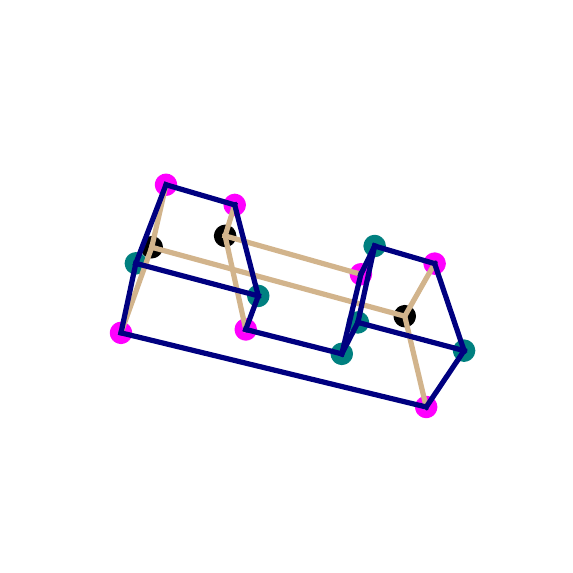}
         & \includegraphics[height=0.12\linewidth]{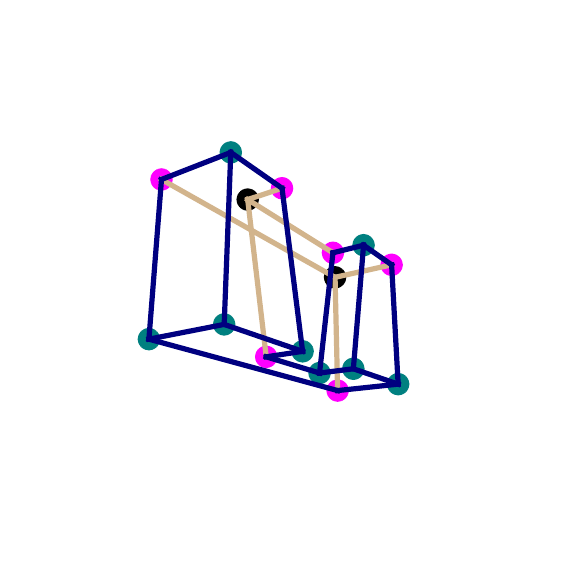}
         & \includegraphics[height=0.12\linewidth]{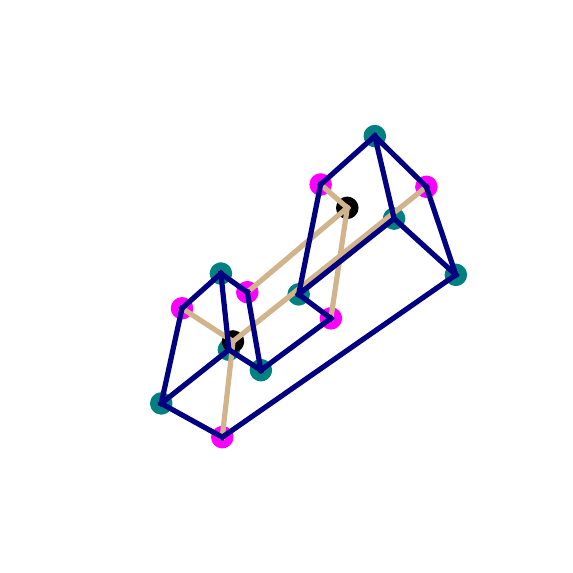}
         & \includegraphics[height=0.12\linewidth]{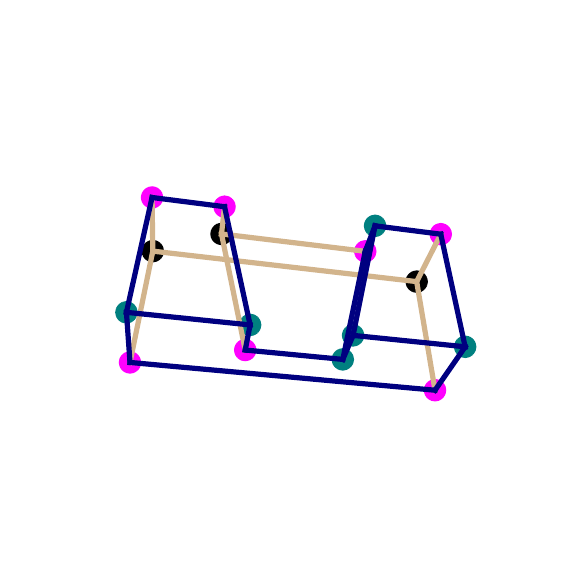}
         & \includegraphics[height=0.12\linewidth]{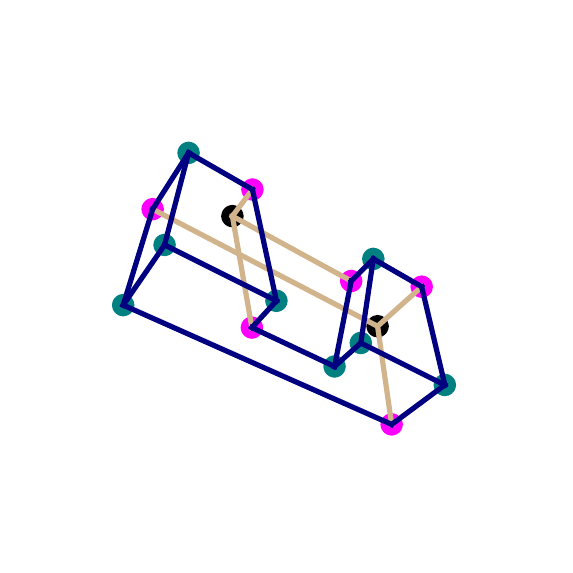}\\
         \midrule
        & \includegraphics[height=0.12\linewidth]{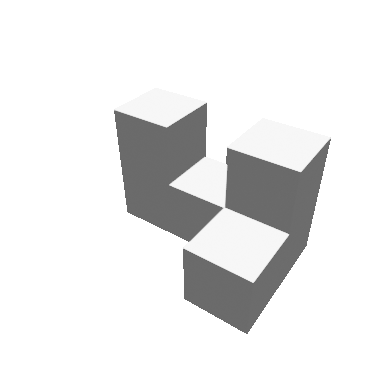} 
         & \includegraphics[height=0.12\linewidth]{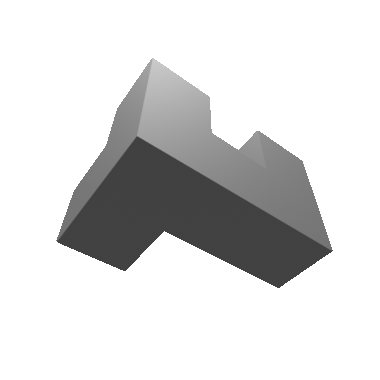}
         & \includegraphics[height=0.12\linewidth]{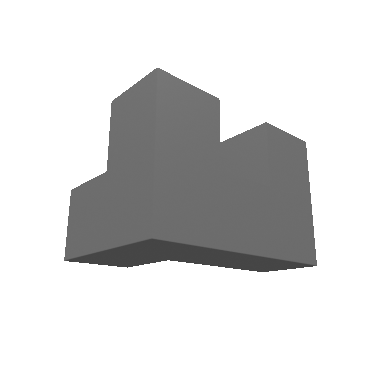}
         & \includegraphics[height=0.12\linewidth]{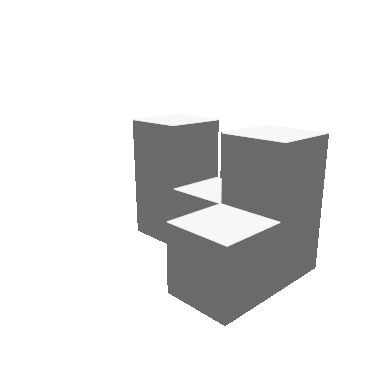}
         & \includegraphics[height=0.12\linewidth]{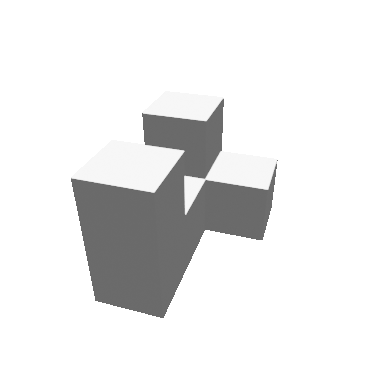}\\
         \multicolumn{6}{c}{
         \begin{tikzpicture}
         \draw [dashed] (3,0) --  ++(8,0);
         \end{tikzpicture}
         }
         \\
         & \includegraphics[height=0.12\linewidth]{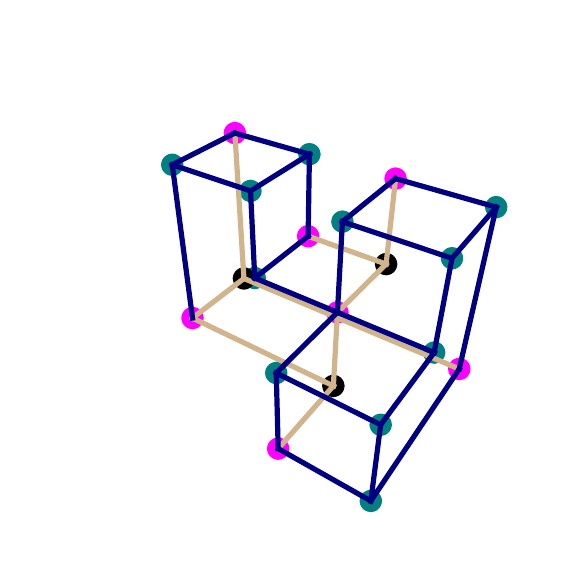}
         & \includegraphics[height=0.12\linewidth]{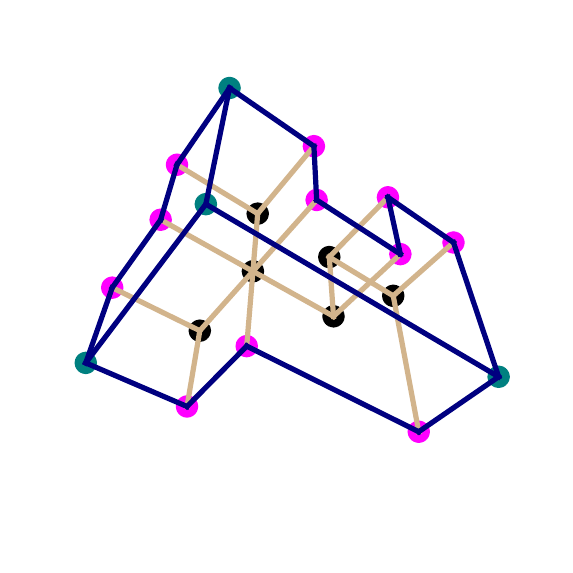}
         & \includegraphics[height=0.12\linewidth]{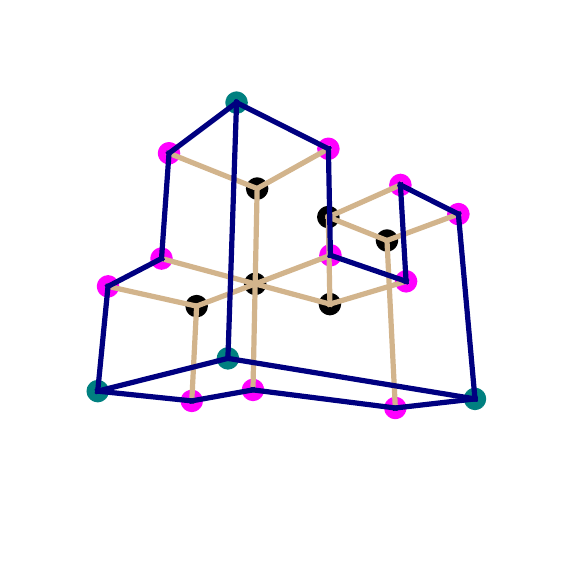}
         & \includegraphics[height=0.12\linewidth]{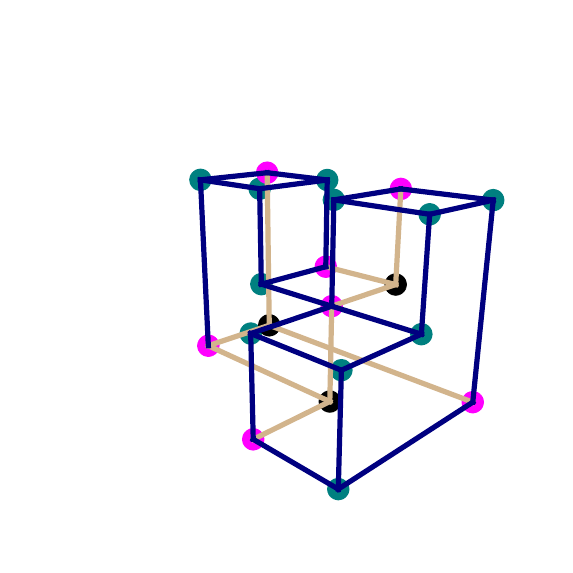}
         & \includegraphics[height=0.12\linewidth]{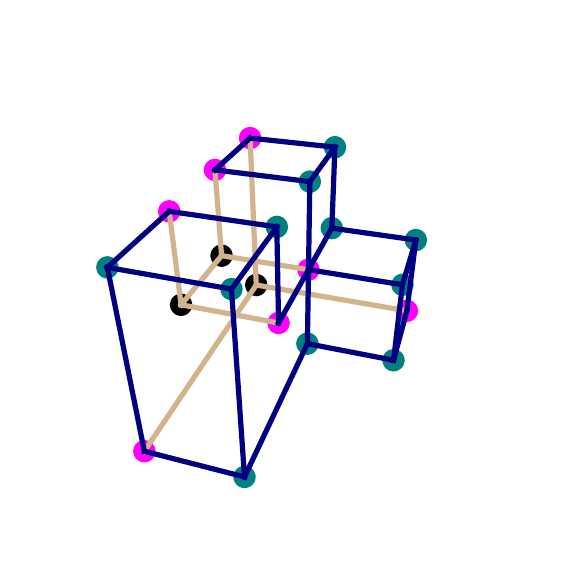}\\
         \midrule
         & \includegraphics[height=0.12\linewidth]{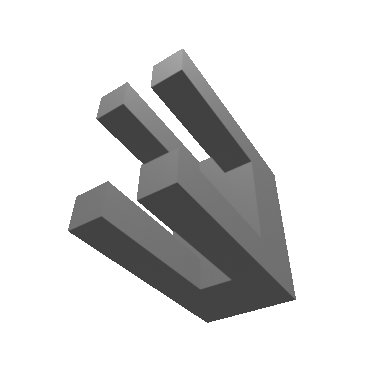} 
         & \includegraphics[height=0.12\linewidth]{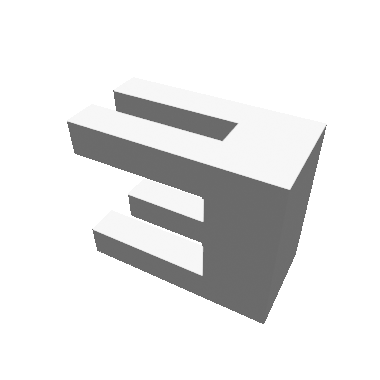}
         & \includegraphics[height=0.12\linewidth]{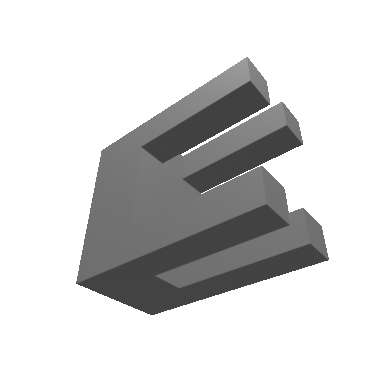}
         & \includegraphics[height=0.12\linewidth]{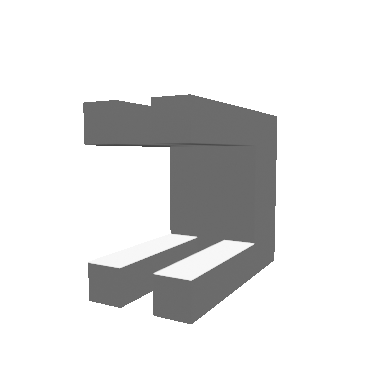}
         & \includegraphics[height=0.12\linewidth]{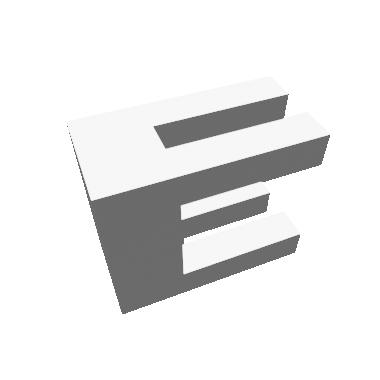}\\
         \multicolumn{6}{c}{
         \begin{tikzpicture}
         \draw [dashed] (3,0) --  ++(8,0);
         \end{tikzpicture}
         }
         \\
         & \includegraphics[height=0.12\linewidth]{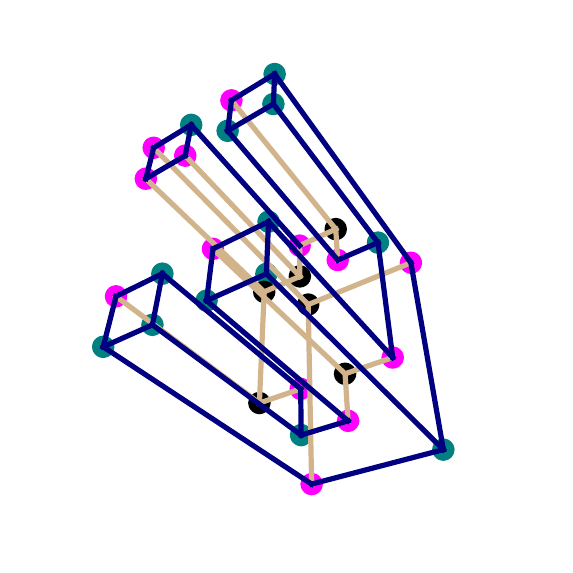}
         & \includegraphics[height=0.12\linewidth]{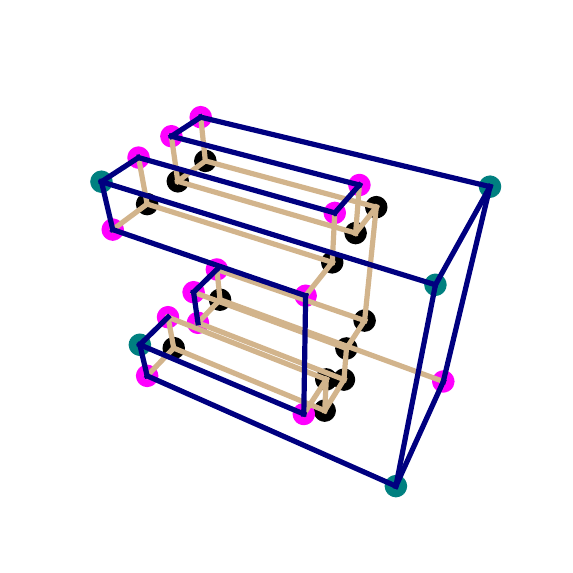}
         & \includegraphics[height=0.12\linewidth]{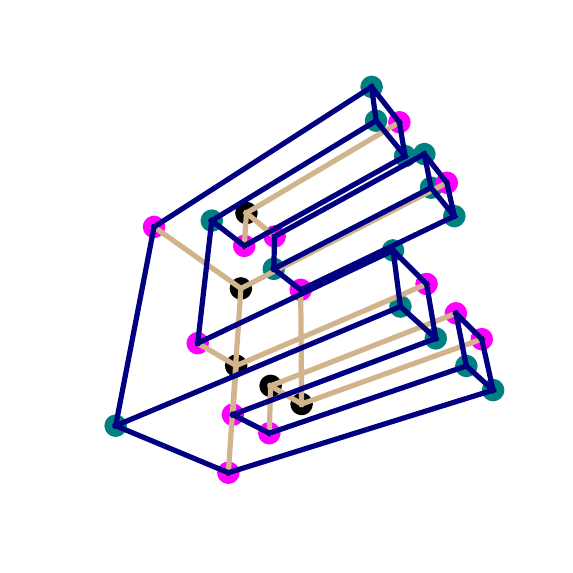}
         & \includegraphics[height=0.12\linewidth]{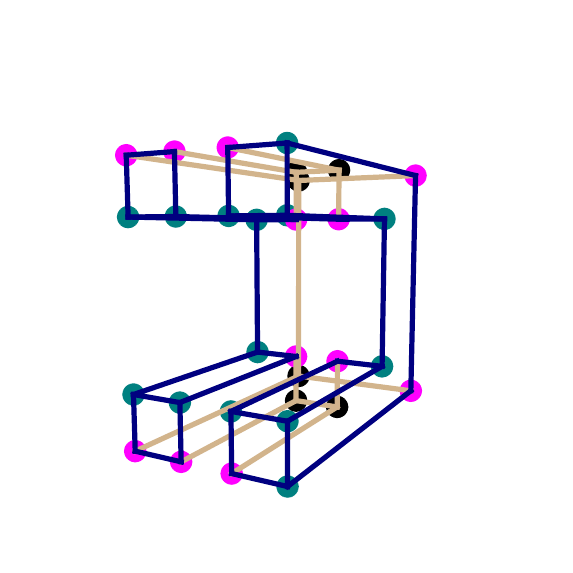}
         & \includegraphics[height=0.12\linewidth]{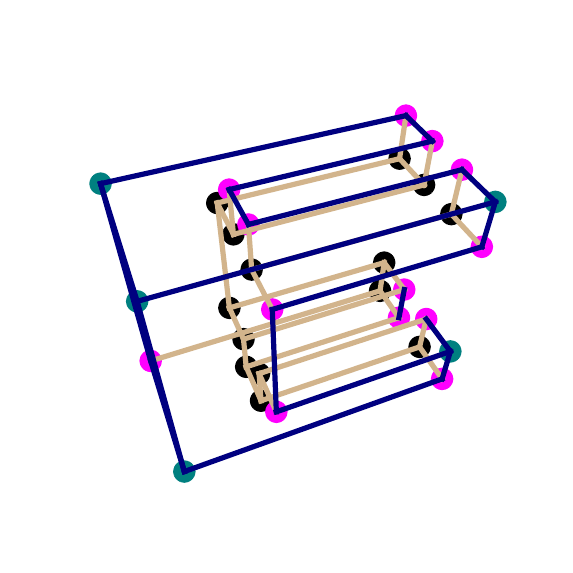}\\
        \bottomrule
    \end{tabular}
    }
    \caption{Examples of our ABC-How dataset.}
    \label{fig:example1}
\end{figure}

\begin{figure}
    \centering
    \resizebox{0.75\linewidth}{!}{
    \begin{tabular}{cccccc}
        \toprule
         & \includegraphics[height=0.12\linewidth]{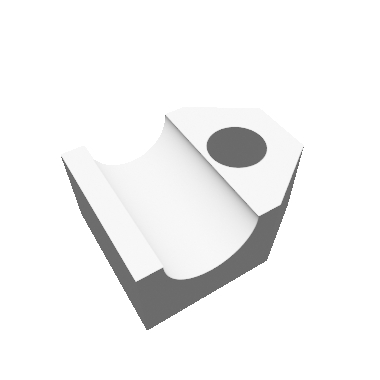} 
         & \includegraphics[height=0.12\linewidth]{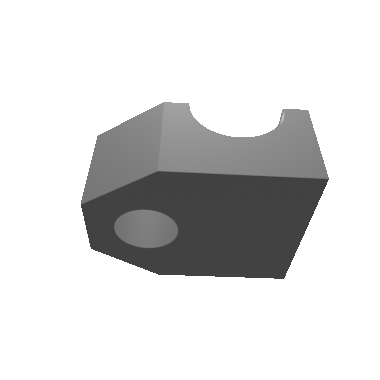}
         & \includegraphics[height=0.12\linewidth]{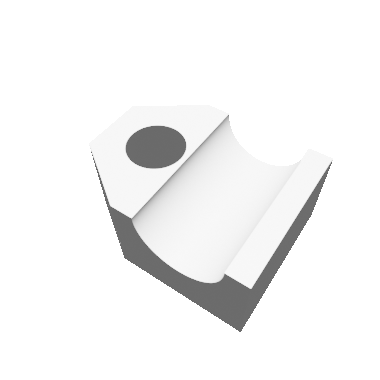}
         & \includegraphics[height=0.12\linewidth]{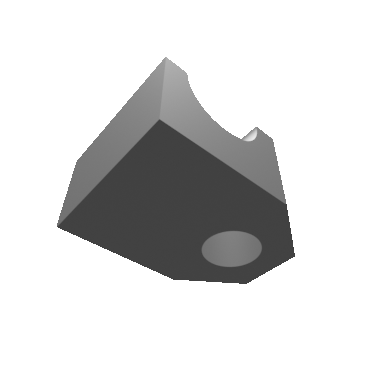}
         & \includegraphics[height=0.12\linewidth]{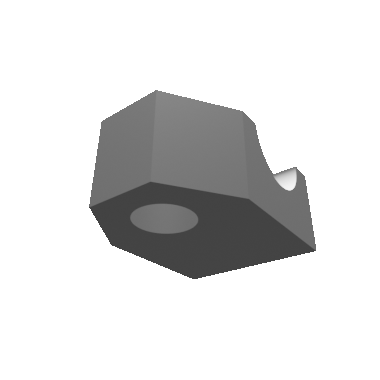}\\
         \multicolumn{6}{c}{
         \begin{tikzpicture}
         \draw [dashed] (3,0) --  ++(8,0);
         \end{tikzpicture}
         }
         \\
         & \includegraphics[height=0.12\linewidth]{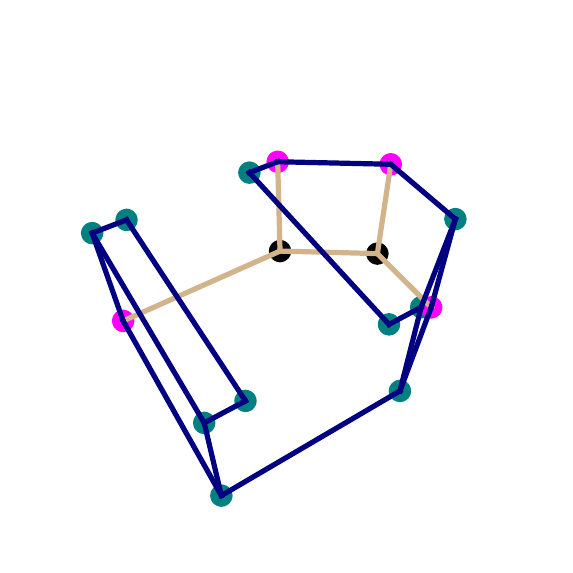}
         & \includegraphics[height=0.12\linewidth]{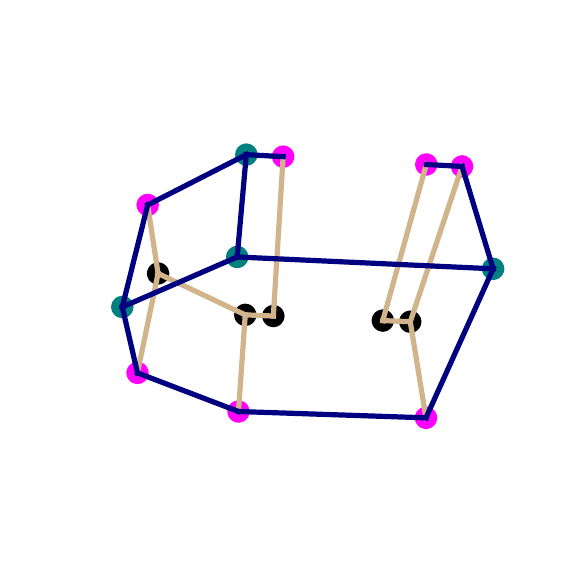}
         & \includegraphics[height=0.12\linewidth]{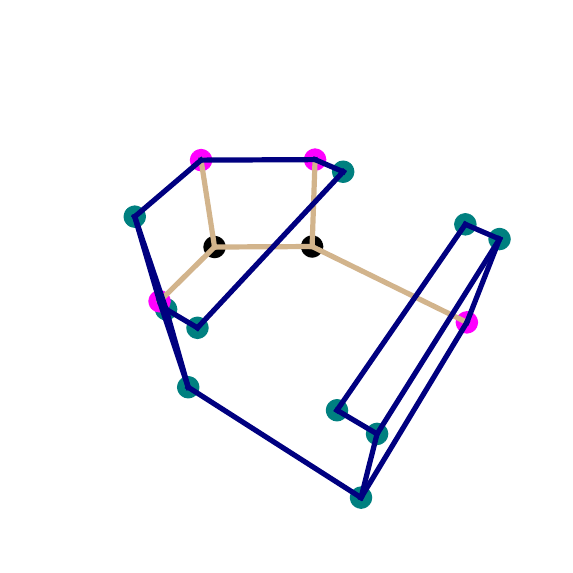}
         & \includegraphics[height=0.12\linewidth]{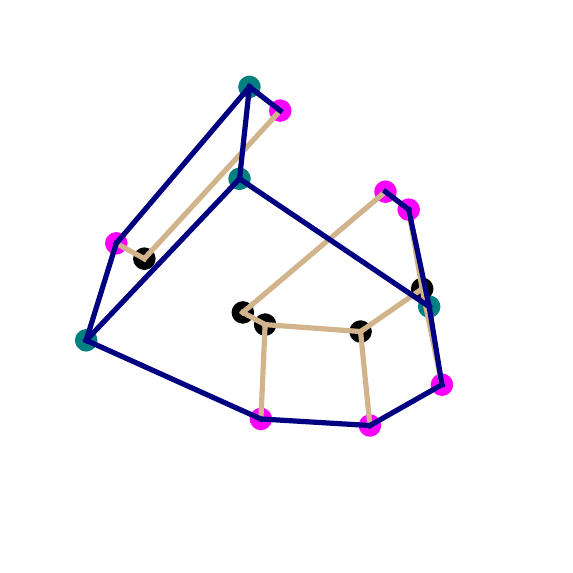}
         & \includegraphics[height=0.12\linewidth]{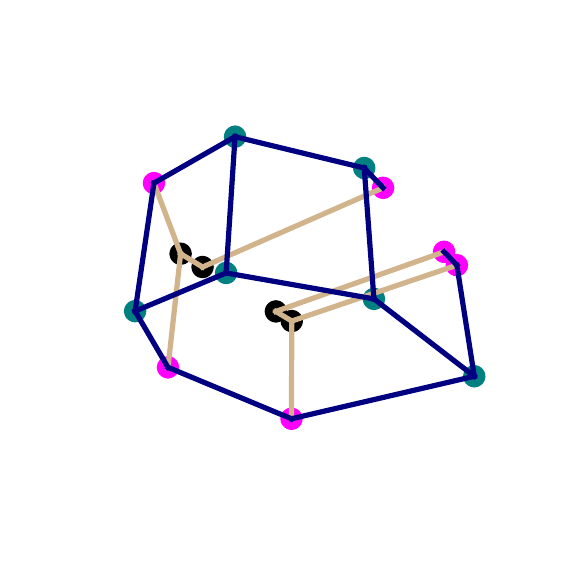}\\
         \midrule
        & \includegraphics[height=0.12\linewidth]{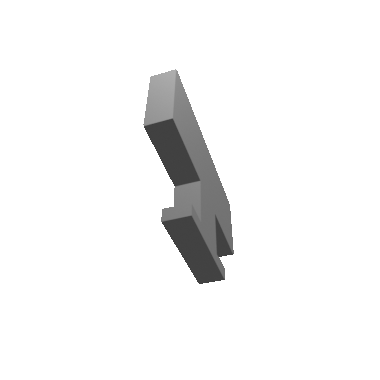} 
         & \includegraphics[height=0.12\linewidth]{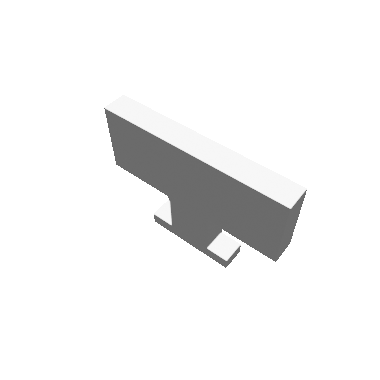}
         & \includegraphics[height=0.12\linewidth]{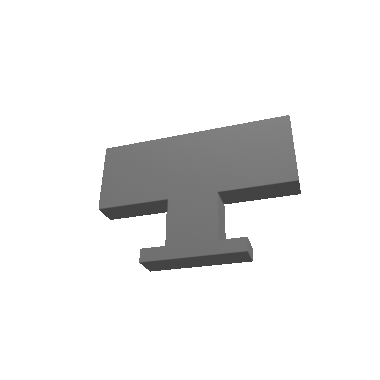}
         & \includegraphics[height=0.12\linewidth]{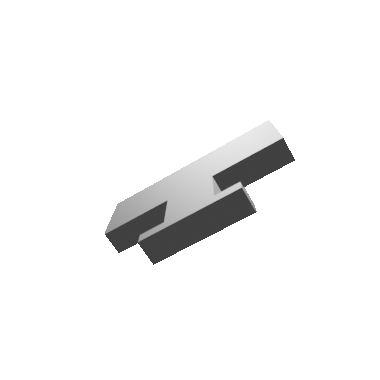}
         & \includegraphics[height=0.12\linewidth]{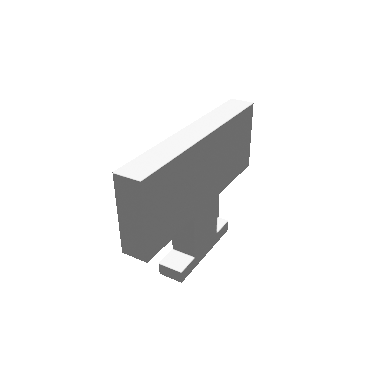}\\
         \multicolumn{6}{c}{
         \begin{tikzpicture}
         \draw [dashed] (3,0) --  ++(8,0);
         \end{tikzpicture}
         }
         \\
         & \includegraphics[height=0.12\linewidth]{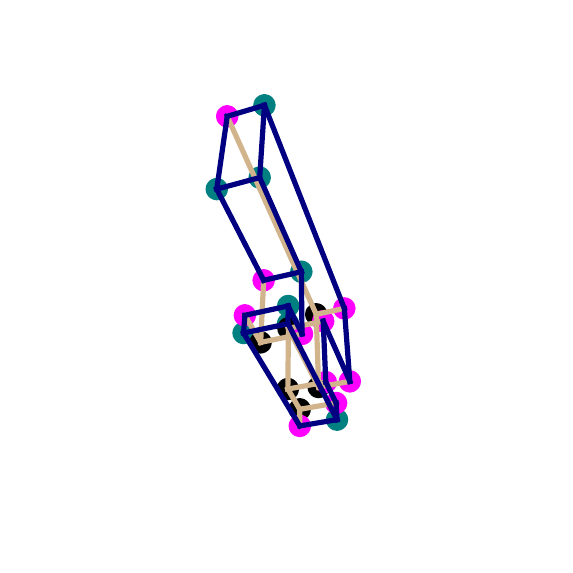}
         & \includegraphics[height=0.12\linewidth]{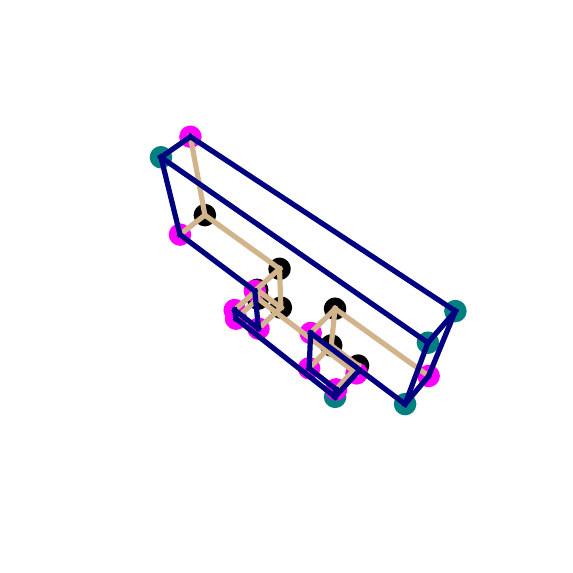}
         & \includegraphics[height=0.12\linewidth]{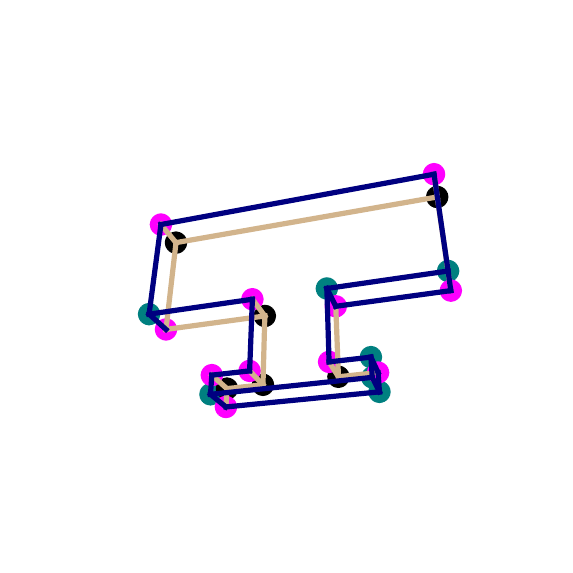}
         & \includegraphics[height=0.12\linewidth]{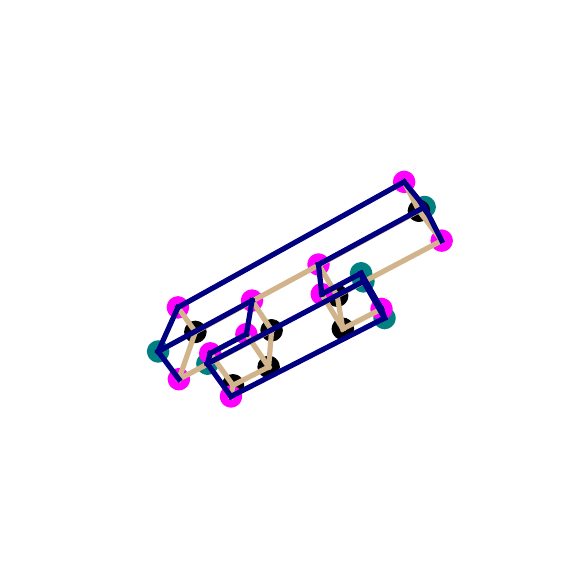}
         & \includegraphics[height=0.12\linewidth]{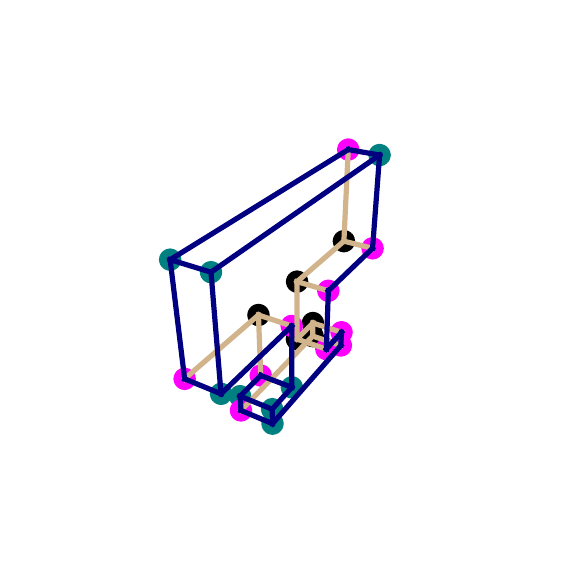}\\
         \midrule
        & \includegraphics[height=0.12\linewidth]{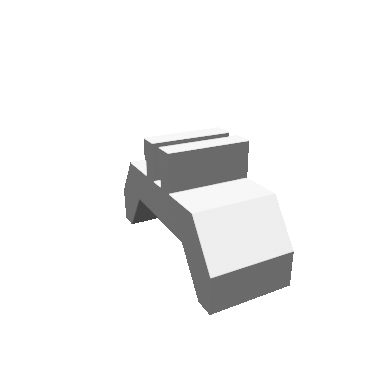} 
         & \includegraphics[height=0.12\linewidth]{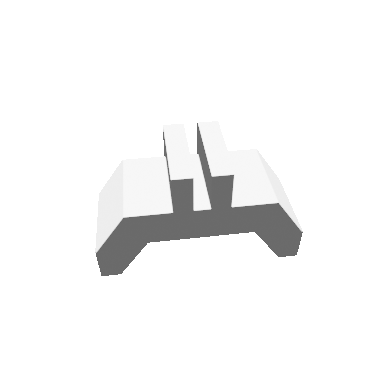}
         & \includegraphics[height=0.12\linewidth]{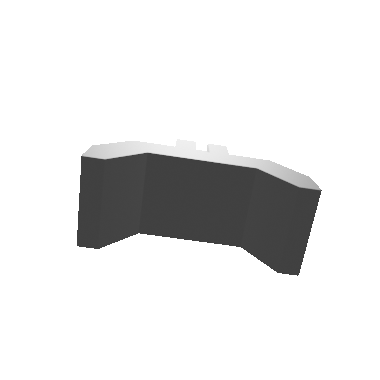}
         & \includegraphics[height=0.12\linewidth]{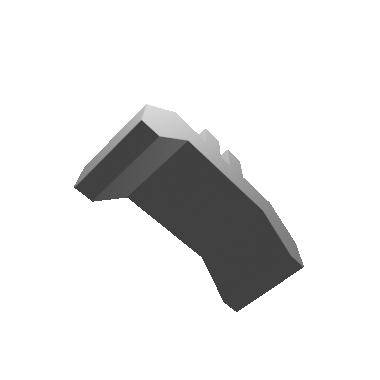}
         & \includegraphics[height=0.12\linewidth]{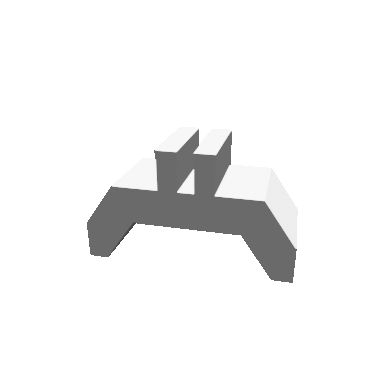}\\
         \multicolumn{6}{c}{
         \begin{tikzpicture}
         \draw [dashed] (3,0) --  ++(8,0);
         \end{tikzpicture}
         }
         \\
         & \includegraphics[height=0.12\linewidth]{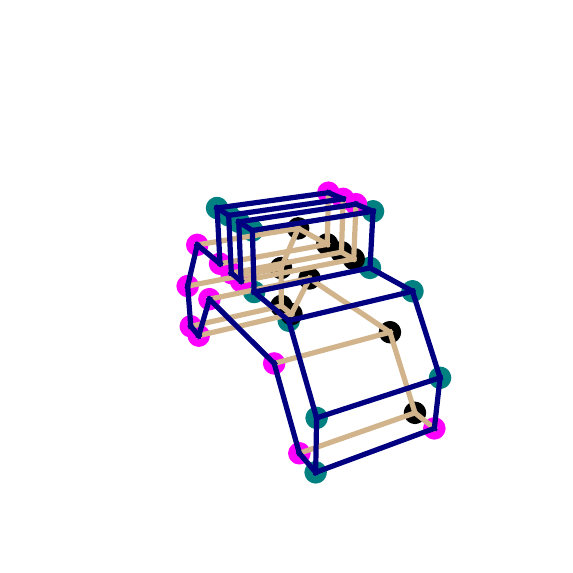}
         & \includegraphics[height=0.12\linewidth]{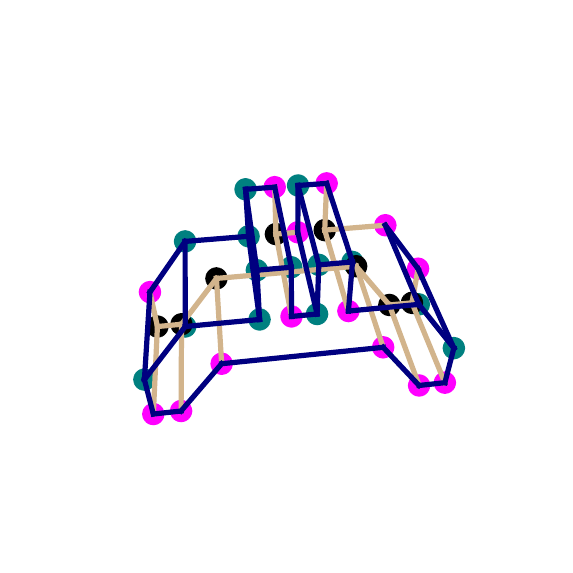}
         & \includegraphics[height=0.12\linewidth]{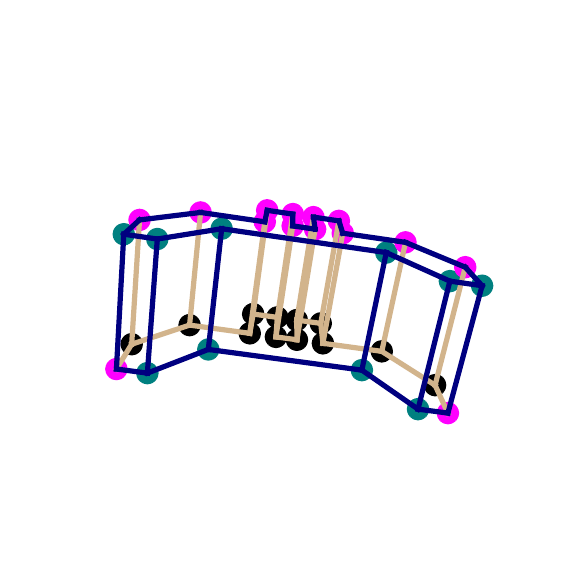}
         & \includegraphics[height=0.12\linewidth]{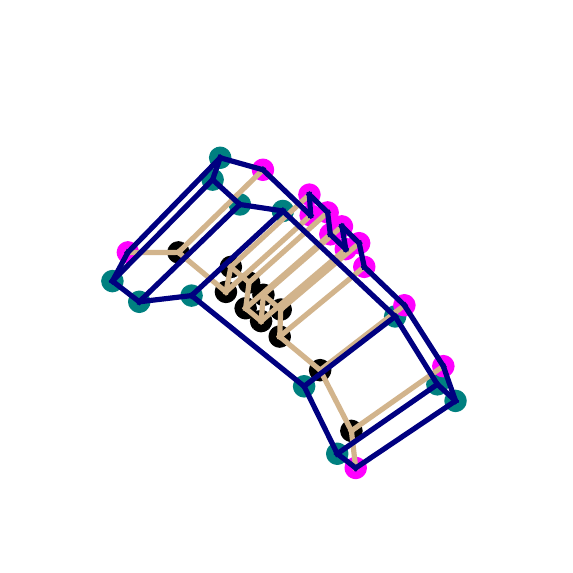}
         & \includegraphics[height=0.12\linewidth]{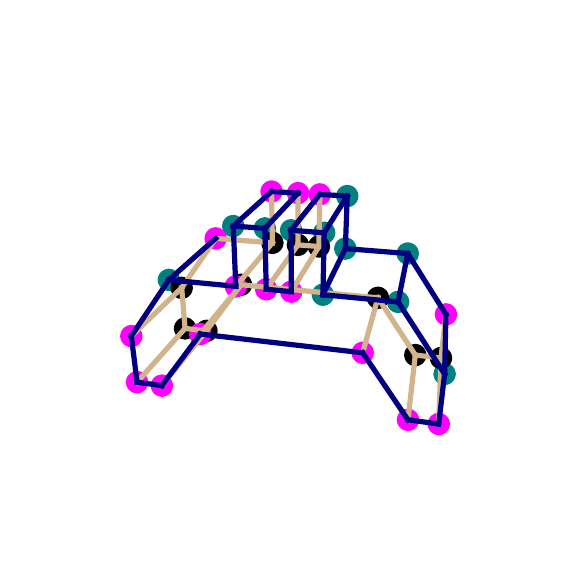}\\
         \midrule
         & \includegraphics[height=0.12\linewidth]{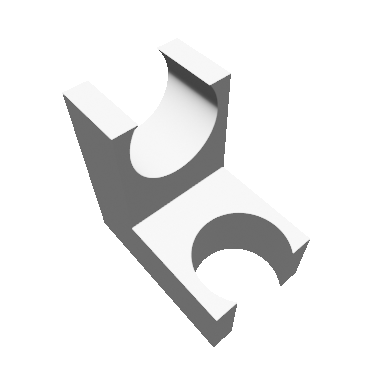} 
         & \includegraphics[height=0.12\linewidth]{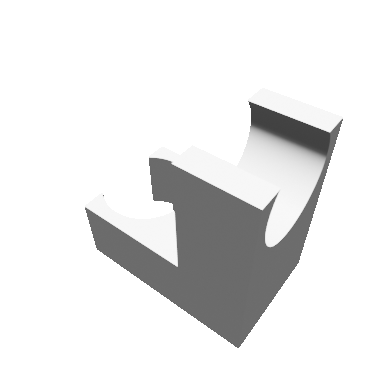}
         & \includegraphics[height=0.12\linewidth]{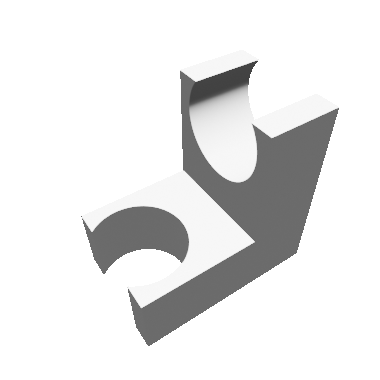}
         & \includegraphics[height=0.12\linewidth]{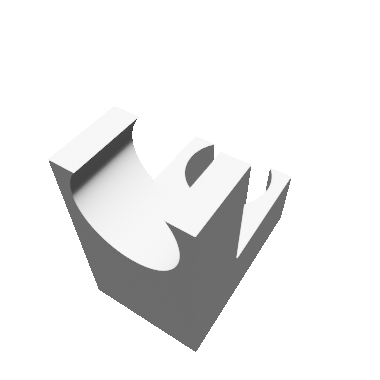}
         & \includegraphics[height=0.12\linewidth]{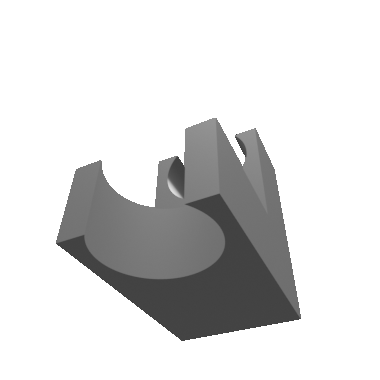}\\
         \multicolumn{6}{c}{
         \begin{tikzpicture}
         \draw [dashed] (3,0) --  ++(8,0);
         \end{tikzpicture}
         }
         \\
         & \includegraphics[height=0.12\linewidth]{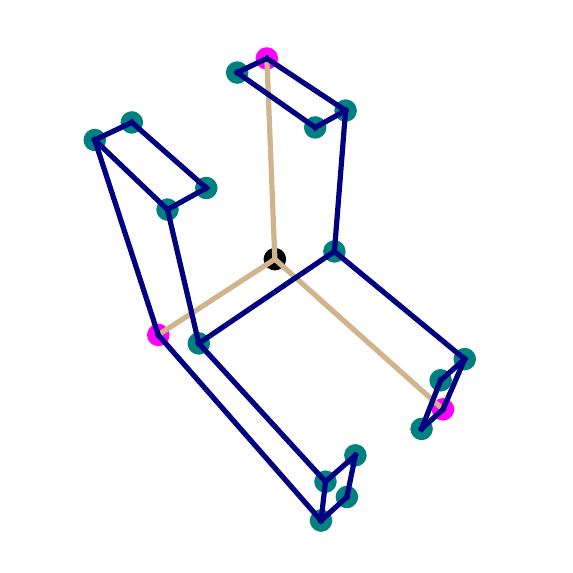}
         & \includegraphics[height=0.12\linewidth]{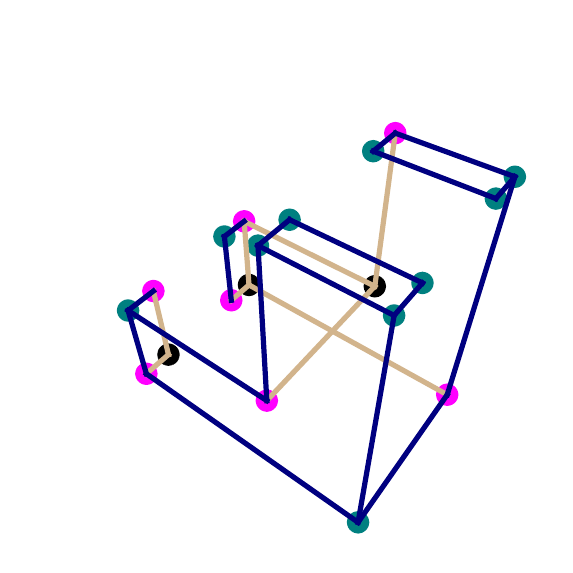}
         & \includegraphics[height=0.12\linewidth]{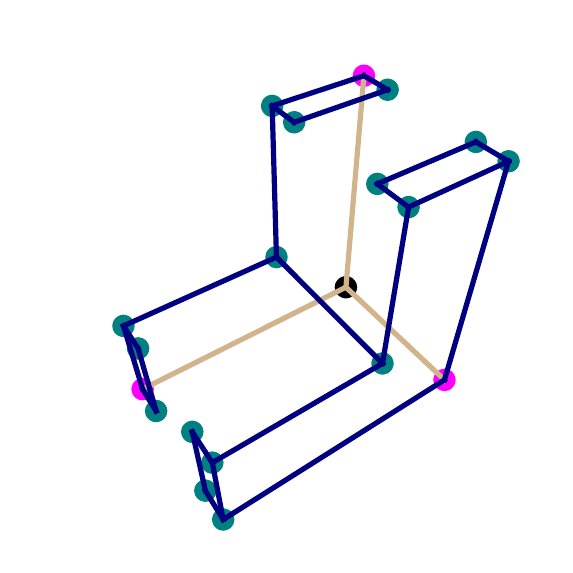}
         & \includegraphics[height=0.12\linewidth]{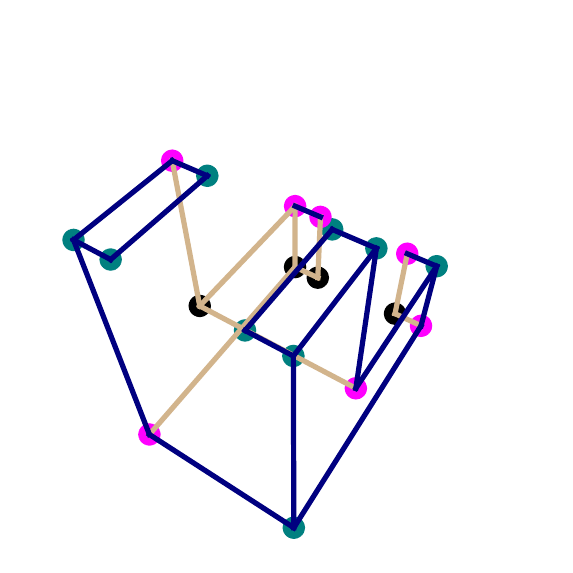}
         & \includegraphics[height=0.12\linewidth]{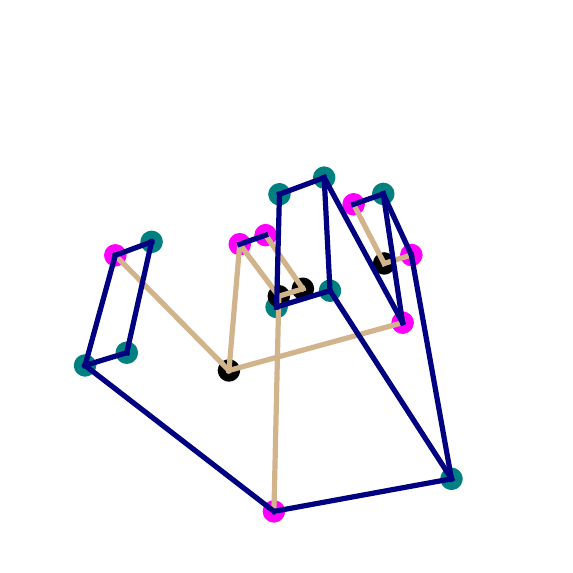}\\
        \bottomrule
    \end{tabular}
    }
    \caption{Examples of our ABC-How dataset.}
    \label{fig:example2}
\end{figure}

Figure~\ref{fig:example1} and Figure~\ref{fig:example2} show some data samples in our ABC-HoW dataset. In each data sample, we show the rendered image and the corresponding holistic 3D wireframe model under the rendering viewpoint together. For the holistic 3D wireframe, the visible and hidden line segments are marked with ``\textcolor{navyblue}{navy blue}" and ``\textcolor{tan}{tan}" respectively, and the junctions are marked as ``\textcolor{teal}{teal}" ,  ``\textcolor{magenta}{magenta}", and ``\textbf{black}" for visible, fleeting, and hidden junctions in the viewpoint. The rendered images are with the resolution of $256\times 256$ in our dataset.

\section{More Visualization Results}
\begin{figure}
\renewcommand{\arraystretch}{0.1}
    \centering
    \resizebox{0.7\linewidth}{!}{
    \begin{tabular}{lcccccc}
        \toprule
         & \includegraphics[height=0.14\linewidth]{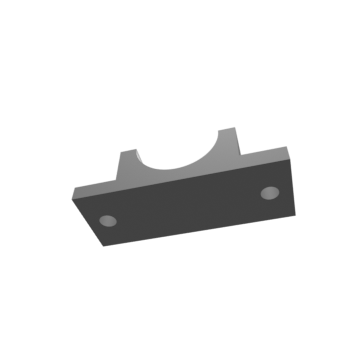} 
         & \includegraphics[height=0.14\linewidth]{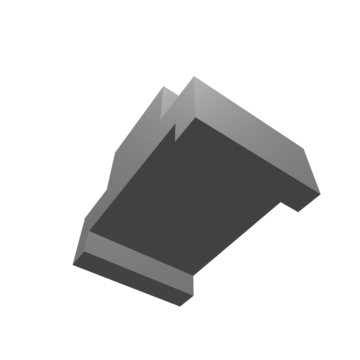}
         & \includegraphics[height=0.14\linewidth]{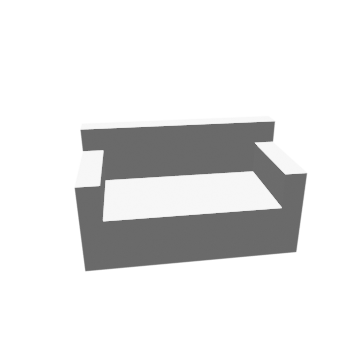}
         & \includegraphics[height=0.14\linewidth]{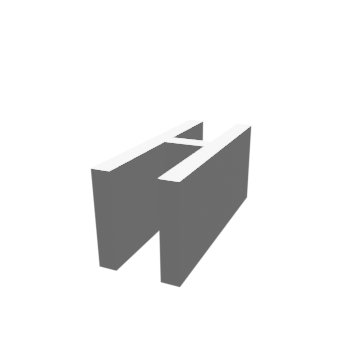}
         & \includegraphics[height=0.14\linewidth]{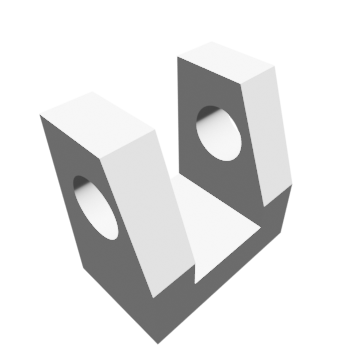}
         & \includegraphics[height=0.14\linewidth]{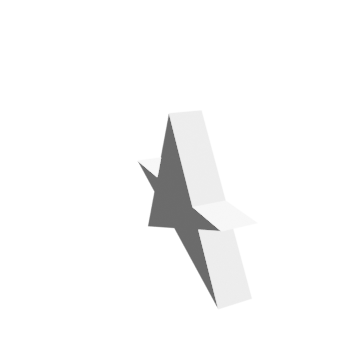}\\
        \midrule
        \multirow{3}{*}{
        \begin{tikzpicture}
         \node [text centered, rotate=90, text width=0.22\linewidth] {Input View};
        \end{tikzpicture}
        }
          & \includegraphics[height=0.14\linewidth]{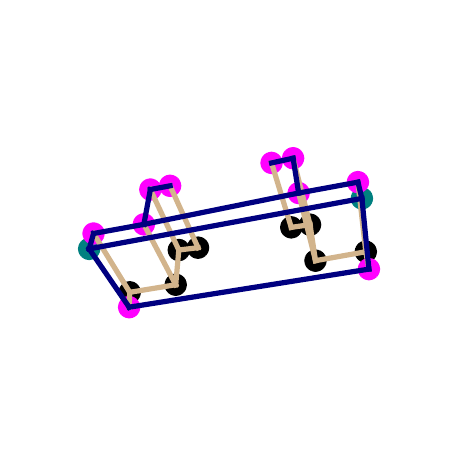} 
         & \includegraphics[height=0.14\linewidth]{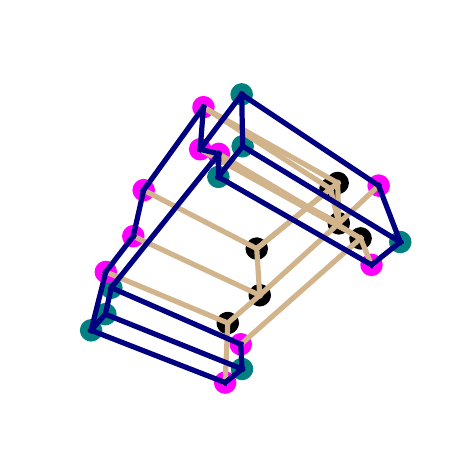}
         & \includegraphics[height=0.14\linewidth]{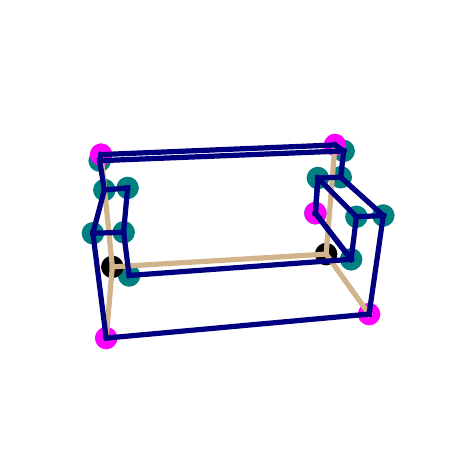}
         & \includegraphics[height=0.14\linewidth]{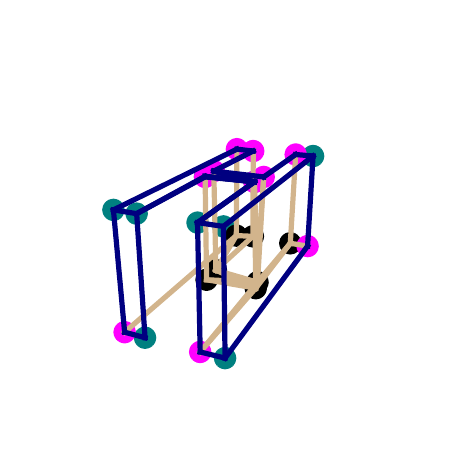}
         & \includegraphics[height=0.14\linewidth]{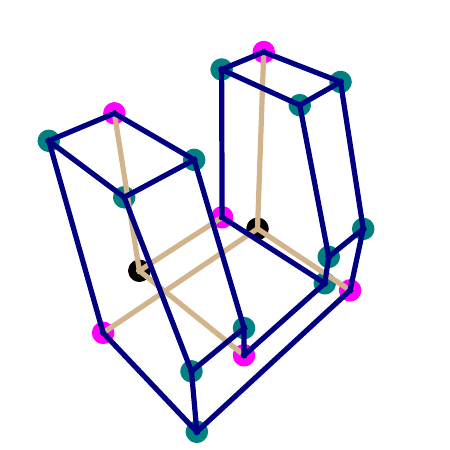}
         & \includegraphics[height=0.14\linewidth]{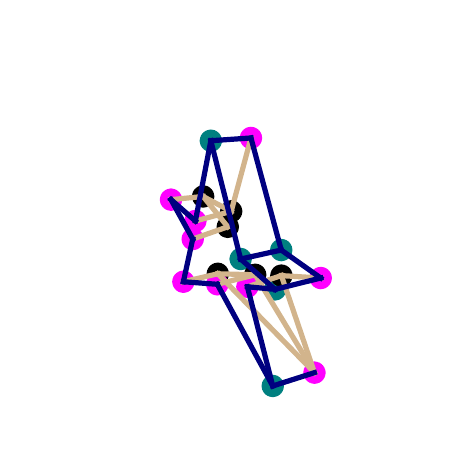}\\
         \multicolumn{7}{c}{
         \begin{tikzpicture}
         \draw [dashed] (1,0) -- node[midway,fill=white] {Our Results} ++(10,0);
         \end{tikzpicture}
         }
         \\
         & \includegraphics[height=0.14\linewidth]{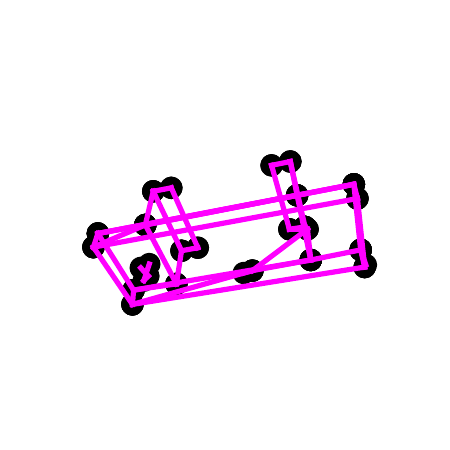}
         & \includegraphics[height=0.14\linewidth]{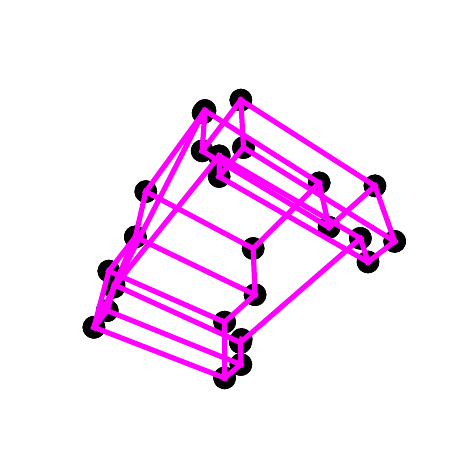}
         & \includegraphics[height=0.14\linewidth]{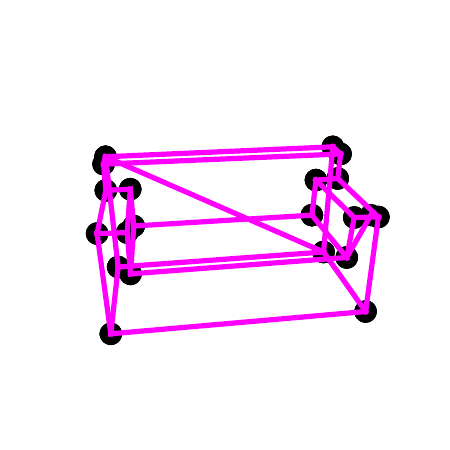}
         & \includegraphics[height=0.14\linewidth]{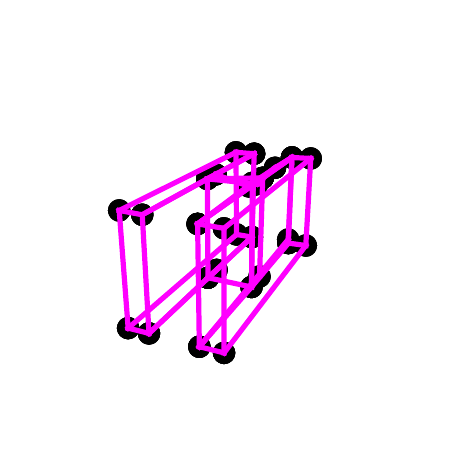}
         & \includegraphics[height=0.14\linewidth]{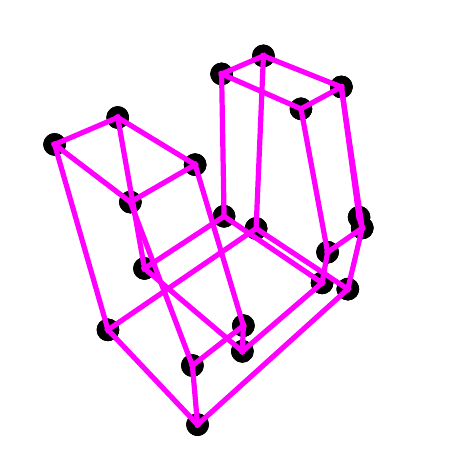}
         & \includegraphics[height=0.14\linewidth]{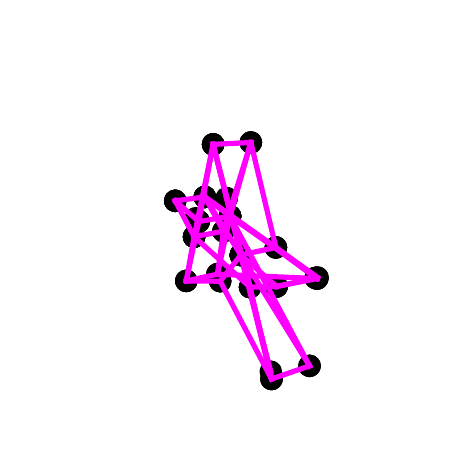}\\
         \multicolumn{7}{c}{
         \begin{tikzpicture}
         \draw [dashed] (1,0) -- node[midway,fill=white] {PC2WF~\cite{liu2021pc2wf}} ++(10,0);
         \end{tikzpicture}
         }\\
         & \includegraphics[height=0.14\linewidth]{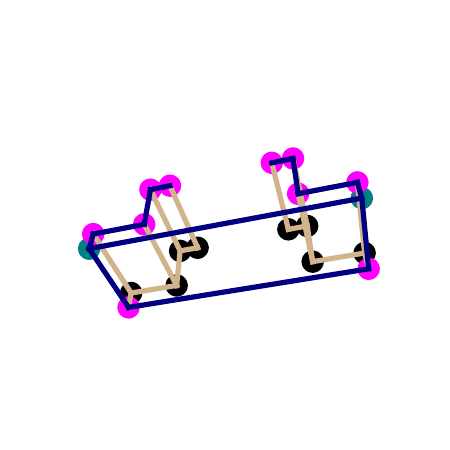} 
         & \includegraphics[height=0.14\linewidth]{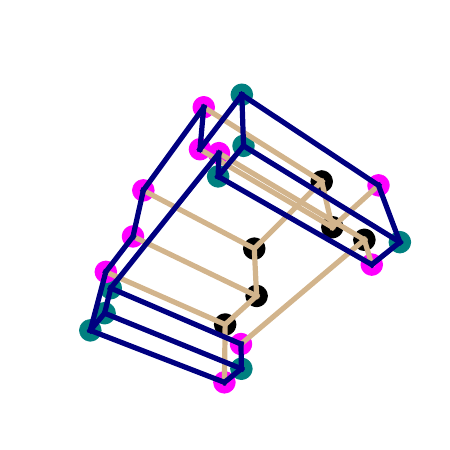}
         & \includegraphics[height=0.14\linewidth]{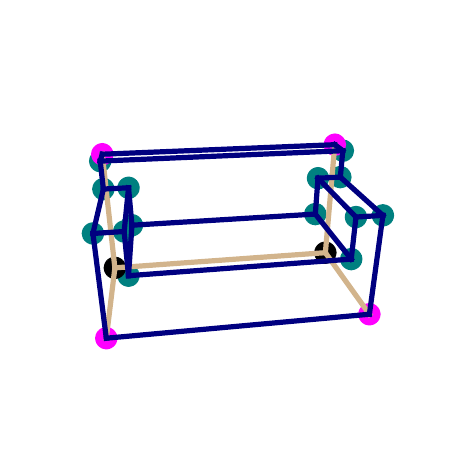}
         & \includegraphics[height=0.14\linewidth]{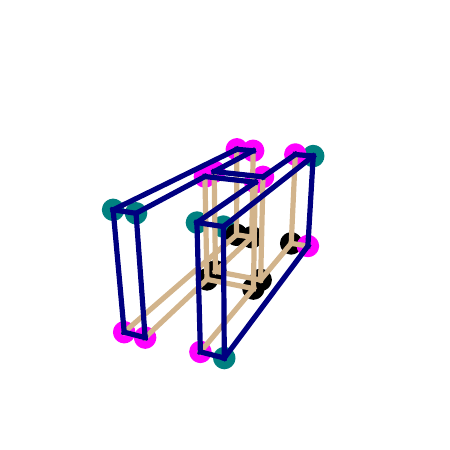}
         & \includegraphics[height=0.14\linewidth]{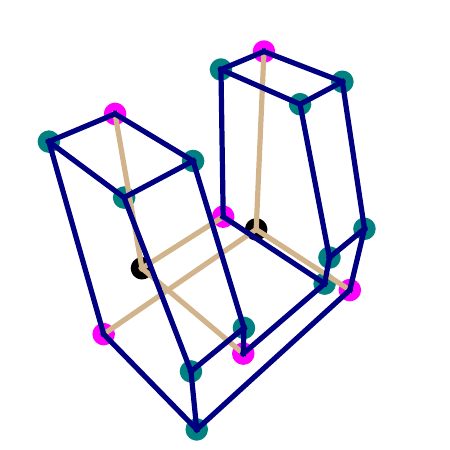}
         & \includegraphics[height=0.14\linewidth]{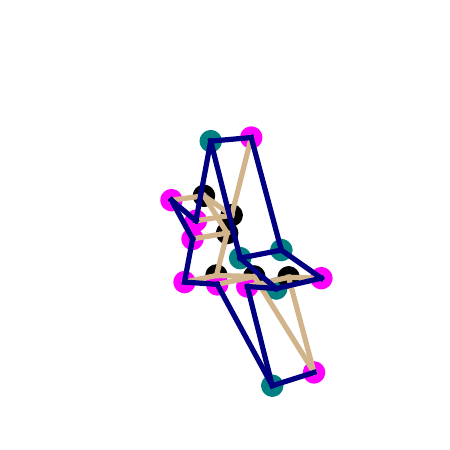}\\
         \multicolumn{7}{c}{Ground Truth}
        \\\midrule
        \multirow{3}{*}{
        \begin{tikzpicture}
         \node [text centered, rotate=90, text width=0.22\linewidth] {Novel View 1};
        \end{tikzpicture}
        }
       & \includegraphics[height=0.14\linewidth]{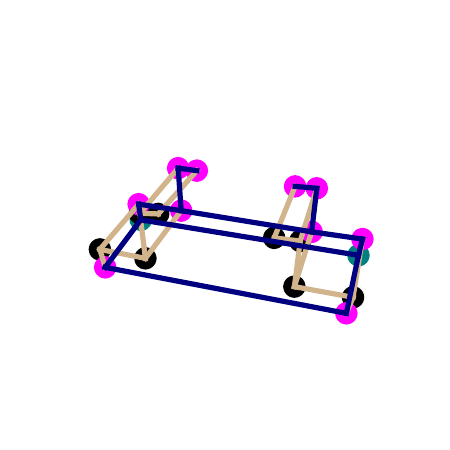} 
         & \includegraphics[height=0.14\linewidth]{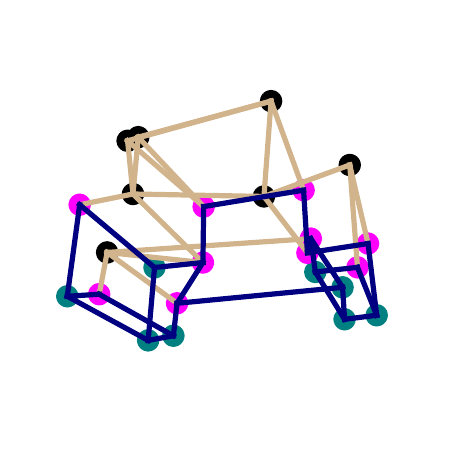}
         & \includegraphics[height=0.14\linewidth]{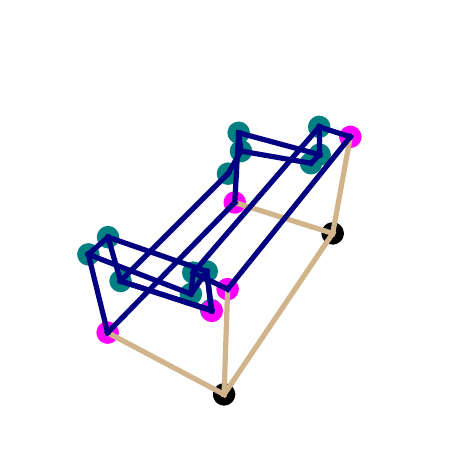}
         & \includegraphics[height=0.14\linewidth]{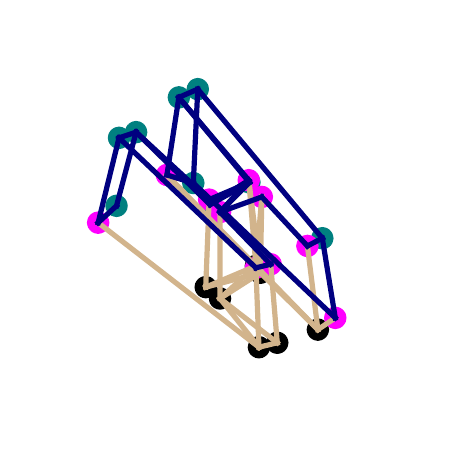}
         & \includegraphics[height=0.14\linewidth]{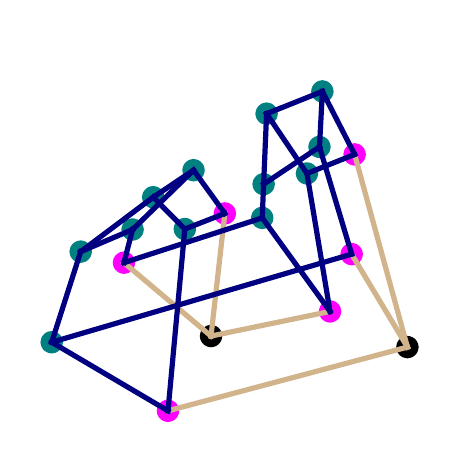}
         & \includegraphics[height=0.14\linewidth]{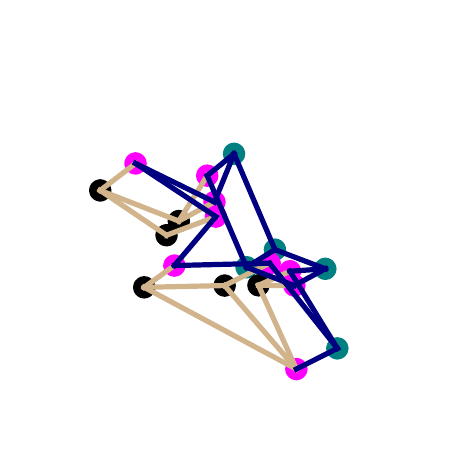}\\
         \multicolumn{7}{c}{
         \begin{tikzpicture}
         \draw [dashed] (1,0) -- node[midway,fill=white] {Our Results} ++(10,0);
         \end{tikzpicture}
         }\\
         & \includegraphics[height=0.14\linewidth]{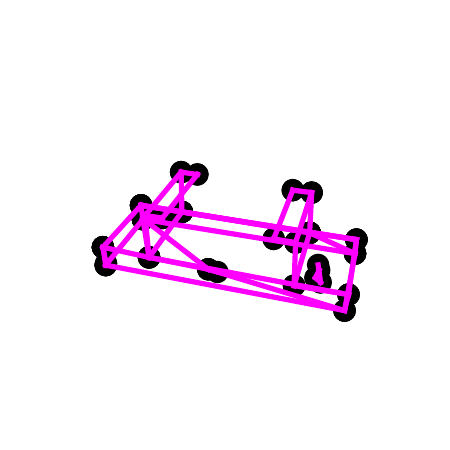}
         & \includegraphics[height=0.14\linewidth]{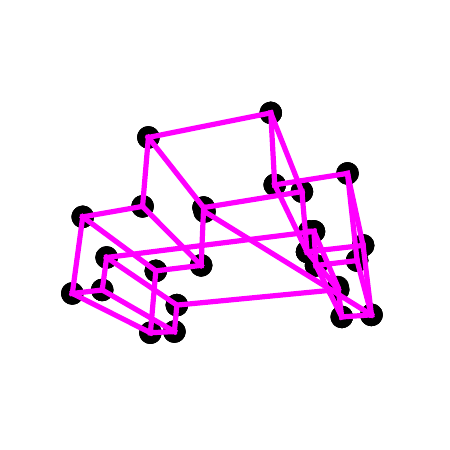}
         & \includegraphics[height=0.14\linewidth]{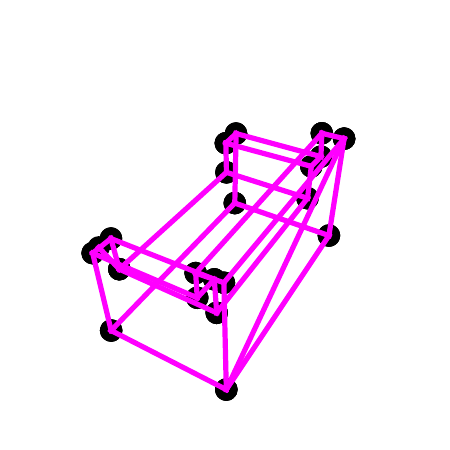}
         & \includegraphics[height=0.14\linewidth]{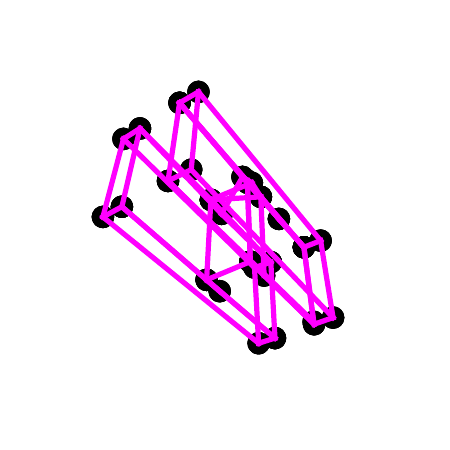}
         & \includegraphics[height=0.14\linewidth]{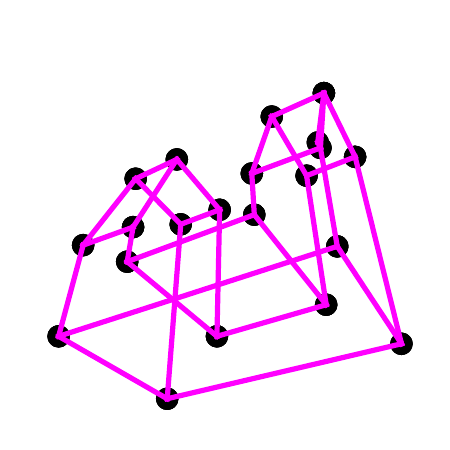}
         & \includegraphics[height=0.14\linewidth]{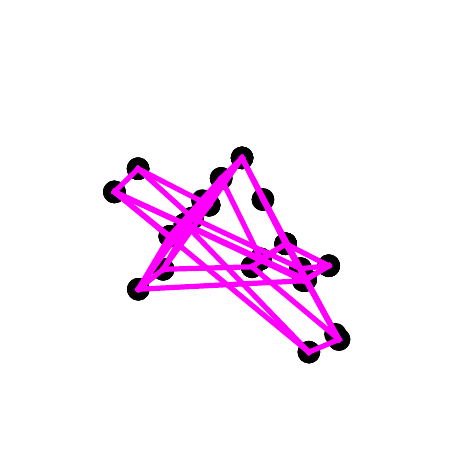}\\
         \multicolumn{7}{c}{
         \begin{tikzpicture}
         \draw [dashed] (1,0) -- node[midway,fill=white] {PC2WF~\cite{liu2021pc2wf}} ++(10,0);
         \end{tikzpicture}
         }\\
        & \includegraphics[height=0.14\linewidth]{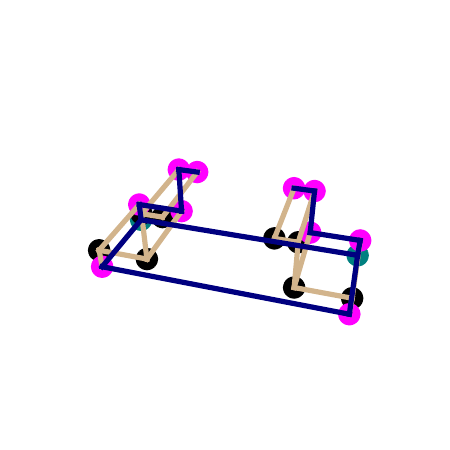} 
         & \includegraphics[height=0.14\linewidth]{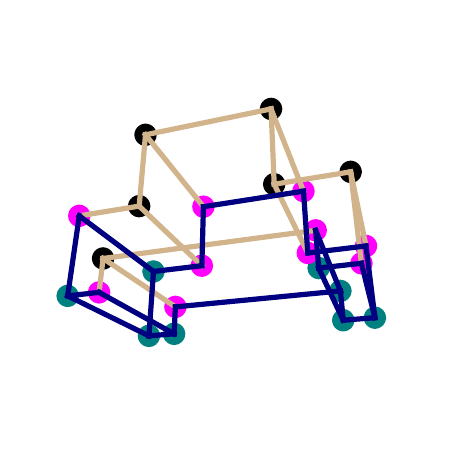}
         & \includegraphics[height=0.14\linewidth]{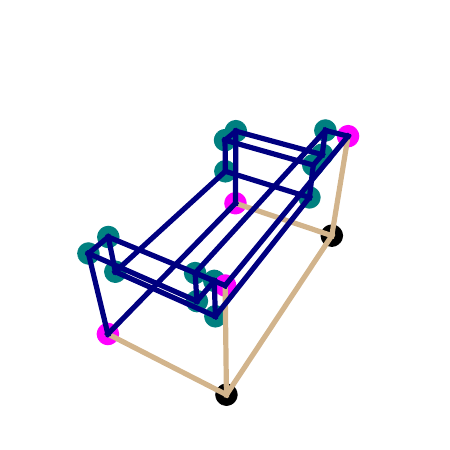}
         & \includegraphics[height=0.14\linewidth]{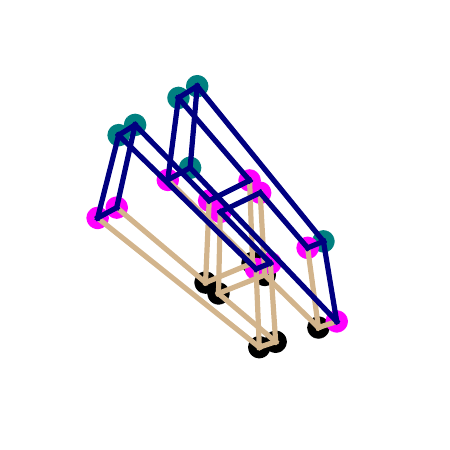}
         & \includegraphics[height=0.14\linewidth]{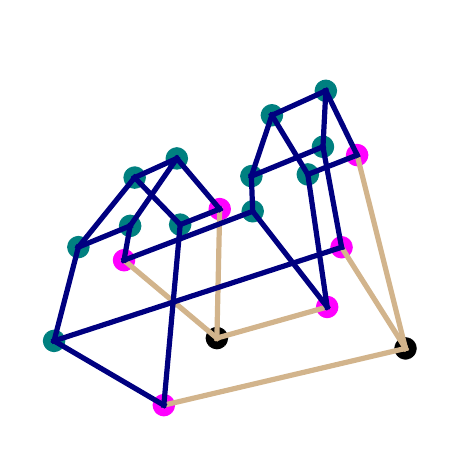}
         & \includegraphics[height=0.14\linewidth]{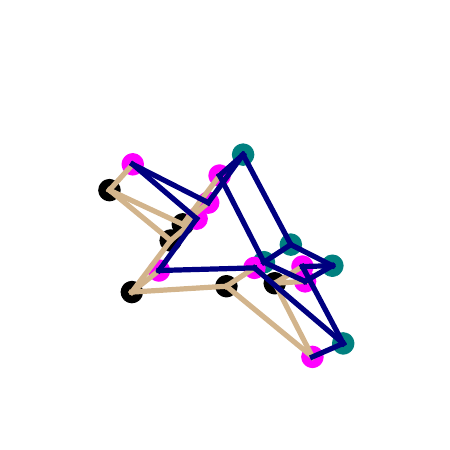}\\
        \multicolumn{7}{c}{Ground Truth}
        \\\midrule
        \multirow{3}{*}{
        \begin{tikzpicture}
         \node [text centered, rotate=90, text width=0.22\linewidth] {Novel View 2};
        \end{tikzpicture}
        }
        & \includegraphics[height=0.14\linewidth]{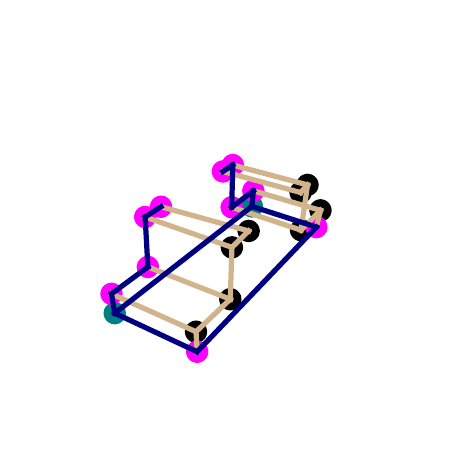} 
         & \includegraphics[height=0.14\linewidth]{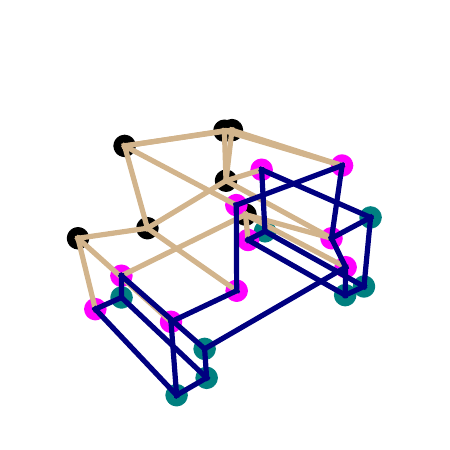}
         & \includegraphics[height=0.14\linewidth]{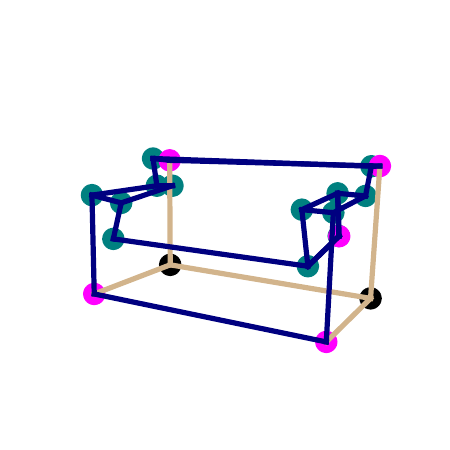}
         & \includegraphics[height=0.14\linewidth]{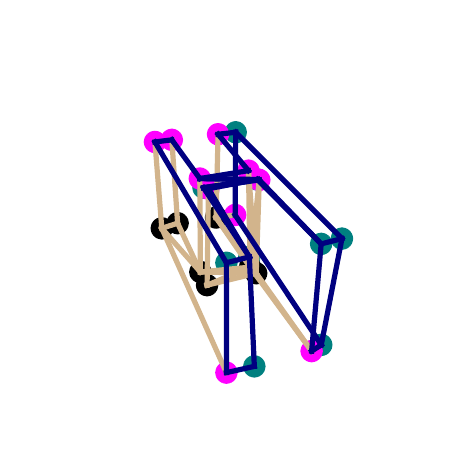}
         & \includegraphics[height=0.14\linewidth]{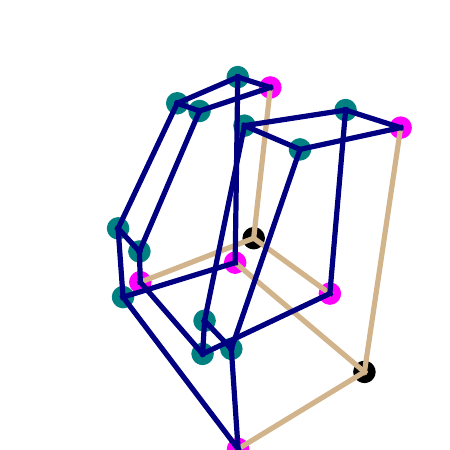}
         & \includegraphics[height=0.14\linewidth]{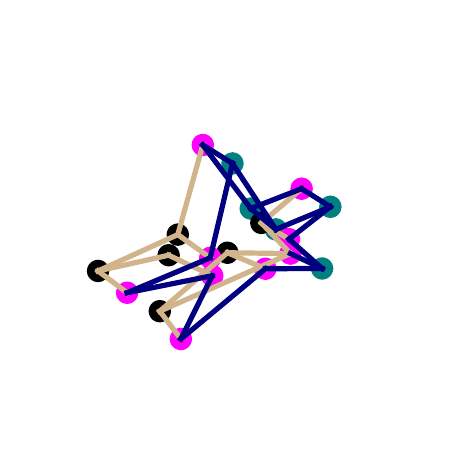}\\
        \multicolumn{7}{c}{
         \begin{tikzpicture}
         \draw [dashed] (1,0) -- node[midway,fill=white] {Our Results} ++(10,0);
         \end{tikzpicture}
         }\\
        & \includegraphics[height=0.14\linewidth]{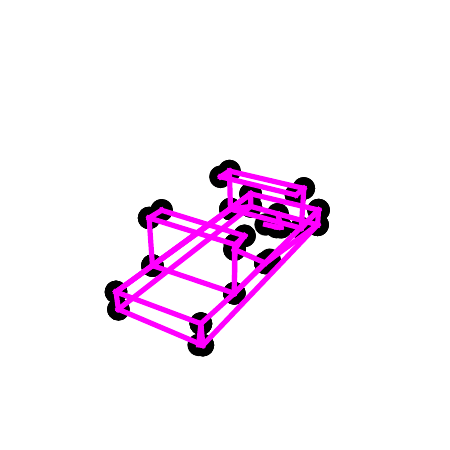}
         & \includegraphics[height=0.14\linewidth]{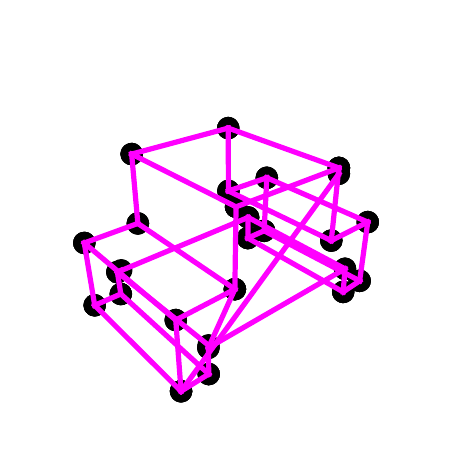}
         & \includegraphics[height=0.14\linewidth]{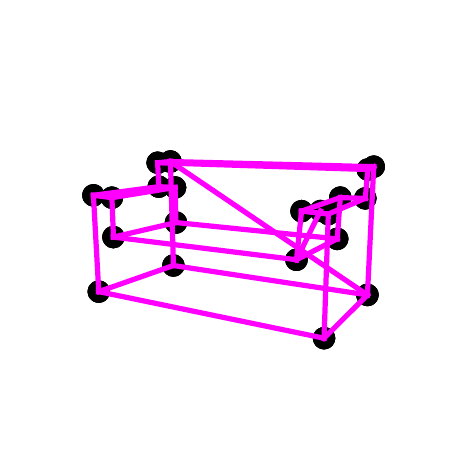}
         & \includegraphics[height=0.14\linewidth]{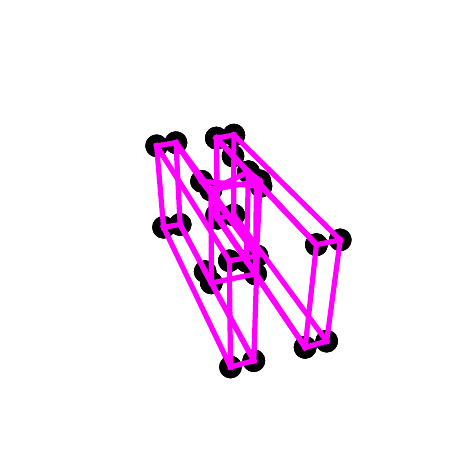}
         & \includegraphics[height=0.14\linewidth]{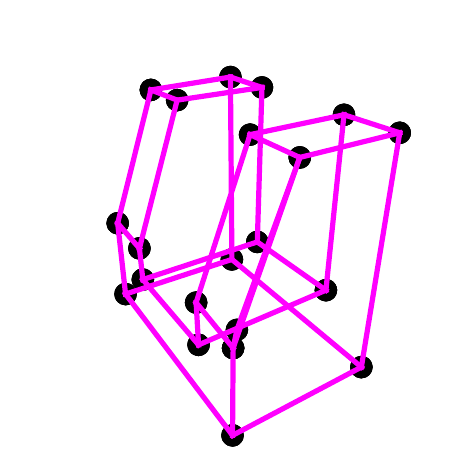}
         & \includegraphics[height=0.14\linewidth]{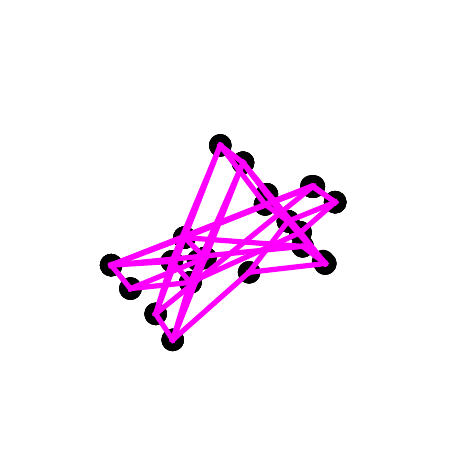}\\
        \multicolumn{7}{c}{
         \begin{tikzpicture}
         \draw [dashed] (1,0) -- node[midway,fill=white] {PC2WF~\cite{liu2021pc2wf}} ++(10,0);
         \end{tikzpicture}
         }\\
        & \includegraphics[height=0.14\linewidth]{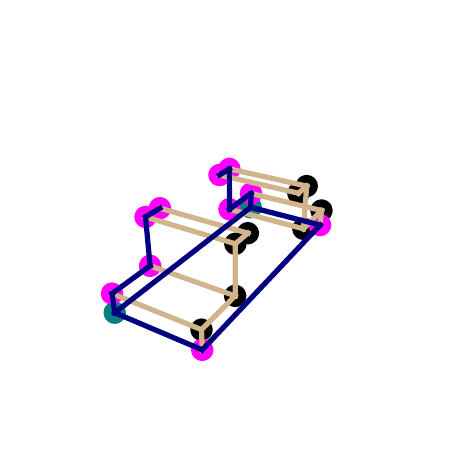} 
         & \includegraphics[height=0.14\linewidth]{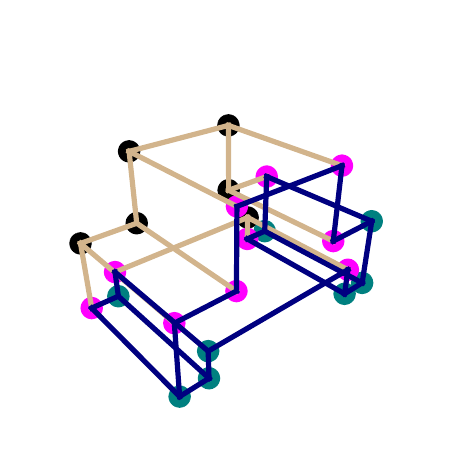}
         & \includegraphics[height=0.14\linewidth]{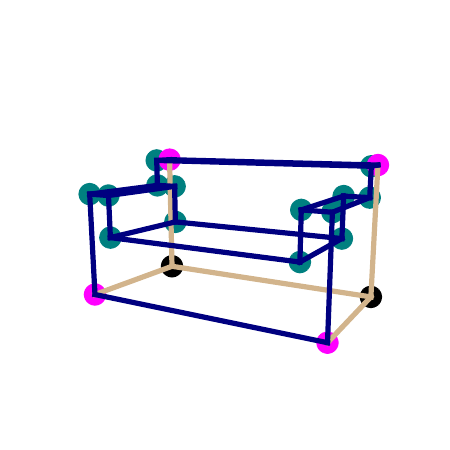}
         & \includegraphics[height=0.14\linewidth]{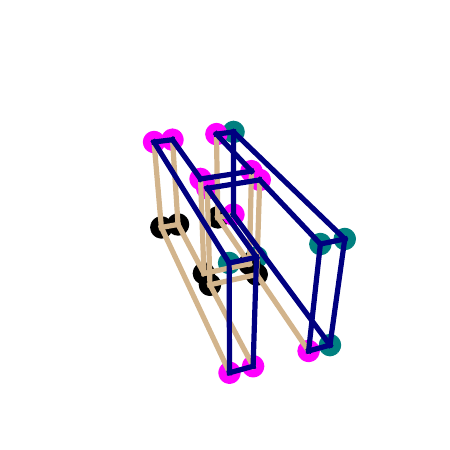}
         & \includegraphics[height=0.14\linewidth]{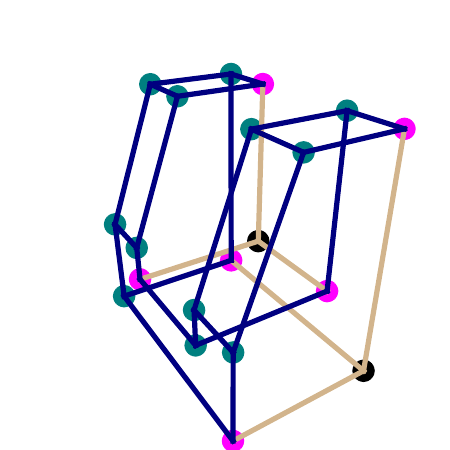}
         & \includegraphics[height=0.14\linewidth]{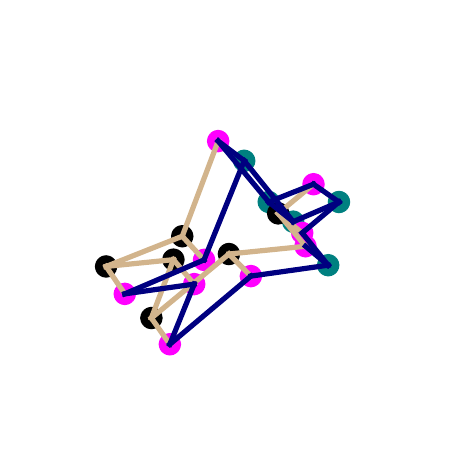}\\
        \multicolumn{7}{c}{Ground Truth}\\\bottomrule
    \end{tabular}
    }
    \caption{Visualization results of our DSG model (from single-view image) and PC2WF\cite{liu2021pc2wf} (from pointclouds) compared with ground truths in different views.}
    \label{fig:visualization-sup}
\end{figure}

We show more visualization results compared with PC2WF~\cite{liu2021pc2wf} and ground truth in ~\cref{fig:visualization-sup}.

\begin{figure}[!t]
    \centering
    \includegraphics[width=0.60\linewidth]{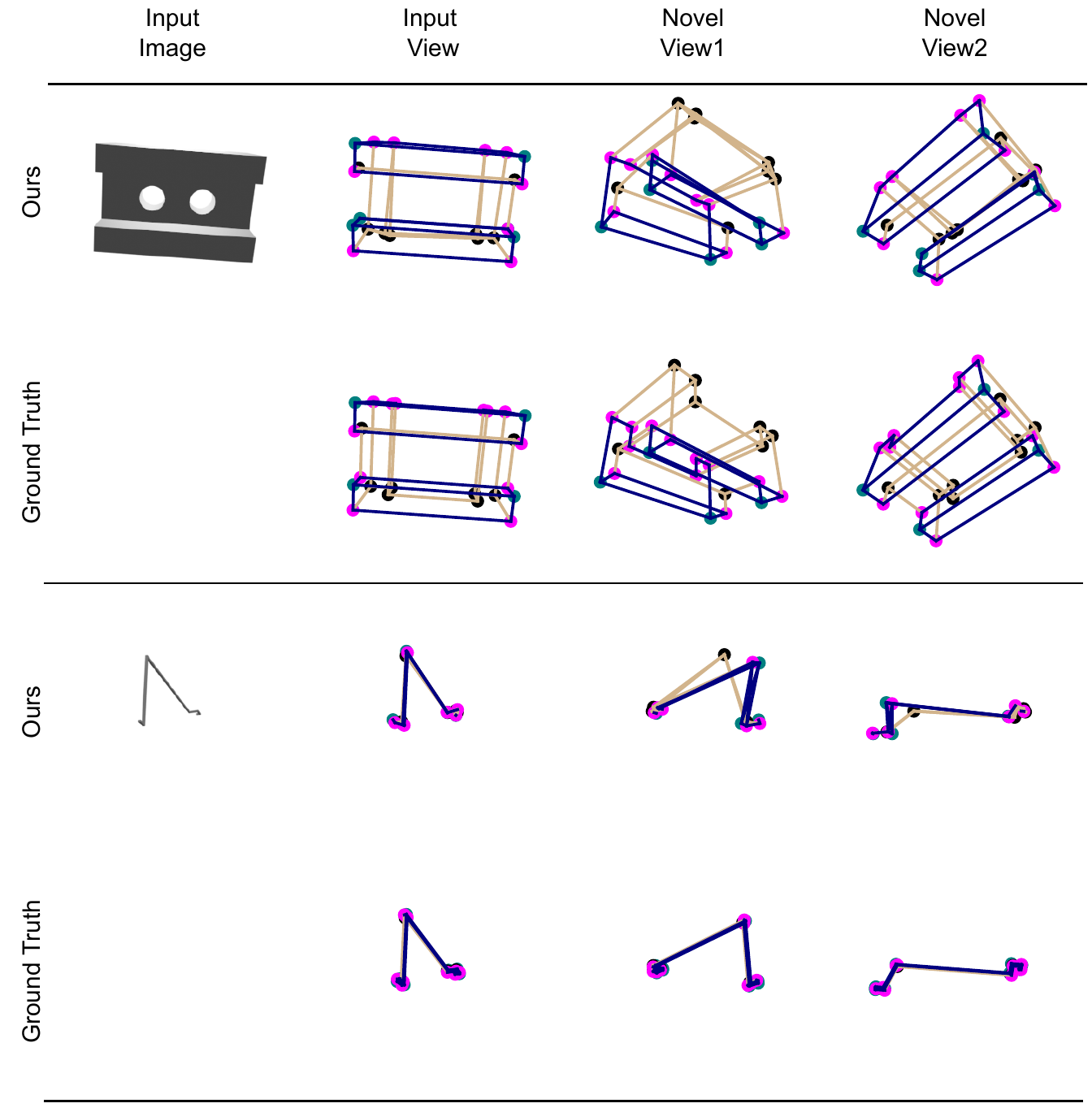}
    \caption{The failure cases of our DSG model}
    \label{fig:failure}
\end{figure}
\section{Failure Case and Limitation}
\cref{fig:failure} shows the failure case of our method compared with the ground truth. As shown at the top of the figure, our method would fail to fill the invisible parts accurately when the observation viewpoint contains much less visible cues. But methods that take point clouds as input would not be limited by the viewpoint. The bottom of the figure shows that our method would fail to reconstruct the ``tiny" parts of the objects, which is mainly caused by the resolution limitation of the input images. 

\end{document}